%% file: main.tex
\pdfoutput=1
\def\finalversion{}
\documentclass[10pt,twocolumn,letterpaper]{article}
\newif\iffinal\ifdefined\finalversion\finaltrue\fi
\iffinal\usepackage[pagenumbers]{cvpr}\else\usepackage[review]{cvpr}\fi
\input{preamble}
\input{sections/counts}
\definecolor{cvprblue}{rgb}{0.21,0.49,0.74}
\usepackage[pagebackref,breaklinks,colorlinks,allcolors=cvprblue]{hyperref}
\hypersetup{pdftitle={IrisMix: Compositional Makeup Transfer and Editing with Layered Diffusion}}
\def\paperID{2685}
\def\confName{CVPR}
\def\confYear{2027}
\begin{document}
\makeatletter
\def\abstract{\iftoggle{cvprpagenumbers}{}{\thispagestyle{empty}}\centerline{\large\bf Abstract}\vspace*{2pt}\noindent\it\ignorespaces}
\def\endabstract{\vspace*{0pt}}
\makeatother
\raggedbottom
\addtocontents{toc}{\protect\setcounter{tocdepth}{-10}}
\title{IrisMix: Compositional Makeup Transfer and Editing with Layered Diffusion}
\iffinal\input{authors}\else\author{Anonymous CVPR submission\\Paper ID \paperID}\fi
\twocolumn[{%
\renewcommand\twocolumn[1][]{#1}%
\maketitle
\vspace{-39pt}%
\begin{center}
\captionsetup{type=figure,skip=0pt}
\includegraphics{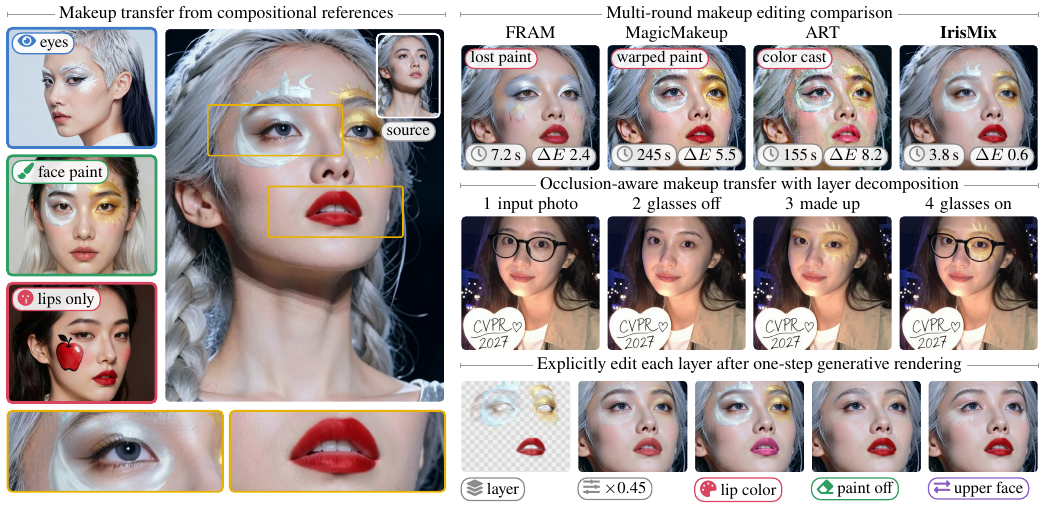}
\captionof{figure}{\textbf{Compositional makeup transfer and editing with layered diffusion.} \method{} recovers each reference as an explicit layer, so selected parts of several references compose on one source face, while FRAM~\cite{gao2026fram} drops the face paint, and MagicMakeup~\cite{magicmakeup} and ART~\cite{art2026}, which apply the references in turn to their own outputs, warp the designs and add color casts ($\Delta E$: mean color change outside the face; clock: time for the three references). Layers also make transfer occlusion-aware, and all layers stay editable after rendering.}
\label{fig:teaser}
\end{center}%
\vspace{-10pt}}]%
\begingroup
\setlength{\textfloatsep}{12pt plus 2pt minus 2pt}
\setlength{\dbltextfloatsep}{12pt plus 2pt minus 2pt}
\setcounter{topnumber}{3}
\begin{figure*}[t]\vspace{-\dbltextfloatsep}\vspace{-5mm}\end{figure*}%
\input{sections/0-abstract}
\input{sections/1-introduction}
\input{sections/2-related}
\input{sections/3-data}
\input{sections/3-method}
\input{sections/4-experiments}

\input{sections/5-conclusion}
\FloatBarrier
\endgroup
\setcounter{topnumber}{99}
\clearpage
\section*{LLM Usage Statement}
We used large language model (LLM) assistants in this work. They helped write and debug code for data processing, training, evaluation and figures. LLMs also helped polish the wording of the paper. LLMs are not part of IrisMix and do not produce any reported result. The authors designed the method and the experiments, checked all code, data and text, and take full responsibility for the content of this paper.
{\small\bibliographystyle{ieeenat_fullname}\bibliography{refs,\iffinal refs_self_final\else refs_self_review\fi}}
\clearpage\appendix
\maketitlesupplementary
\supplementfloats
\setcounter{dbltopnumber}{3}
\renewcommand{\dbltopfraction}{0.95}
\renewcommand{\textfraction}{0.05}
\setlength{\textfloatsep}{10pt plus 2pt minus 2pt}
\setlength{\dbltextfloatsep}{10pt plus 2pt minus 2pt}
\setlength{\floatsep}{8pt plus 2pt minus 2pt}
\setlength{\dblfloatsep}{8pt plus 2pt minus 2pt}
\addtocontents{toc}{\protect\setcounter{tocdepth}{2}}
{\makeatletter\renewcommand{\contentsname}{Contents}\renewcommand\l@section[2]{\@dottedtocline{1}{0em}{1.6em}{\bfseries #1}{#2}}\renewcommand\l@subsection{\@dottedtocline{2}{1.6em}{2.3em}}\footnotesize\tableofcontents\makeatother}
\input{sections/A-analysis}
\input{sections/A-model}
\input{sections/B-impl}
\input{sections/C-data}
\input{sections/G-gallery}
\input{sections/D-protocol}
\input{sections/E-eval}
\input{sections/G-realworld}\nocite{avatarmakeup,curatedmakeup2026,diffclean,dreammakeup,genomnimatte,gorgeous,iclight,kips2022,layerd,sdmatte,synthlight,tinybeauty,zim}
\end{document}

%% file: preamble.tex
\usepackage{graphicx}
\usepackage{amsmath,amsfonts,amssymb}
\usepackage{booktabs,multirow}
\usepackage{xcolor,colortbl,array}
\usepackage{caption}
\usepackage{subcaption}
\usepackage{pifont}
\usepackage{placeins}
\definecolor{pEye}{HTML}{3E7CC9}\definecolor{pLip}{HTML}{D9455F}\definecolor{pPaint}{HTML}{2E9E62}\definecolor{pUp}{HTML}{8A5CC7}\definecolor{pLow}{HTML}{E07B22}
\definecolor{oursrow}{HTML}{EAF1FB}\definecolor{refgray}{gray}{0.45}
\colorlet{top1}{pLip!42}\colorlet{top2}{pLip!24}\colorlet{top3}{pLip!11}   %
\newcommand{\method}{IrisMix}
\newcommand{\up}{\ensuremath{\uparrow}}\newcommand{\dn}{\ensuremath{\downarrow}}
\newcommand{\cmark}{\ding{51}}\newcommand{\xmark}{\textcolor{refgray}{\ding{55}}}

\newcommand{\supplementfloats}{\renewcommand{\dbltopfraction}{0.90}\setcounter{dbltopnumber}{2}\renewcommand{\textfraction}{0.10}\renewcommand{\dblfloatpagefraction}{0.8}}

%% file: sections/counts.tex
\newcommand{\nCreativePairs}{13{,}273}  %
\newcommand{\nCreativePairsK}{13.3k}  %
\newcommand{\nSynPairs}{4{,}794}  %
\newcommand{\nSynPairsK}{4.8k}  %

%% file: authors.tex
\author{Chia-Ming Lee$^{1,2}$\enspace Ming-Ching Chang$^{1}$\enspace Xin Li$^{1}$\enspace Hong-Han Shuai$^{2}$\enspace Yu-Lun Liu$^{2}$\enspace Chih-Chung Hsu$^{2}$\\[2pt]
$^{1}$State University of New York at Albany\qquad $^{2}$National Yang Ming Chiao Tung University\\[2pt]
{\small\url{https://ming053l.github.io/IrisMix/}}}

%% file: sections/0-abstract.tex
\begin{abstract}
Makeup design involves combining references, refining individual parts, and revisiting earlier choices. Diffusion-based makeup transfer produces realistic results, but its flattened outputs keep no explicit makeup, so subsequent revisions may require another generation pass that can alter identity and accepted regions. We introduce \method{}, which couples explicit makeup layers for editing with diffusion rendering for appearance adaptation. One diffusion transformer with dual LoRAs first decomposes each reference into a bare face and reusable color and coverage layers by triangulation matting. After layers from one or more references are aligned and composed on the source face that receives the makeup, the same transformer renders the composite, correcting compositing artifacts and adapting the appearance to facial geometry and scene illumination. Because occluders are layers too, glasses can be lifted off before transfer and restored afterwards. The rendered result is an editable layer over the source face, so color, strength and selected regions can be adjusted without resampling. When a change needs another render, \method{} recomposes cached layers on the source face instead of editing previous outputs. To train the decomposer and renderer, we construct LayerBeauty, which pairs portraits with explicit compositing layers: bare-first synthesis gives independent bare targets for everyday makeup, and makeup-first synthesis keeps intricate artistic designs. The decomposer beats layered and non-layered methods; composing its layers alone preserves identity best on three transfer benchmarks, and rendering further improves design fidelity. \method{} decomposes and renders a full look in under 4\,s, and over repeated edits it drifts far less in never-edited regions, enabling interactive and personalized makeup design.
\end{abstract}

%% file: sections/1-introduction.tex
\vspace{-2mm}
\section{Introduction}
\label{sec:intro}
\begin{figure}[t]
\centering
\includegraphics[width=\columnwidth]{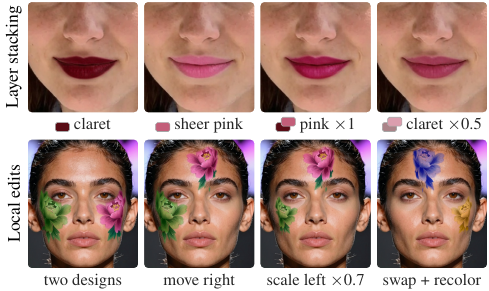}
\caption{\textbf{Composability and editability of \method{}.} By lifting the makeup off a photograph and decomposing it into layers, users can locally edit, stack and change the color and placement of any part, while the renderer adapts each edited look to the source's geometry and lighting so that the makeup stays photorealistic.}\vspace{-4mm}
\label{fig:apps}
\end{figure}

Makeup transfer enables users to preview different looks by adapting makeup from reference photographs to a source face. Early methods use generative adversarial networks to transfer makeup while preserving facial identity~\cite{beautygan,psgan}. More recent diffusion models have improved the visual quality and fidelity of transferred makeup~\cite{stablemakeup,shmt,fluxmakeup,art2026}, expanding the possibilities for personalized makeup design.

In practice, makeup design often involves several rounds of revision. Users may combine elements from multiple references, adjust individual parts, and revisit earlier choices while retaining makeup they have already accepted. Regional control~\cite{elegant,gao2026fram,magicmakeup} supports such customization by allowing users to specify where makeup is applied. However, the resulting images do not retain makeup as a separate representation that users can directly edit and reuse. Revising an existing look may therefore require another generation pass. When revisions operate on previously generated images, unintended changes to facial identity, colors, and accepted regions can accumulate across rounds.

To reduce reliance on repeated generation, makeup can be retained as an explicit \emph{layer representation}, with coverage and color separated from the underlying face. Users can then select, stack, recolor, and adjust the strength of makeup through direct layer operations. However, using layers for transfer presents two challenges: extracting detailed makeup from photographs and adapting it to a new face. Earlier layered approaches rely on fixed facial regions or canonical face representations~\cite{chau2026realtime,makeupextract2023,makeupprior2024}, which can restrict the designs they represent. General layer generators~\cite{qwenlayered,stablelayers} are computationally expensive and are not specialized for makeup, while decomposition through makeup removal~\cite{mad} can miss fine details. Even accurately extracted layers can produce broken strokes, visible boundaries, and illumination mismatches when warped and blended onto another face. This calls for a framework that efficiently recovers reusable makeup layers and adapts their composition to the receiving face while retaining editability.

We introduce \method{}, which combines reusable makeup layers with diffusion rendering for iterative makeup design, as illustrated in \cref{fig:teaser}. A shared diffusion transformer~\cite{flux2} uses separate LoRA adapters~\cite{hu2022lora} for decomposition and rendering, each in a single step. The decomposer predicts a bare face and makeup over white and black backgrounds, enabling triangulation matting~\cite{smithblinn} to recover coverage and color. Selected parts from one or more references are then aligned and composed on the source face, and the renderer repairs compositing artifacts and adapts the makeup to the source's geometry and illumination. The rendered result is represented as an editable layer over the source face, allowing color, strength, and selected regions to be adjusted without another generation pass. When a revision requires rendering, \method{} recomposes cached layers on the original source face instead of modifying a previously generated image.~\cref{fig:apps} illustrates the resulting editing capabilities, including layer stacking and local adjustments to makeup color, strength, and placement.

To train this framework, we construct \textbf{LayerBeauty}, a dataset pairing portraits with explicit makeup layers for everyday and artistic looks. For everyday makeup, we add makeup to independently generated bare faces, preserving bare targets that do not depend on makeup removal. For artistic looks, we generate the makeup designs first and obtain their bare counterparts through removal, retaining intricate patterns in the original portraits. These complementary data provide supervision for layer decomposition and rendering. A harmonization cycle improves layer extraction from real photographs, while self-transfer supervision teaches the renderer to repair alignment artifacts without requiring paired examples of transfer between different identities. The contributions are summarized as follows.\nopagebreak
\begin{itemize}[beginpenalty=10000]

\item We introduce \method{}, a framework for compositional makeup transfer that retains an editable representation after rendering. Direct layer operations and recomposition from cached layers support repeated revisions without repeatedly modifying previously generated images.

\item We develop a shared diffusion transformer for layer decomposition and rendering, each in a single step. Triangulation matting recovers makeup coverage and color, while self-transfer supervision teaches the renderer to repair compositing artifacts without cross-identity pairs.

\item We construct \textbf{LayerBeauty}, comprising \nSynPairs{} everyday and \nCreativePairs{} artistic looks paired with explicit makeup layers. Its complementary synthesis provides independent bare targets for everyday makeup and preserves intricate artistic designs.

\item Experiments on three transfer benchmarks show that rendering improves makeup fidelity over warping alone. Across seven edits on 30 looks, \method{} reduces mean color drift outside edited regions by 88\% relative to sequential FLUX.2-klein editing, while producing more accurate strength adjustments and reference replacements.

\end{itemize}

%% file: sections/2-related.tex
\vspace{-1.5mm}
\section{Related Works}
\begin{figure*}[t]
\centering
\vspace{-5mm}
\includegraphics[width=0.98\textwidth]{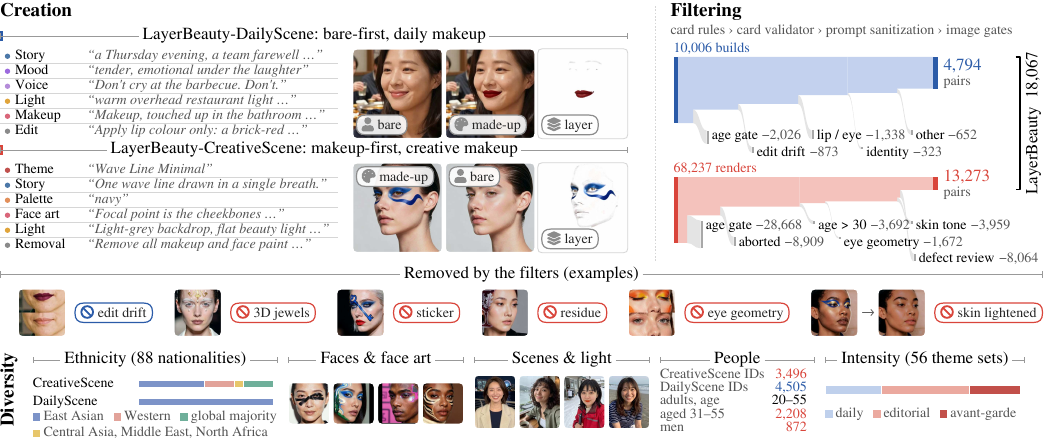}\vspace{-1.5mm}
\caption{\textbf{LayerBeauty} is a carefully curated dataset of everyday and creative makeup with explicit layers, spanning many themes, identities and ethnicities. Bare-first DailyScene adds makeup to bare faces; makeup-first CreativeScene removes it.}\vspace{-2mm}
\label{fig:datapipe}\vspace{-2mm}
\end{figure*}
\begin{table}[t]
\centering
\caption{\textbf{Makeup transfer datasets.} Counts are images for unpaired sets and pairs otherwise. Bare: same-person bare pairs; Indep.: independent of removal. Layer lists supplied representations; min.\ $\boldsymbol\alpha$: minimum coverage; +syn.: synthetic makeup.}
\label{tab:datasets}
\scriptsize\setlength{\tabcolsep}{1.7pt}

\resizebox{\linewidth}{!}{%
\begin{tabular}{@{}lrrccccl@{}}
\toprule
 & & & \multicolumn{3}{c}{Supervision} & & \\
\cmidrule(lr){4-6}
Dataset & Size & Resolution & Bare & Indep. & Layer & Styles & Faces\\\midrule
MT~\cite{beautygan} & 3.8k & 361 & \xmark & -- & \xmark & daily & real\\
Makeup-Wild~\cite{psgan} & 0.8k & 256 & \xmark & -- & \xmark & daily & real\\
LADN~\cite{ladn} & ${\sim}$0.8k & -- & \xmark & -- & \xmark & dramatic & real\\
CPM-Real~\cite{cpm} & 3.9k & -- & \xmark & -- & \xmark & patterns & real\\
BeautyFace~\cite{beautyrec} & 3.0k & 512 & \xmark & -- & \xmark & varied & real\\
MF2K~\cite{art2026} & 8.6k & 2048 & \xmark & -- & \xmark & daily+art & real\\\midrule
FFHQ-Makeup~\cite{ffhqmakeup} & 90k & 512 & \cmark & \cmark & residual & MT/LADN & real+syn.\\
BMS~\cite{beautybank} & 324k & 512 & \cmark & \cmark & \xmark & prompts & real+syn.\\
HQMT~\cite{fluxmakeup} & 56.9k & -- & \cmark & \cmark & \xmark & prompts & real+syn.\\
MagicMakeup~\cite{magicmakeup} & 6.8k & 1024 & \cmark & \xmark & \xmark & varied & real\\\midrule
\rowcolor{oursrow}\multicolumn{8}{@{}l}{LayerBeauty (ours)}\\
\rowcolor{oursrow}\quad CreativeScene & \nCreativePairsK{} & 1024 & \cmark & \xmark & min.\ $\boldsymbol\alpha$ & artistic & synth.\\
\rowcolor{oursrow}\quad DailyScene & \nSynPairsK{} & $768{\times}1024$ & \cmark & \cmark & \cmark & daily & synth.\\\bottomrule
\end{tabular}}\vspace{-3mm}
\end{table}

\label{sec:related}
\noindent\textbf{Generative makeup transfer.} Prior work has steadily improved fidelity and control, but every method still returns a flat image. GANs learn transfer from unpaired photographs through histogram losses, correspondence or style codes~\cite{beautygan,psgan,scgan,elegant,csdmt}, yet implicit correspondence struggles with heavy and creative makeup. Diffusion models raise fidelity, from reference-conditioned UNets~\cite{stablemakeup,shmt,mad} to diffusion transformers trained on filtered synthetic pairs, rewards or real-reference cycles~\cite{fluxmakeup,realbeauty,art2026}, but they regenerate the face and can alter identity. Region-aware features and tokens~\cite{gao2026fram,clip,magicmakeup,evomakeup,makeupmirror} add control over where makeup goes, yet each edit is a new generation pass. In contrast, \method{} recovers reusable layers that support direct edits and one-step rendering on a new source.

\noindent\textbf{Layered makeup representations.} Chau \etal~\cite{chau2026realtime} predict semi-transparent RGBA layers for fixed regions (eyes, lips, cheeks), learned from template pseudo-labels, and alpha-blend them in real time, so designs crossing regions are cut and lighting is ignored. Yang \etal~\cite{makeupextract2023,makeupprior2024} decompose the albedo of a FLAME UV texture into skin, makeup and alpha, later with learned makeup priors; the layer is confined to the skin UV region and inherits 3D-fit errors. RealBeauty's~\cite{realbeauty} layer is a rule-based color residual on a canonical face, without alpha. \method{} decomposes in image space with a generative prior; its layer can be opaque and span any region, and a learned renderer fits it.

\noindent\textbf{Generative layer decomposition.} Diffusion models recover transparent layers~\cite{layerdiffuse,artlayer2025}: LayerDecomp~\cite{layerdecomp} separates foreground and effects, DiffDecompose~\cite{diffdecompose} predicts contextual layers, and Qwen-Image-Layered~\cite{qwenlayered} generates multiple RGBA layers. Gu \etal~\cite{cycledecomp2026} couple decomposition and composition with a shared LoRA and cycle. Object masks, boxes and instructions~\cite{refdecomp2026,reveallayer2026} provide limited guidance for translucent makeup spanning facial parts.
\vspace{-2mm}

%% file: sections/3-data.tex
\section{LayerBeauty Dataset}
\label{sec:data}
\begin{figure*}[t]
\centering
\vspace{-5mm}
\includegraphics[width=\textwidth]{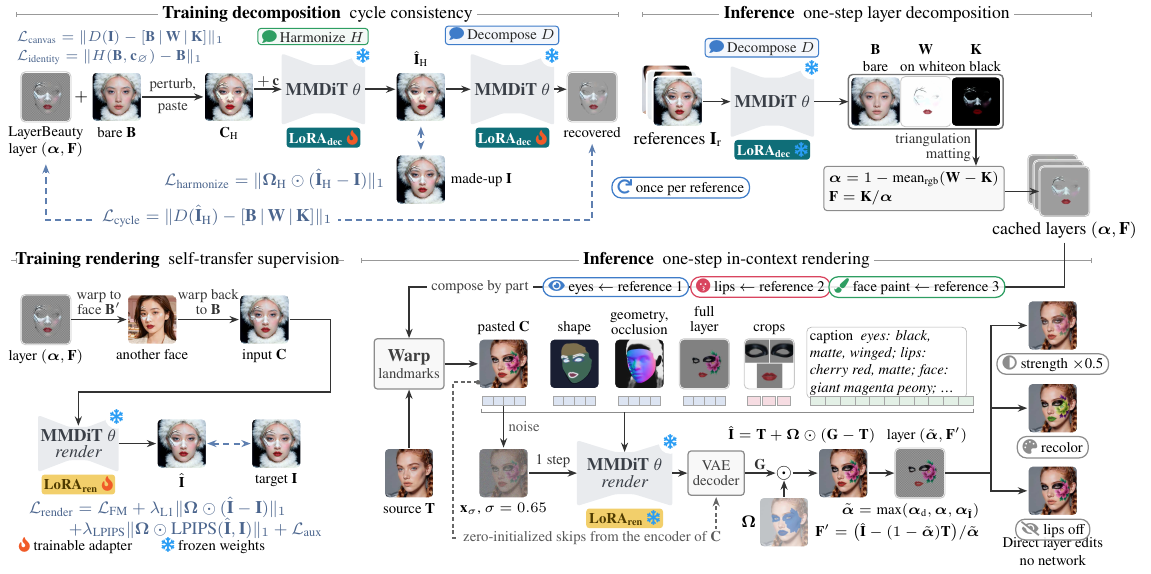}\vspace{-1.5mm}
\caption{\textbf{Overview of \method{}.} Two LoRAs on one frozen MMDiT (left) separate decomposition from rendering (right). The decomposer, trained with a harmonization cycle, recovers reusable layers $({\boldsymbol\alpha},{\mathbf F})$ from a predicted bare/white/black canvas $[{\mathbf B}\,|\,{\mathbf W}\,|\,{\mathbf K}]$, so each reference is decomposed once. The renderer composes the selected layers on ${\mathbf T}$ and renders natural, photorealistic makeup in one step, supervised by self-transfer through another face ${\mathbf B}'$. Source pixels outside ${\boldsymbol\Omega}$ are kept, and $(\tilde{\boldsymbol\alpha},{\mathbf F}')$ stays editable afterwards for persistent editability.}\vspace{-2mm}
\label{fig:method}\vspace{-2mm}
\end{figure*}
\begin{figure}[t]
\centering
\includegraphics[width=\linewidth]{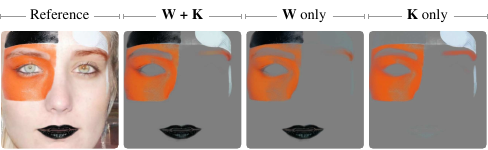}
\caption{\textbf{Triangulation matting for makeup decomposition.} Layers are shown over gray and computed from the same decomposer prediction. One background loses pigment close to its color: white paint over white (${\mathbf{W}}$ only), black lips over black (${\mathbf{K}}$ only).}\vspace{-2mm}
\label{fig:trimat}
\end{figure}
\begin{figure}[t]
\centering
\includegraphics[width=\linewidth]{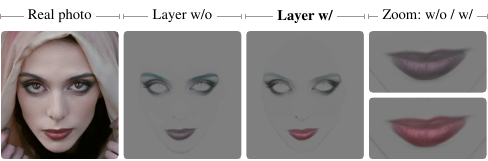}
\caption{\textbf{Effect of recomposition training} on a real MF2K~\cite{art2026} photograph unseen in training. Without harmonization and cycle consistency, the layer loses color; with them, it keeps the photograph's lip and eye colors, thereby generalizing to real photos.}\vspace{0mm}
\label{fig:harmcycle}
\end{figure}
Existing datasets such as MF2K~\cite{art2026}, BMS~\cite{beautybank} and HQMT~\cite{fluxmakeup} provide image pairs or regional masks~\cite{magicmakeup} (Table~\ref{tab:datasets}). The synthetic ones are limited by their generators and often fall short of contemporary aesthetics, and none provides the layer representation needed to reuse makeup on another face. LayerBeauty adds explicit layers for both everyday and creative looks, which supervise layer decomposition and in-context rendering. To obtain them, we model a made-up face ${\mathbf{I}}$ as an alpha-over composite~\cite{porterduff} of a makeup layer on the bare face ${\mathbf{B}}$,
\begin{equation}
{\mathbf{I}}={\boldsymbol{\alpha}}\odot {\mathbf{F}}+(1-{\boldsymbol{\alpha}})\odot {\mathbf{B}},
\label{eq:layer}
\end{equation}
with coverage ${\boldsymbol{\alpha}}\in[0,1]$ and pigment color ${\mathbf{F}}\in[0,1]^3$. Since such pairs are hard to collect in the real world, we follow agentic data generation~\cite{xia2026sage}: as shown in Figure~\ref{fig:datapipe}, a coding agent~\cite{claude} uses MiniMax-M2~\cite{minimaxm2} to batch-write prompts for diverse themes; Z-Image-Turbo~\cite{zimage} generates faces, FLUX.2-klein~\cite{flux2} adds makeup (DailyScene) or removes it (CreativeScene), and the agent screens out unnatural looks, with humans deciding disputed ones via MCP agent tools (Appendix~\ref{app:data}).

\noindent\textbf{LayerBeauty-DailyScene} is \emph{bare-first}: the editor adds light everyday makeup part by part to a rendered bare face, so the original bare face is kept and residual blush or liner is never labeled as skin; compositing gives pairs whose layers reproduce the made-up image by construction, with independent bare-face targets.

\noindent\textbf{LayerBeauty-CreativeScene} is \emph{makeup-first} and covers heavy and artistic looks: made-up portraits are rendered from validated theme cards, the editor removes the makeup, and pairs failing eye, skin-tone or defect checks are dropped. Since Eq.~\eqref{eq:layer} does not uniquely determine $({\boldsymbol{\alpha}},{\mathbf{F}})$, we take the minimum-alpha (color-to-alpha) solution~\cite{smithblinn,gimpcolortoalpha} with ${\mathbf{d}}={\mathbf{I}}-{\mathbf{B}}$, the least coverage keeping every channel in $[0,1]$, so unchanged skin stays uncovered:
\begin{equation}
{\boldsymbol{\alpha}}=\max_{c\in\{r,g,b\}}\max\big\{{\mathbf{d}}_c/(1-{\mathbf{B}}_c),\,-{\mathbf{d}}_c/{\mathbf{B}}_c\big\},
\label{eq:minalpha}
\end{equation}
and ${\mathbf{F}}={\mathbf{B}}+{\mathbf{d}}/{\boldsymbol{\alpha}}\in[0,1]^3$ where ${\boldsymbol{\alpha}}>0$. These layers reconstruct the image, but bare faces remain editor estimates. To evaluate generalization, we build a \emph{Composite} set by warping CreativeScene designs onto independent DailyScene bare faces (Appendix~\ref{app:protocol}).

%% file: sections/3-method.tex
\vspace{-2mm}
\section{Methodology}

\label{sec:method}
\begin{figure*}[t]\vspace{-\dbltextfloatsep}\vspace{-6mm}\end{figure*}%
As shown in Figure~\ref{fig:method}, one transformer with dual LoRAs decomposes references into layers and renders their composite on the source face as a persistently editable layer.
\vspace{1mm}
\subsection{Layer Representation and Backbone}
\label{sec:layer}

\noindent\textbf{Layer representation.} We represent makeup as a layer $({\boldsymbol{\alpha}},{\mathbf{F}})$ of coverage and pigment color in a face's coordinate frame, composited by Eq.~\eqref{eq:layer}, so that editing reduces to closed-form operations. Soft part weights ${\boldsymbol{\pi}}$ select lips, eyes, brows, face paint or forehead by scaling their coverage; strength scales ${\boldsymbol{\alpha}}$, recoloring changes ${\mathbf{F}}$, moving or rescaling warps the layer, and stacking and blending use alpha-over compositing and interpolation (Appendix~\ref{app:layerops}); each edit stays local to the selected part. Because rendering returns the same representation $(\tilde{\boldsymbol{\alpha}},{\mathbf{F}}')$ on the source face ${\mathbf{T}}$, these operations remain available afterwards, while cached reference layers let designs be replaced or their layer order exchanged without re-extracting makeup.

\noindent\textbf{Shared diffusion backbone.} Decomposition and rendering share one frozen MMDiT~\cite{sd3} $\theta$, FLUX.2-klein~\cite{flux2}, so both draw on the same generative prior; each adds its own LoRA~\cite{hu2022lora}, LoRA$_{\text{dec}}$ or LoRA$_{\text{ren}}$. Trained with rectified flow~\cite{liu2023flow,lipman2023flowmatching}, it predicts the velocity ${\mathbf{v}}_\theta({\mathbf{x}}_t,t,{\mathbf{c}})\approx{\boldsymbol{\epsilon}}-{\mathbf{x}}_0$ of latents ${\mathbf{x}}_t=(1-t){\mathbf{x}}_0+t{\boldsymbol{\epsilon}}$, with conditions ${\mathbf{c}}$ given as image tokens attending jointly with the target~\cite{ominicontrol,kontext}; ${\mathbf{z}}_{\mathbf{X}}$ is the latent of an image ${\mathbf{X}}$. We predict the decomposition canvas $[{\mathbf{B}}\,|\,{\mathbf{W}}\,|\,{\mathbf{K}}]$ from noise conditioned on the reference. For harmonization and rendering, we start from the noised composite ${\mathbf{x}}_\sigma=(1-\sigma){\mathbf{z}}_{\mathbf{X}}+\sigma{\boldsymbol{\epsilon}}$, so the input provides an appearance anchor~\cite{sdedit,art2026}. Zero-initialized encoder--decoder skips~\cite{pix2pixturbo,controlnet} carry input detail, and a masked residual confines the decoded change to an explicit editable region. 

\vspace{-1.5mm}
\input{sections/4b-gpt-tab}\subsection{One-Step Layer Decomposition}
\label{sec:decomposer}
\noindent\textbf{Triangulation matting for makeup layers.} One background gives three equations per pixel for four unknowns, so faint makeup cannot be told from pigment near the background color; triangulation matting~\cite{smithblinn} adds a second background (Figure~\ref{fig:trimat}). We therefore predict the makeup composited over white and black backgrounds:
\begin{equation}
{\mathbf{W}}={\boldsymbol{\alpha}}\odot{\mathbf{F}}+(1-{\boldsymbol{\alpha}})\odot\mathbf 1,\qquad {\mathbf{K}}={\boldsymbol{\alpha}}\odot{\mathbf{F}},
\end{equation}
and place both panels beside the bare face in a single canvas $[{\mathbf{B}}\,|\,{\mathbf{W}}\,|\,{\mathbf{K}}]$, which expresses layer recovery entirely in the pretrained RGB output space, without adding an alpha channel~\cite{layerdiffuse,qwenlayered}, so the pretrained VAE and prior carry over.

\begin{figure}[t]
\centering
\includegraphics[width=\linewidth]{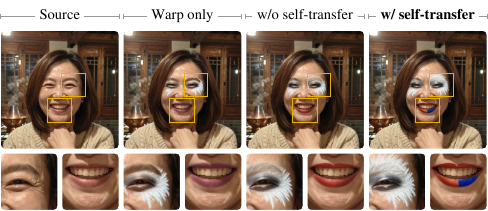}
\caption{\textbf{Effect of self-transfer training.} Self-transfer corrupts the paste with real alignment and resampling errors, so the DiT cannot copy its input and must recover details from the reference-layer tokens. This strengthens its representation of fine makeup details: without it, the strokes soften and the two-tone lip collapses into a uniform red; with it, both are restored.}\vspace{-2mm}
\label{fig:selftransfer}
\end{figure}
\noindent\textbf{Training.} The decomposer $D$ (LoRA$_{\text{dec}}$) maps a reference ${\mathbf{I}}_{\text{r}}$ and a fixed instruction to the whole canvas in one step (Appendix~\ref{app:onecanvas}). As the output is tightly constrained by the reference, we directly supervise the one-step estimate ${\mathbf{x}}_0={\boldsymbol{\epsilon}}-{\mathbf{v}}_\theta({\boldsymbol{\epsilon}},1,{\mathbf{c}})$ with decoded-image losses ($\mathcal L_{\text{canvas}}$), following pix2pix-Turbo~\cite{pix2pixturbo}, so that one step gives a sharp, pixel-accurate canvas without distilling a sampler~\cite{instaflow,add}.
Supervising the canvas alone, however, does not force the layers to explain the photograph, and on real photographs the recovered colors wash out (Figure~\ref{fig:harmcycle}). We therefore add the inverse task, harmonization $H$, which shares LoRA$_{\text{dec}}$ and is selected by another instruction (Figure~\ref{fig:twotasks})~\cite{cyclegan,pix2pixturbo,cycledecomp2026}. For a LayerBeauty item, we perturb its clean layer and paste it onto its bare face ${\mathbf{B}}$, giving ${\mathbf{C}}_{\text{H}}$; $H$ restores the photograph ${\mathbf{I}}$ locally, $\hat {\mathbf{I}}_{\text{H}}={\mathbf{C}}_{\text{H}}+{\boldsymbol{\Omega}}_{\text{H}}\odot(H({\mathbf{C}}_{\text{H}},{\mathbf{c}})-{\mathbf{C}}_{\text{H}})$, conditioned on coverage, part and hair maps ${\mathbf{c}}$, where ${\boldsymbol{\Omega}}_{\text{H}}$ covers the dilated union of the clean and perturbed layers. The two tasks are chained in a cycle: $D$ must recover the \emph{clean} canvas from $H$'s output, so $H$ learns to remove the perturbation and $D$ to produce layers that recomposite the photograph, while an identity term keeps harmonizing a bare face a no-op~\cite{cyclegan}. In total (Figure~\ref{fig:method}),
\begin{equation}
\begin{aligned}
\mathcal L_{\text{decompose}}={}&\mathcal L_{\text{canvas}}+\lambda_{\text{cycle}}\mathcal L_{\text{cycle}}\\
&+\mathcal L_{\text{harmonize}}+\lambda_{\text{identity}}\mathcal L_{\text{identity}},
\end{aligned}
\label{eq:decompose}
\end{equation}
where paired terms add LPIPS~\cite{lpips}, ${\mathbf{c}}_\varnothing$ denotes an empty layer, and Appendix~\ref{app:impl} lists the weights.

\noindent\textbf{Inference.} We run $D$ once per reference; harmonization is used only in training. Subtracting the panels cancels pigment color and leaves uncovered background, giving ${\boldsymbol{\alpha}}=1-\mathrm{mean}_{\text{rgb}}({\mathbf{W}}-{\mathbf{K}})$ and ${\mathbf{F}}={\mathbf{K}}/{\boldsymbol{\alpha}}$ wherever ${\boldsymbol{\alpha}}>0$. As ${\mathbf{W}}$ and ${\mathbf{K}}$ are predicted separately, we clip ${\boldsymbol{\alpha}}$ and ${\mathbf{F}}$ to $[0,1]$, which alters few pixels and keeps re-compositing far below the decomposition error (Appendix~\ref{app:validity}). We keep ${\mathbf{K}}={\boldsymbol{\alpha}}\odot{\mathbf{F}}$ as premultiplied color~\cite{porterduff}, so warping does not mix the pigment of transparent pixels into visible boundaries, and cache the layers for reuse across faces.

\input{sections/4b-gpt-fig}\vspace{-1.5mm}
\subsection{One-Step In-Context Rendering}
\label{sec:renderer}
\noindent\textbf{Alignment and composition.} We align each recovered layer to the source ${\mathbf{T}}$ by jointly warping coverage and premultiplied color over MediaPipe landmarks~\cite{mediapipe,kartynnik2019facemesh}, extended across the forehead, fit lips and brows to the source and restrict coverage to paintable skin (Appendix~\ref{app:protocol}). Since all references share this representation, selected regions of several references are composed before a single render. Selection must respect the design, as a face-parsing boundary~\cite{segformer} can cut through paint spanning eyes, brows and forehead, so we split a layer along a minimum-cost coverage seam found by dynamic programming~\cite{seamcarving}, choosing upper/lower or left/right zones by which crosses less makeup; finer parts serve designs confined to one part. Selected weights ${\mathbf{s}}_n$ scale the coverage of each layer $({\boldsymbol{\alpha}}^{(n)},{\mathbf{F}}^{(n)})$, and repeated alpha-over compositing (Eq.~\eqref{eq:layer}) gives the composite ${\mathbf{C}}$ on ${\mathbf{T}}$. Warping alone looks pasted on, so we render ${\mathbf{C}}$ locally to fit the makeup to the skin and repair artifacts such as broken strokes.

\noindent\textbf{Renderer.} LoRA$_{\text{ren}}$ denoises the noised paste ${\mathbf{C}}$~\cite{sdedit}. As in in-context conditioning~\cite{ominicontrol,easycontrol,kontext}, all conditions become image tokens concatenated with the target tokens, so full self-attention relates each target pixel to the intended makeup and the receiving face. The full warped reference layer over gray retains strokes removed by rules, eye and lip crops supply texture, shape maps, normals~\cite{kartynnik2019facemesh} and hair/glasses masks~\cite{celebamaskhq,faceparsing_hf} describe alignment, geometry and occlusion, and part captions set finish. The renderer predicts a decoded image ${\mathbf{G}}$ that contributes only through the masked residual $\hat{\mathbf{I}}={\mathbf{T}}+{\boldsymbol{\Omega}}\odot({\mathbf{G}}-{\mathbf{T}})$, where the editable region ${\boldsymbol{\Omega}}$ is computed by morphological operations on the warped design and pasted coverage, with design-aware protection for eyes, mouth, hair, glasses and unrequested parts (Eq.~\eqref{eq:omega}), so source pixels outside it stay unchanged, preserving identity.

\noindent\textbf{Training.} Without paired cross-identity targets, we build supervision from explicit layers: we send an item's layer to another identity's bare face ${\mathbf{B}}'$ and back to its own ${\mathbf{B}}$ using the transfer rules. This round trip introduces alignment and resampling errors while retaining ${\mathbf{I}}$ as the target and leaving its pixels unchanged outside the design, so the renderer learns to restore the design from the full reference layer (Figure~\ref{fig:selftransfer}). We also train on decompositions of real photographs, initializing LoRA$_{\text{ren}}$ from the decomposer before occluder fine-tuning, with
\begingroup\setlength{\abovedisplayskip}{2pt}\setlength{\belowdisplayskip}{2pt}
\begin{equation}
\begin{aligned}
\mathcal L_{\text{render}}={}&\mathcal L_{\text{FM}}+\lambda_{\text{L1}}\big\lVert{\boldsymbol{\Omega}}\odot(\hat{\mathbf{I}}-{\mathbf{I}})\big\rVert_1\\
&+\lambda_{\text{LPIPS}}\big\lVert{\boldsymbol{\Omega}}\odot\mathrm{LPIPS}(\hat{\mathbf{I}},{\mathbf{I}})\big\rVert_1+\mathcal L_{\text{aux}},
\end{aligned}
\label{eq:render}
\end{equation}
\endgroup
where $\mathcal L_{\text{FM}}$ is a token-weighted flow-matching loss~\cite{lipman2023flowmatching} on a straight path from the target latent ${\mathbf{z}}_{\mathbf{I}}$ to the noised paste ${\mathbf{x}}_\sigma$ (Appendix~\ref{app:impl}); sampling the endpoint $t=\sigma$ trains the one-step render directly. Its token weights emphasize the editable region and makeup changes, and we draw most samples at $\sigma=0.7$ on the four-step time grid of Table~\ref{tab:hparams}. The image losses supervise the decoded one-step estimate ${\mathbf{x}}_t-t\,{\mathbf{v}}_\theta$ after the masked residual; Appendix~\ref{app:impl} details the conditioning inputs, numerical handling and auxiliary losses $\mathcal L_{\text{aux}}$, which preserve skin, structure and color.

\noindent\textbf{Inference.} We render each composition once, in one Euler step from $\sigma=0.65$. To retain direct control after rendering, we express $\hat{\mathbf{I}}$ as a layer over ${\mathbf{T}}$: $\tilde{\boldsymbol{\alpha}}=\max({\boldsymbol{\alpha}}_{\text{d}},{\boldsymbol{\alpha}},{\boldsymbol{\alpha}}_{\hat{\mathbf{I}}})$ combines warped design coverage on paintable skin, pasted coverage and the minimum coverage needed to explain the rendered image, and Eq.~\eqref{eq:layer} gives ${\mathbf{F}}'=\big(\hat{\mathbf{I}}-(1-\tilde{\boldsymbol{\alpha}})\odot{\mathbf{T}}\big)/\tilde{\boldsymbol{\alpha}}$ where $\tilde{\boldsymbol{\alpha}}>\tau$. The result $(\tilde{\boldsymbol{\alpha}},{\mathbf{F}}')$ supports strength, color and blending edits without another network evaluation, giving persistent editability across repeated revisions.

%% file: sections/4b-gpt-tab.tex
\providecommand{\pend}{\textcolor{red}{?}}%
\begin{table*}[t]
\centering
\vspace{-7mm}
\caption{\textbf{Quantitative comparison with state-of-the-art methods.} We shade the top three per column; time includes decomposition.}
\label{tab:compare}
\scriptsize\setlength{\tabcolsep}{3pt}\renewcommand{\arraystretch}{0.95}
\resizebox{\linewidth}{!}{%
\begin{tabular}{@{}lrrrrrrrr@{\hspace{6pt}}rrrr@{\hspace{6pt}}rrrr@{}}
\toprule
 & \multicolumn{8}{c}{Multi-reference hybrid looks} & \multicolumn{4}{c}{Single-reference artistic} & \multicolumn{4}{c}{Single-reference pattern}\\
 & \multicolumn{8}{c}{LayerBeauty-CreativeScene test (100 looks, 3 references each, $1024^2$)} & \multicolumn{4}{c}{MF2K test (32 pairs, $512^2$)} & \multicolumn{4}{c}{CPM-Real (190 pairs, $512^2$)}\\
\cmidrule(lr){2-9}\cmidrule(lr){10-13}\cmidrule(l){14-17}
Method & ID\up & L2-M\dn & DINO$_{\text{part}}$\up & CLIP$_{\text{part}}$\up & \begin{tabular}[b]{@{}r@{}}Leak\\$\Delta E$\dn\end{tabular} & \begin{tabular}[b]{@{}r@{}}Color\\$\Delta E_{\text{part}}$\dn\end{tabular} & Assign\up & \begin{tabular}[b]{@{}r@{}}Time\\s\dn\end{tabular} & ID\up & DINO\up & CLIP\up & L2-M\dn & ID\up & DINO\up & CLIP\up & L2-M\dn\\\midrule
\textcolor{refgray}{Source} & \textcolor{refgray}{1.00} & \textcolor{refgray}{0.0} & \textcolor{refgray}{0.71} & \textcolor{refgray}{0.85} & \textcolor{refgray}{0.0} & \textcolor{refgray}{25.4} & \textcolor{refgray}{--} & \textcolor{refgray}{--} & \textcolor{refgray}{1.000} & \textcolor{refgray}{0.424} & \textcolor{refgray}{0.587} & \textcolor{refgray}{0.0} & \textcolor{refgray}{1.000} & \textcolor{refgray}{0.345} & \textcolor{refgray}{0.538} & \textcolor{refgray}{0.0}\\\midrule
EleGANt & \cellcolor{top2}0.73 & 170.8 & 0.73 & 0.87 & 16.7 & 16.9 & 0.51 & \cellcolor{top1}1.0 & \cellcolor{top3}0.748 & 0.420 & 0.627 & 412.1 & 0.747 & 0.375 & 0.607 & 409.3\\
MAD & 0.62 & \cellcolor{top1}10.3 & 0.73 & 0.87 & \cellcolor{top2}12.3 & 14.9 & 0.55 & 121.8 & 0.636 & 0.474 & 0.657 & \cellcolor{top3}11.0 & 0.663 & 0.405 & 0.617 & \cellcolor{top2}10.5\\
Stable-Makeup & 0.28 & 472.2 & 0.71 & \cellcolor{top3}0.89 & 23.7 & 17.5 & 0.56 & 15.4 & 0.329 & \cellcolor{top2}0.611 & \cellcolor{top1}0.784 & 131.6 & 0.330 & 0.597 & \cellcolor{top2}0.756 & 137.0\\
FLUX-Makeup & 0.62 & 248.4 & 0.74 & 0.88 & 20.3 & 17.4 & 0.50 & 97.4 & 0.701 & 0.519 & 0.700 & 109.4 & \cellcolor{top3}0.801 & 0.508 & 0.667 & 128.1\\
Qwen-Image-Edit & 0.30 & 7794.3 & \cellcolor{top1}0.81 & \cellcolor{top1}0.91 & 34.9 & 18.0 & 0.33 & 261.3 & 0.484 & \cellcolor{top1}0.624 & \cellcolor{top2}0.752 & 1341.5 & 0.445 & \cellcolor{top1}0.682 & \cellcolor{top1}0.761 & 1220.5\\
FLUX.2-klein & 0.43 & 2172.5 & 0.76 & \cellcolor{top3}0.89 & 22.0 & 20.5 & 0.27 & 7.0 & 0.348 & 0.549 & 0.720 & 977.2 & 0.388 & \cellcolor{top3}0.628 & 0.726 & 72.7\\
FRAM & 0.54 & 55.9 & \cellcolor{top3}0.77 & \cellcolor{top2}0.90 & \cellcolor{top3}14.7 & 13.7 & \cellcolor{top3}0.67 & 7.2 & 0.602 & 0.555 & \cellcolor{top3}0.739 & 68.3 & 0.595 & 0.582 & 0.720 & 66.8\\
MagicMakeup & 0.63 & 109.7 & \cellcolor{top3}0.77 & 0.88 & 22.3 & \cellcolor{top3}13.1 & 0.61 & 245.1 & 0.677 & 0.574 & 0.727 & 21.4 & 0.670 & 0.588 & 0.706 & 26.4\\
ART & 0.59 & 227.4 & 0.69 & 0.87 & 20.5 & 17.9 & 0.39 & 155.3 & 0.669 & \cellcolor{top3}0.582 & 0.729 & 51.5 & 0.525 & \cellcolor{top2}0.658 & \cellcolor{top3}0.746 & 79.4\\
\rowcolor{oursrow}\method{} (warp only) & \cellcolor{top1}0.79 & \cellcolor{top2}11.8 & 0.76 & \cellcolor{top3}0.89 & \cellcolor{top1}6.8 & \cellcolor{top2}9.5 & \cellcolor{top2}0.68 & \cellcolor{top2}2.6 & \cellcolor{top1}0.924 & 0.486 & 0.661 & \cellcolor{top1}0.0 & \cellcolor{top1}0.894 & 0.542 & 0.650 & \cellcolor{top1}6.9\\
\rowcolor{oursrow}\method{} (+ rendering) & \cellcolor{top3}0.71 & \cellcolor{top3}15.4 & \cellcolor{top2}0.80 & \cellcolor{top2}0.90 & 15.6 & \cellcolor{top1}8.3 & \cellcolor{top1}0.78 & \cellcolor{top3}3.8 & \cellcolor{top2}0.866 & 0.509 & 0.681 & \cellcolor{top2}0.6 & \cellcolor{top2}0.829 & 0.571 & 0.674 & \cellcolor{top3}11.7\\\bottomrule
\end{tabular}}\vspace{-4mm}
\end{table*}

%% file: sections/4b-gpt-fig.tex
\begin{figure*}[t]
\centering
\vspace{1.5mm}%
\includegraphics{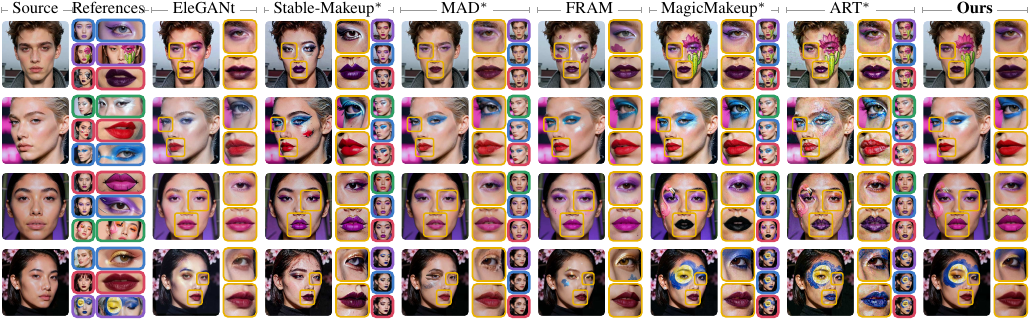}
\caption{\textbf{Comparison of multi-reference looks.} Frame colors mark the part taken from each reference; $^\ast$ marks single-reference methods.}\vspace{-2mm}
\label{fig:gpt}\vspace{-2mm}
\end{figure*}

%% file: sections/4-experiments.tex
\vspace{-2mm}
\section{Experiments}
\label{sec:exp}

\begin{figure*}[t]\vspace{-\dbltextfloatsep}\vspace{-5mm}\end{figure*}%

\noindent\textbf{Evaluation setup.} We evaluate transfer on three benchmarks (Table~\ref{tab:compare}): hybrid looks that combine three references on one source face, artistic MF2K~\cite{art2026} and pattern-based CPM-Real~\cite{cpm}. We compare with makeup transfer methods~\cite{elegant,stablemakeup,fluxmakeup,gao2026fram,magicmakeup,art2026,mad} and general image editors~\cite{qwenimageedit2511,flux2}, with official settings (Appendix~\ref{app:hybrid}), select ours on identity-disjoint validation sets (Appendix~\ref{app:protocol}), and evaluate decomposition on items with known layers (Table~\ref{tab:decomp_baselines}).

\noindent\textbf{Metrics.} Face-ID (ID) and the error outside the face (L2-M) measure source preservation~\cite{magicmakeup,adaface}. An unchanged source scores perfectly on both, and whole-image DINO and CLIP~\cite{dino,clip} also reward copying the reference's face, so we compare each requested part only with its own reference: part-wise DINO and CLIP measure its appearance, Color $\Delta E_{\text{part}}$ its CIEDE2000 color error, and Assign the fraction of parts that follow their designated reference. Leak $\Delta E$ measures the CIE76 color change of unrequested parts.

\vspace{-1.5mm}
\subsection{Makeup Transfer and Compositional Control}
\label{sec:comparison}

\input{sections/4b-gpt}

\begin{figure}[t]
\centering
\includegraphics[width=\linewidth]{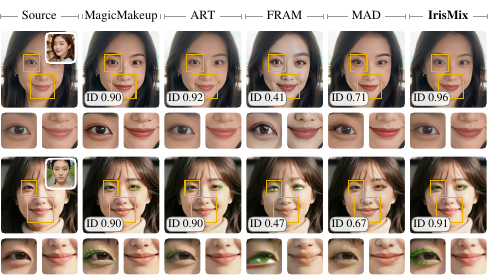}\vspace{-1mm}
\caption{\textbf{Comparison of single-reference looks.} \method{} keeps the source's identity and the reference makeup while adapting it to the scene lighting and appearance, ensuring photorealistic results.}
\label{fig:daily}\vspace{-2mm}
\end{figure}
\noindent\textbf{Results on traditional benchmarks.} On artistic MF2K~\cite{art2026} and pattern-based CPM-Real~\cite{cpm}, compositing keeps the person most recognizable (Table~\ref{tab:compare}): the layer changes only the pixels it covers. Regenerative methods score higher whole-image DINO and CLIP, partly by copying the reference's face, and lose identity: in Figure~\ref{fig:daily}, FRAM reshapes the face, MAD and ART transfer the liner faintly, and MagicMakeup overdraws it, whereas \method{} keeps the source's identity, skin texture and lighting.

\begin{table}[t]
\centering\setlength{\aboverulesep}{0.2ex}\setlength{\belowrulesep}{0.3ex}\setlength{\abovecaptionskip}{2pt}\setlength{\belowcaptionskip}{2pt}
\caption{\textbf{Comparison with layer decomposition methods.} Composite pairs warp CreativeScene layers onto independent bare faces (Appendix~\ref{app:protocol}). For layered generators, we use the generated layer subset with the lowest white plus black L1 error. Errors are $\times100$; runtime is in seconds per image.}
\label{tab:decomp_baselines}
\scriptsize\setlength{\tabcolsep}{1.7pt}\renewcommand{\arraystretch}{0.85}
\resizebox{\linewidth}{!}{%
\begin{tabular}{@{}lrrrrrrrrr@{}}
\toprule
 & \multicolumn{4}{c}{DailyScene test} & \multicolumn{4}{c}{Composite test} & Time\\
\cmidrule(lr){2-5}\cmidrule(lr){6-9}
Method & L1\dn & LPIPS\dn & $\mathbf W$\dn & $\mathbf K$\dn & L1\dn & LPIPS\dn & $\mathbf W$\dn & $\mathbf K$\dn & s\dn\\\midrule
FLUX.2-klein & 4.38 & 10.40 & 1.54 & 0.89 & 4.15 & 11.13 & 2.18 & 3.29 & 4.4\\
MAD ($256^2$) & 1.91 & 4.48 & 0.95 & 1.30 & 4.61 & 12.52 & 3.81 & 5.14 & 116.7\\\midrule
Qwen-Layered & 3.84 & 7.86 & 3.38 & 1.57 & 4.74 & 11.58 & 4.36 & 2.17 & 111.9\\
Stable-Layers & 10.59 & 15.24 & 11.56 & 6.67 & 8.09 & 13.53 & 5.22 & 3.18 & 164.4\\\midrule
\rowcolor{oursrow}Ours & \textbf{1.62} & \textbf{1.71} & \textbf{0.35} & \textbf{0.20} & \textbf{1.88} & \textbf{3.08} & \textbf{0.89} & \textbf{0.73} & \textbf{0.45}\\\bottomrule
\end{tabular}}\vspace{-2mm}
\end{table}
\begin{figure*}[t]
\centering
\vspace{-8mm}
\includegraphics{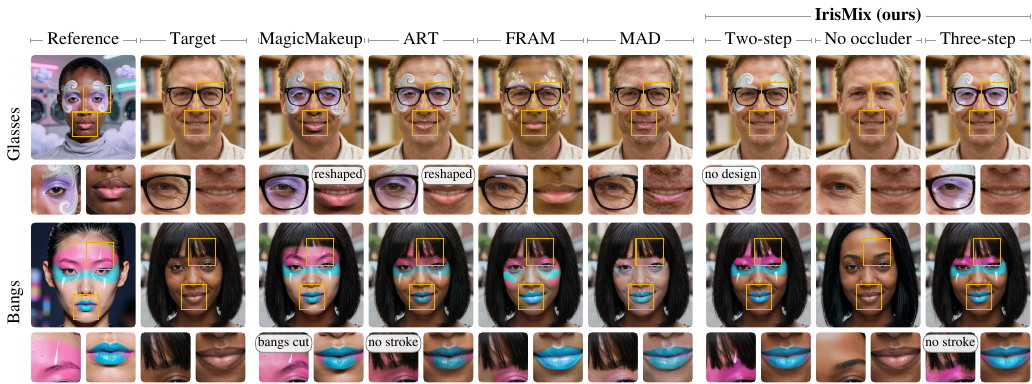}
\caption{\textbf{Comparison of occlusion-aware transfer across ethnicities.} IrisMix's semi-transparent layers keep the source's skin tone, lip shape and hair without loss of identity, and removing, transferring and restoring the occluder continues the design behind lenses and under strands, whereas the baselines carry the reference's skin tone into the source, reshape its lips, cut the bangs or transfer little design.}
\label{fig:occluder}\vspace{-5mm}
\end{figure*}
\vspace{-1.5mm}
\subsection{Analysis and Extended Results}
\label{sec:analysis}
\noindent\textbf{Decomposition baselines.} Table~\ref{tab:decomp_baselines} and Figure~\ref{fig:decomp} show that our decomposer has the lowest errors, even against best-matching generator layers~\cite{qwenlayered,stablelayers}, with a single network evaluation. The white and black backgrounds separate faint makeup and skin-colored pigment, whereas generic layer generators split images into objects, and removal-based MAD degrades on heavy designs.

\input{sections/4c-renderer}

\vspace{-1.5mm}
\subsection{Ablation Studies}
\label{sec:ablation}
\begin{table}[t]
\centering
\caption{\textbf{Effect of decomposer training data and objectives.} We reduce bare-face error with diverse data and cycle training, then improve occluder separation by fine-tuning. We report validation L1 errors ($\times100$); $\mathbf W$ and $\mathbf K$ denote white- and black-background layer composites, and Table~\ref{tab:decomposer_data} lists the decomposer training sources.}
\label{tab:cycle}
\scriptsize\setlength{\tabcolsep}{3pt}
\resizebox{\linewidth}{!}{%
\begin{tabular}{@{}lrrrrrrr@{}}
\toprule
 & \multicolumn{3}{c}{Decompose} & \multicolumn{2}{c}{Harmonize} & Cycle & Identity\\
\cmidrule(lr){2-4}\cmidrule(lr){5-6}\cmidrule(lr){7-7}\cmidrule(l){8-8}
Decomposer training & bare\dn & $\mathbf W$\dn & $\mathbf K$\dn & L1\dn & LPIPS\dn & L1\dn & L1\dn\\\midrule
\textcolor{refgray}{Paste} & \textcolor{refgray}{--} & \textcolor{refgray}{--} & \textcolor{refgray}{--} & \textcolor{refgray}{2.37} & \textcolor{refgray}{7.84} & \textcolor{refgray}{--} & \textcolor{refgray}{--}\\\midrule
Decompose only, CreativeScene data & 5.64 & 1.80 & 1.22 & -- & -- & -- & --\\
\quad + all training sources & 2.79 & 0.85 & 0.70 & 2.59 & 7.96 & 2.55 & 3.17\\\midrule
\quad + harmonize and cycle & 2.05 & 0.88 & \textbf{0.69} & 2.37 & 7.66 & 2.24 & \textbf{1.11}\\
\quad + stronger perturbations & 2.18 & 0.86 & 0.70 & 2.29 & 7.40 & 2.24 & 1.16\\
\quad + Harmonize L1 in $\boldsymbol\Omega_{\text{H}}$ & 2.02 & 0.87 & 0.73 & 2.15 & 7.30 & 2.09 & 1.20\\
\rowcolor{oursrow}\quad + occluder fine-tuning (ours) & \textbf{1.89} & \textbf{0.79} & \textbf{0.69} & \textbf{2.10} & \textbf{7.23} & \textbf{2.01} & 1.12\\\bottomrule
\end{tabular}}\vspace{-2mm}
\end{table}

\begin{table}[t]
\centering\setlength{\aboverulesep}{0.2ex}\setlength{\belowrulesep}{0.3ex}\setlength{\abovecaptionskip}{2pt}\setlength{\belowcaptionskip}{2pt}
\caption{\textbf{Effect of renderer components.} The reference layer and self-transfer trade identity for fidelity; the masked residual and one step improve both, showing that the reference supplies the design while the masked residual and the noised paste keep identity.}
\label{tab:s2abl}
\scriptsize\setlength{\tabcolsep}{3pt}\renewcommand{\arraystretch}{0.8}
\resizebox{\linewidth}{!}{\input{figures/tables/model_main}}\vspace{-2mm}
\end{table}

\noindent\textbf{Decomposer components.} The decomposer trains on CreativeScene, DailyScene and Composite pairs, real FFHQ-Makeup photographs (bare panel only) and, for fine-tuning, glasses and bangs pairs (Table~\ref{tab:decomposer_data}). As shown in Table~\ref{tab:cycle}, diverse data lowers bare-face error the most, as makeup must be separated from many skin tones and lights; harmonization with cycle consistency, which requires the layers to recomposite the photograph, and occluder fine-tuning lower it further, especially around glasses, bangs and hair.

\noindent\textbf{Renderer components.} In Table~\ref{tab:s2abl}, without the full reference layer or self-transfer, the renderer mostly copies the paste: identity is higher, but strokes removed by the transfer rules are lost and colors drift, so we trade some identity for design fidelity. The masked residual protects the source at no cost in similarity, and one-step sampling from $\sigma=0.65$ keeps the noised paste as an anchor, improving identity and color over four steps at a quarter of the cost.

%% file: sections/4b-gpt.tex
\noindent\textbf{Results on LayerBeauty.} \emph{Quantitatively} (Table~\ref{tab:compare}), our composed layers keep the person most consistent with the source, change unrequested parts least and match the reference colors most closely; one-step rendering then follows each reference more faithfully at a small cost in identity. \emph{Qualitatively} (Figure~\ref{fig:gpt}), each baseline fails in its own way: multi-image editors repaint the whole face, single-reference methods overwrite earlier references when they regenerate their own outputs, and FRAM and EleGANt fade fine details and drop most face paint. Our method avoids these failures because each layer changes only the pixels it covers and all layers are rendered together once, so errors cannot accumulate between references.

%% file: sections/4c-renderer.tex
\noindent\textbf{In-context rendering.} Because the MMDiT attends jointly over the target, the source and the full reference layer, each rendered pixel draws on both the intended design and the source's geometry and lighting, which improves part similarity, assignment and color over warping (Figure~\ref{fig:renderer_cmp}, Table~\ref{tab:compare}); in exchange, it changes more of the source than warping but less than the baselines.

\noindent\textbf{Occlusion-aware transfer.} Because occluders are layers above the makeup, glasses can first be removed, the face then made up and rendered accordingly, and only the frames restored on top afterwards. As shown in Figure~\ref{fig:occluder}, the eye design then continues behind clear lenses, which two-step transfer loses, whereas MagicMakeup cuts the bangs and MAD transfers little of the design.

\noindent\textbf{Transfer across ethnicities.} Figure~\ref{fig:occluder} also crosses ethnicities, and every baseline carries the reference into the source: MagicMakeup, FRAM and MAD shift its skin tone and ART reshapes the lips. Our semi-transparent layer lets the source's own skin show through, keeping its tone.

\begin{figure}[t]
\centering
\includegraphics[width=\linewidth]{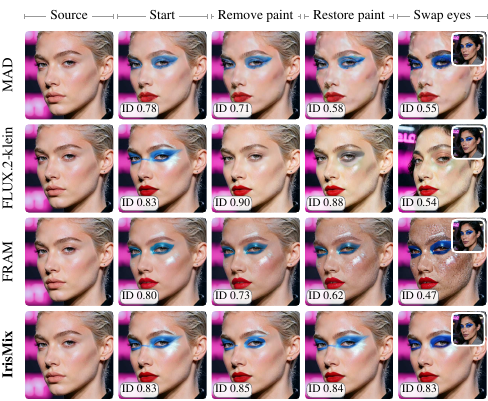}
\caption{\textbf{Comparison of multi-round editing.} Because each edit changes only an explicit layer, the background and the unedited parts of the face stay nearly unchanged across rounds.}\vspace{-2mm}
\label{fig:multiround}
\end{figure}
\noindent\textbf{Multi-round editing.} Figure~\ref{fig:multiround} removes the face paint, restores it and swaps the eyes step by step. With explicit layers, each edit acts directly on its target layer: removing and restoring are closed-form, and the swap recomposes cached layers and renders once on the source face, keeping the result natural. Baselines must express every edit through prompts or reference conditions and regenerate their previous output, so MAD, FLUX.2-klein and FRAM lose identity with each successive edit. Over seven edits on many looks, our method also drifts much less in never-edited regions than FLUX.2-klein, even when it receives the full history (Tables~\ref{tab:multiround} and~\ref{tab:multiround_hist}); Figures~\ref{fig:teaser} and~\ref{fig:apps} show more edits.

%% file: figures/tables/model_main.tex
\begin{tabular}{@{}lrrrr@{}}
\toprule
Renderer & ID\up & DINO$_{\text{part}}$\up & Color $\Delta E$\dn & L2-M\dn\\\midrule
4 steps, w/o full reference layer & 0.778 & 0.731 & 9.32 & 6.9\\
\quad + full reference layer & 0.666 & 0.758 & 8.72 & 7.7\\
\midrule
4 steps, w/o masked residual & 0.640 & 0.756 & 9.09 & 13.8\\
\quad + masked residual & 0.666 & 0.758 & 8.72 & 7.7\\
\rowcolor{oursrow}\quad + 1 step, $\sigma=0.65$ (ours) & 0.687 & 0.756 & 8.47 & 7.3\\
\midrule
Training w/o self-transfer & 0.743 & 0.743 & 9.07 & 6.5\\
\quad + self-transfer & 0.687 & 0.750 & 8.40 & 6.9\\
\bottomrule
\end{tabular}

%% file: sections/5-conclusion.tex
\section{Conclusion}
We present IrisMix, a compositional makeup transfer and editing framework based on layered diffusion, and LayerBeauty, a dataset of everyday and artistic looks with explicit makeup layers. By treating makeup as a layer of coverage and pigment color rather than as pixels of an output image, IrisMix makes transfer editable: one diffusion transformer with dual LoRAs decomposes references into layers by triangulation matting and renders composed layers to fit the source's geometry and illumination, each in a single step. Because the render is returned as a layer, selection, stacking, recoloring and blending stay closed-form, and further renders restart from cached layers instead of earlier outputs. Experiments show lower decomposition error, stronger identity preservation, references combined without overwriting one another, and far less drift over repeated edits, enabling interactive and personalized makeup design.

%% file: sections/A-analysis.tex
\section{Layer Editing and Decomposition Analysis}
\label{app:analysis}
\subsection{Layer Operations and Editing}
\label{app:layerops}
\begin{figure*}[t]
\centering
\includegraphics[width=0.98\textwidth]{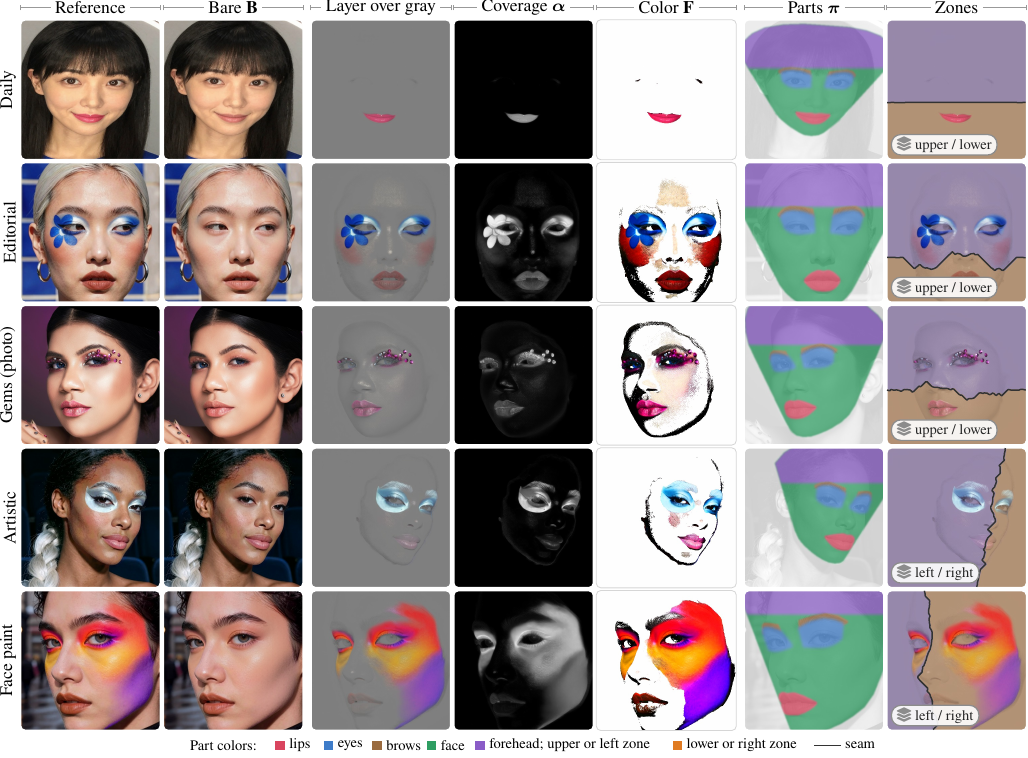}
\caption{\textbf{Layer anatomy.} One decomposition step turns each reference into a bare face $\mathbf{B}$ and a layer whose coverage ${\boldsymbol{\alpha}}$ and color $\mathbf{F}$ (shown where ${\boldsymbol{\alpha}}>0.05$) separate the design from the skin, from daily washes to heavy face paint. Part weights ${\boldsymbol{\pi}}$ select lips, eyes, brows, face paint and forehead, and zones split a layer along a minimum-cost seam through bare skin, so cuts avoid splitting a design mid-stroke. Third reference: MF2K photograph by Monika Kesharwani (Unsplash License).}
\label{fig:anatomy}
\end{figure*}

\begin{figure*}[t]
\centering
\includegraphics[width=0.90\textwidth]{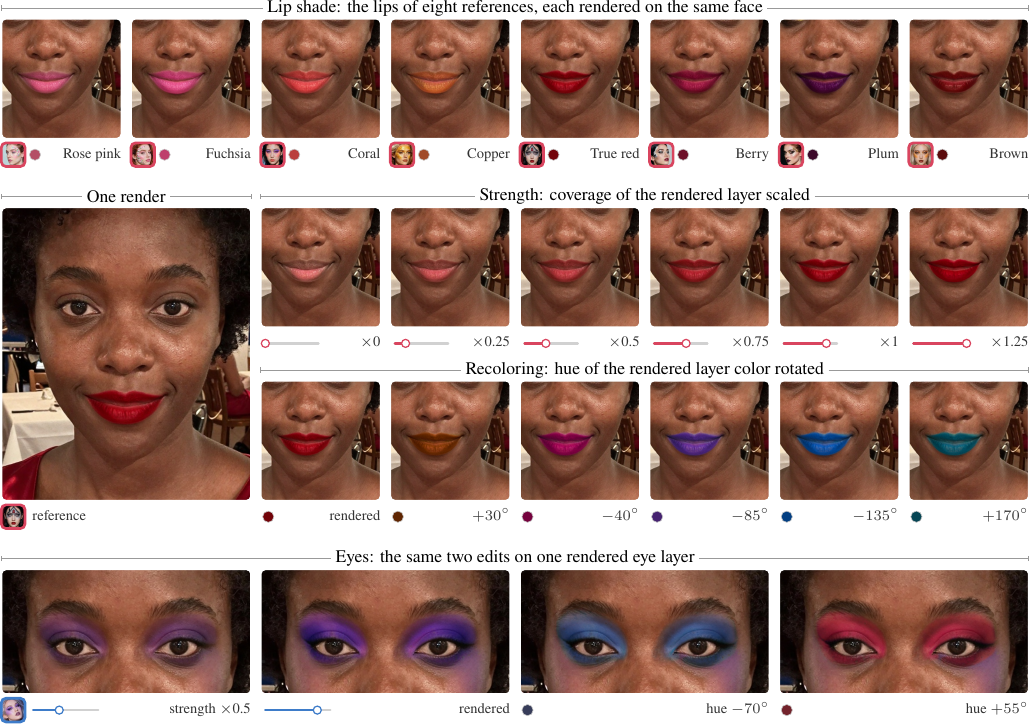}
\caption{\textbf{Fine control of a transferred layer.} Top: the lips of eight test references (red frame) rendered on one face by the reported renderer, with the median rendered lip color (dot). Middle and bottom: the layer $(\tilde{\boldsymbol{\alpha}},{\mathbf{F}}')$ of one render, with $\tilde{\boldsymbol{\alpha}}$ scaled (clipped at 1) or the CIELAB hue of ${\mathbf{F}}'$ rotated on lips or eyes at fixed lightness and chroma; DailyScene test source face; all nine references unseen.}
\label{fig:control}
\end{figure*}

\begin{figure*}[t]
\centering
\includegraphics[width=0.95\textwidth]{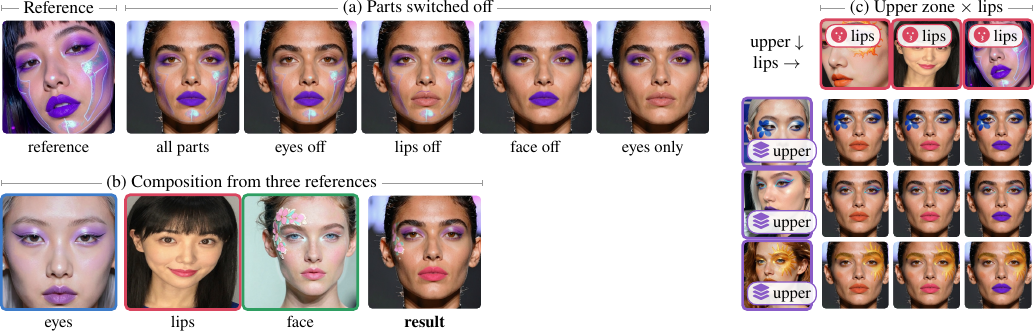}
\caption{\textbf{Editing and composition on one source.} Because each reference becomes an explicit layer, parts can be switched off (a), eyes, lips and face paint can come from three different references (b), and the upper zone of one reference can be paired with the lips of another (c). All tiles are warped and composited without rendering.}
\label{fig:edit}
\end{figure*}

\begin{figure*}[t]
\centering
\includegraphics[width=0.82\textwidth]{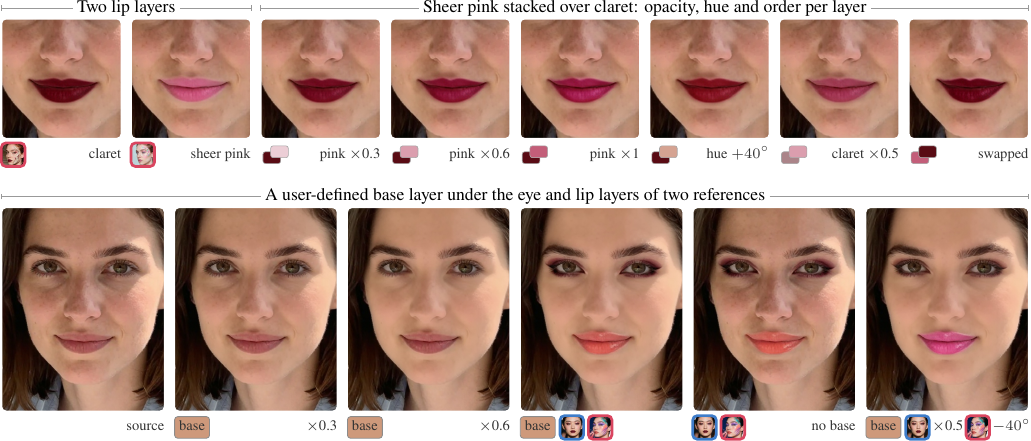}
\\[1mm]\includegraphics[width=0.82\textwidth]{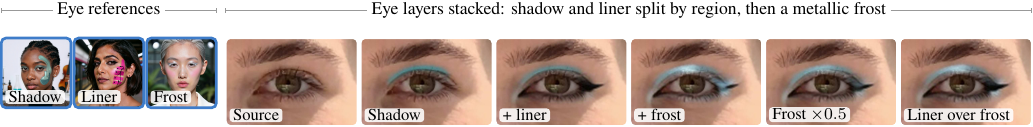}
\caption{\textbf{Layer stacking.} Each tile is one render of a stack composited bottom to top (layers shown under each tile) on a source face unseen during training. Top: the lips of two test references, claret and a sheer pink, stacked at strengths 0.3 to 1, recolored and swapped; pink over claret gives a berry shade, while swapped, the claret hides it. Middle: a user-defined base tinted from the source's skin tone, under the eyes and lips of two other references, which are then faded or recolored. Last row: eye layers from three references; a region rule splits shadow and liner, and only the top layer supplies the renderer's eye crops.}
\label{fig:stack}
\end{figure*}

\begin{figure*}[t]
\centering
\includegraphics[width=0.90\textwidth]{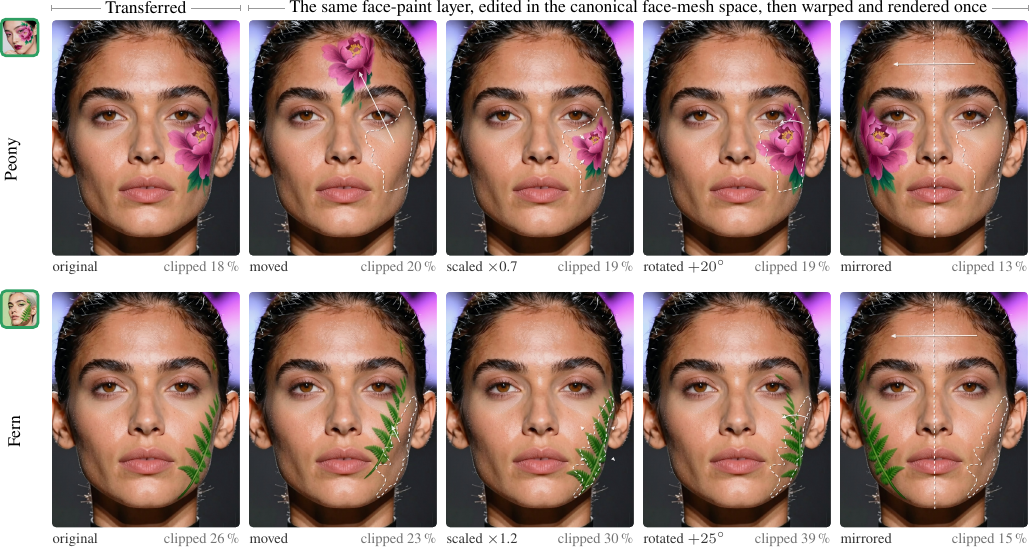}
\caption{\textbf{Design repositioning.} Because the face paint is an explicit layer, it can be moved, scaled, rotated or mirrored on a canonical face mesh before it is warped to a new face and rendered once, so the design keeps its shape wherever it lands on paintable skin; ``clipped'' gives the share of the warped design removed by the paintable-skin rules (eyes, brows, lips, hair and beyond the face contour), and dashed outlines mark the original position.}
\label{fig:reposition}
\end{figure*}

\begin{figure*}[t]
\centering
\includegraphics[width=0.95\textwidth]{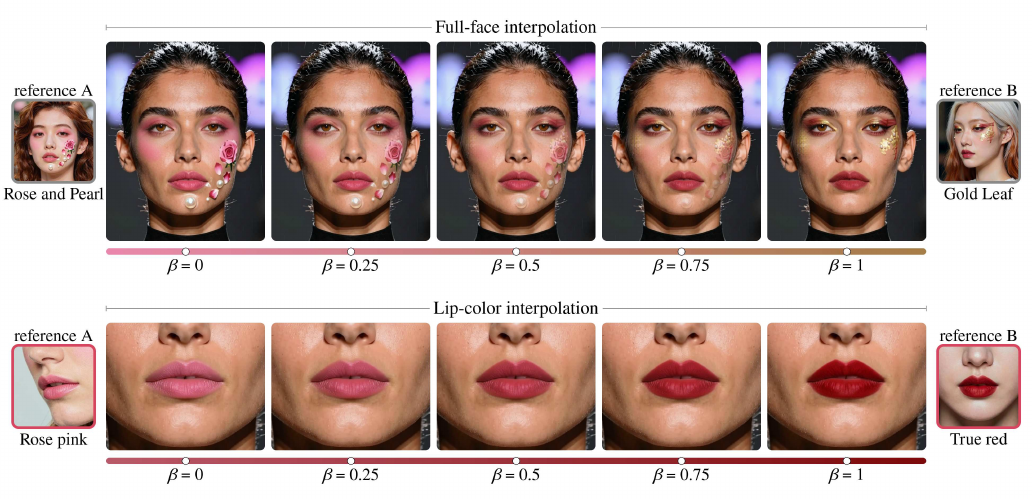}
\caption{\textbf{Look blending.} In-between frames interpolate the layers recovered from two renders on one unseen face ($\tilde{\boldsymbol{\alpha}}$ and premultiplied color linear in $\beta$, composited over the source). Top: whole looks; bottom: lipstick shades.}
\label{fig:blend}
\end{figure*}

For a layer (Figure~\ref{fig:anatomy}) on a fixed source $\mathbf T$, write $\mathbf K=\boldsymbol\alpha\odot\mathbf F$ for premultiplied color. The layer composites onto the source as $\mathbf K+(1-\boldsymbol\alpha)\odot\mathbf T$. The following operations act on the layer in closed form (Figures~\ref{fig:control}--\ref{fig:blend}).

\noindent\textbf{Selection and strength.} A spatial selection $\mathbf s\in[0,1]$ multiplies both coverage and premultiplied color: $(\boldsymbol\alpha,\mathbf K)\mapsto(\mathbf s\odot\boldsymbol\alpha,\mathbf s\odot\mathbf K)$. Strength $k$ uses $\boldsymbol\alpha_k=\min(1,k\boldsymbol\alpha)$ and $\mathbf K_k=\boldsymbol\alpha_k\odot\mathbf F$; setting $k=0$ removes the layer.

\noindent\textbf{Ordered stacking.} Placing layer 2 over layer 1 gives
\begin{equation}
\begin{aligned}
\boldsymbol\alpha^{(2\text{ over }1)}&=\boldsymbol\alpha^{(2)}+(1-\boldsymbol\alpha^{(2)})\odot\boldsymbol\alpha^{(1)},\\
\mathbf K^{(2\text{ over }1)}&=\mathbf K^{(2)}+(1-\boldsymbol\alpha^{(2)})\odot\mathbf K^{(1)}.
\end{aligned}
\end{equation}
Order matters: an opaque upper layer hides the lower one (Figure~\ref{fig:stack}). Cached component layers let their order be exchanged after a combined render without re-extracting makeup.

\noindent\textbf{Linear blending.} For two aligned layers on the same source and $\beta\in[0,1]$,
\begin{equation}
(\boldsymbol\alpha_\beta,\mathbf K_\beta)=(1-\beta)(\boldsymbol\alpha^{(1)},\mathbf K^{(1)})+\beta(\boldsymbol\alpha^{(2)},\mathbf K^{(2)}).
\end{equation}
Composited over $\mathbf T$, it cross-fades the endpoint images in appearance and also blends layers from two renders (Figure~\ref{fig:blend}).

\begin{figure*}[t]
\centering
\includegraphics[width=0.95\textwidth]{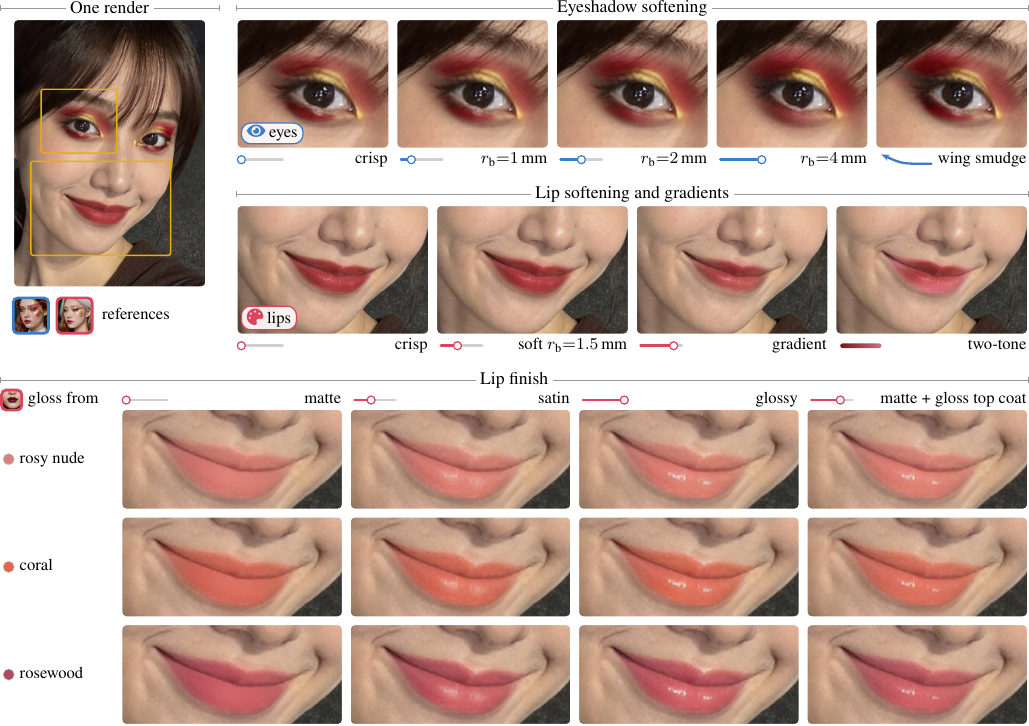}
\caption{\textbf{Softening and finish.} Because the makeup is an explicit layer, its softness and finish stay editable: each edited layer is rendered once, blurring by a radius $r_{\text{b}}$ (in millimeters, relative to the inter-ocular distance) softens edges, glossy highlights painted into the lip layer set the finish, a sheer glossy top coat stacks over a matte layer, and the matte tiles also damp the rendered highlights in closed form.}
\label{fig:finish}
\end{figure*}
\noindent\textbf{Softening and finish.} In Figure~\ref{fig:finish}, we blur the eye or lip layer, extend its color by normalized convolution, or drag, fade and shade it before one render. The layer has no specular term, so its painted highlights set the lip finish, and the highlight area grows from matte to satin to glossy.

\begin{figure*}[t]
\centering
\includegraphics[width=0.95\textwidth]{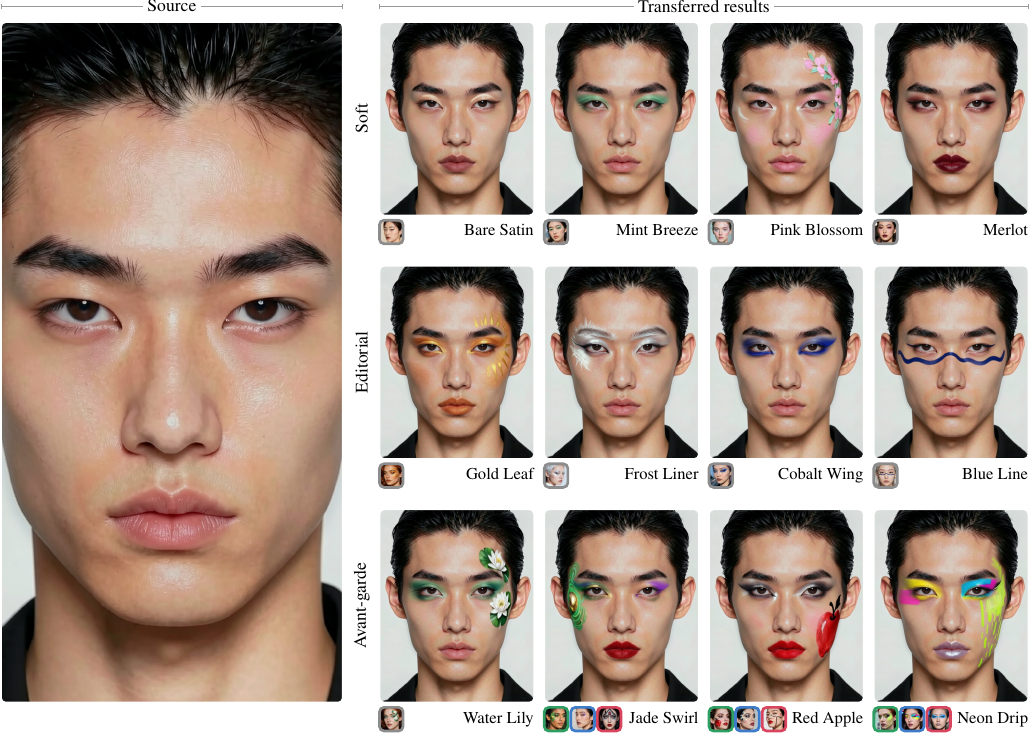}
\caption{\textbf{Explicit per-part control on one face.} Twelve looks, from soft daily makeup to avant-garde face painting, on one unseen face, from one reference (gray frame) or three (eyes blue, lips red, face paint green). All 18 references are unseen in training and used once.}
\label{fig:oneface}
\end{figure*}

\begin{figure*}[t]
\centering
\includegraphics{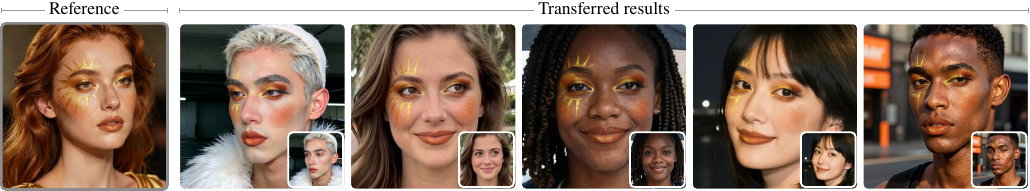}
\caption{\textbf{Transfer of one look to many faces.} The reference is decomposed once and its layer reused: the same gold design follows each new face's shape, skin tone and lighting, from men to women and from light to dark skin (the insets show the sources).}
\label{fig:onelook}
\end{figure*}

\begin{figure}[t]
\centering
\includegraphics[width=\linewidth]{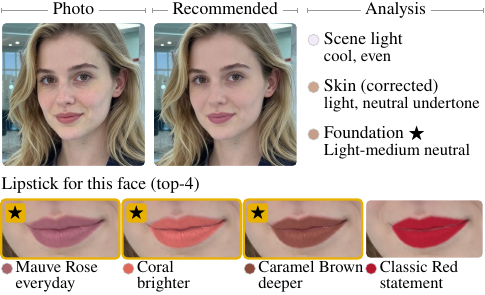}\vspace{-1mm}
\caption{\textbf{Simple personalized makeup recommendation.} Skin and lip colors are estimated under corrected lighting to rank foundations and select lipsticks; stars mark the recommended shades. Each shade re-composites only one layer in closed form and needs no network evaluation. Since face, makeup and light are separated, each suggestion is previewed on the user's own face in its own lighting.}\vspace{-2mm}
\label{fig:shades}
\end{figure}

\begin{figure}[t]
\centering
\includegraphics{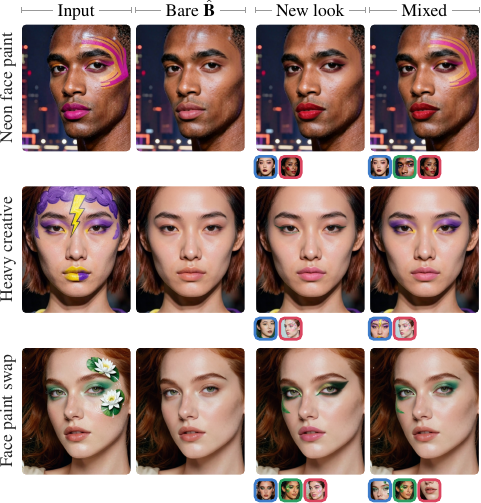}
\caption{\textbf{Re-make-up.} The decomposer strips the makeup of an already made-up photo, so parts of other references (the frame color marks the part) can be applied to the recovered bare face $\hat{\mathbf B}$ and rendered once. \emph{Mixed} keeps some of the input's own parts (thumbnails cut from the input). Inputs and references are unseen in training.}
\label{fig:remakeup}
\end{figure}

\begin{figure*}[t]
\centering
\includegraphics{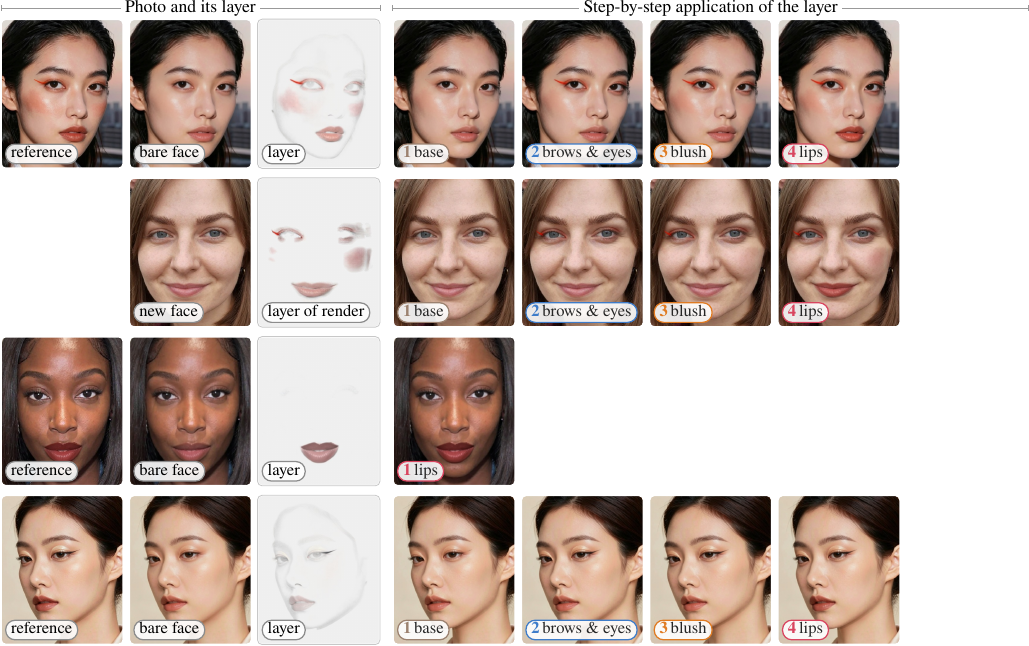}
\caption{\textbf{Step-by-step makeup from one photo.} Fixed rules on the part weights ${\boldsymbol{\pi}}$ split a made-up photo into base, brows and eyes, blush and lips, each step a closed-form composite; parts with almost no coverage are skipped (third row) and the blush is softened (Gaussian feathering of about 3\,mm, 60\,\% strength). Second row: the first look on a new face, split likewise from one render. Third and fourth rows: two further references.}
\label{fig:tutorial}
\end{figure*}

\noindent\textbf{Shade preview.} Because face, makeup and light are separated, cosmetic shades can be previewed on the user's own face in its own lighting (Figure~\ref{fig:shades}); a simple color heuristic suggests foundations and lipsticks, and switching shades re-composites only one layer in closed form. Figures~\ref{fig:oneface}--\ref{fig:tutorial} show further applications.

\subsection{Decomposition Analysis}
\label{app:steps}
\begin{figure}[t]
\centering
\includegraphics[width=\columnwidth]{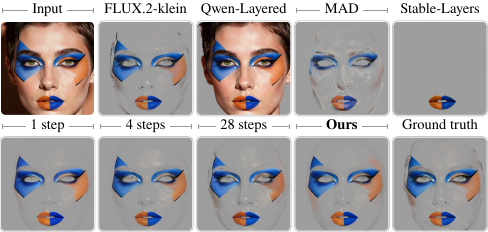}
\caption{\textbf{Decomposition quality.} Removal and multi-step sampling leave skin in the layer, and layered generators keep the whole person or miss the design; ours isolates the makeup.}
\label{fig:decomp}
\end{figure}

\noindent\textbf{Sampling steps.} Table~\ref{tab:steps} shows that a few Euler steps improve bare-face recovery, while smaller budgets give lower layer errors. Direct one-step training has lower errors than every tested sampling budget, so its benefit comes from the training objective rather than from fewer sampling steps. Figure~\ref{fig:decomp} shows that extra steps leave face contours and noise in the recovered layers, and that layered generators and makeup removal do not isolate the makeup.

\noindent\textbf{Joint canvas.}\label{app:onecanvas} Predicting the white and black panels together in one canvas, instead of in two separate generations, makes them agree at soft edges, where agreement matters most. There the coverage error drops on both Composite and DailyScene validation looks, and reconstruction improves as well.

\begin{table}[t]
\centering
\caption{\textbf{Validity of decomposed layers.} We report pixel fractions and the median RGB spread inside the makeup (${\boldsymbol\alpha}>0.02$) and L1 ($\times100$) between re-composited and predicted panels on the test images of Table~\ref{tab:decomp_baselines}.}
\label{tab:validity}
\footnotesize\setlength{\tabcolsep}{4.2pt}
\begin{tabular}{@{}lcc@{}}
\toprule
 & DailyScene test & Composite test\\\midrule
Coverage outside $[0,1]$ & 0\% & $<$0.01\%\\
Median RGB spread of ${\mathbf W}-{\mathbf K}$ & 0.027 & 0.035\\
Clipped by more than 0.05 & 1.4\% & 4.6\%\\\midrule
Re-composited ${\mathbf K}$ & 0.001 & 0.017\\
Re-composited ${\mathbf W}$ & 0.03 & 0.12\\\bottomrule
\end{tabular}
\end{table}

\noindent\textbf{Layer validity.}\label{app:validity} The closed form assumes that ${\mathbf W}$ and ${\mathbf K}$ come from one layer: ${\mathbf W}-{\mathbf K}=(1-{\boldsymbol\alpha})\mathbf 1$ and $\mathbf 0\le{\mathbf K}\le{\boldsymbol\alpha}$ per channel. The decomposer predicts the panels without this constraint, and panels in the valid color range can still yield a color outside it. We therefore store ${\boldsymbol\alpha}=\mathrm{clip}(1-\mathrm{mean}_{\text{rgb}}({\mathbf W}-{\mathbf K}),0,1)$ and ${\mathbf F}=\mathrm{clip}({\mathbf K}/\max({\boldsymbol\alpha},\delta),0,1)$ with a small $\delta$, and every later warp, edit and composite uses this valid layer. Table~\ref{tab:validity} shows that clipping changes few pixels by more than 0.05 and that the re-compositing error of the stored layer is far below the errors against ground truth in Table~\ref{tab:decomp_baselines}. The residual concentrates in ${\mathbf W}$ because a single-channel coverage cannot absorb a colored ${\mathbf W}-{\mathbf K}$. The layer returned after rendering is recovered separately, with minimum coverage (Section~\ref{sec:renderer}).

\begin{table}[t]
\centering
\caption{\textbf{Effect of sampling steps in decomposition.} We report L1 ($\times100$) of the bare face and of the layer over white ($\mathbf W$) and black ($\mathbf K$); NFE counts transformer evaluations per image, and underlining marks the best sampling budget. All rows, including the reported one-step model, use the test images of Table~\ref{tab:decomp_baselines}.}
\label{tab:steps}
\footnotesize\setlength{\tabcolsep}{4.2pt}
\begin{tabular}{@{}lccccccc@{}}
\toprule
 & & \multicolumn{3}{c}{DailyScene test} & \multicolumn{3}{c}{Composite test}\\
\cmidrule(lr){3-5}\cmidrule(l){6-8}
Model & NFE & Bare\dn & $\mathbf W$\dn & $\mathbf K$\dn & Bare\dn & $\mathbf W$\dn & $\mathbf K$\dn\\\midrule
\multirow{5}{*}{Flow matching} & 1 & 5.19 & \underline{1.33} & \underline{0.42} & 5.24 & \underline{1.89} & 1.20\\
 & 2 & 3.63 & 1.69 & 0.49 & 3.81 & 2.13 & \underline{1.19}\\
 & 4 & \underline{3.30} & 1.88 & 0.53 & \underline{3.52} & 2.30 & 1.28\\
 & 8 & 3.43 & 2.06 & 0.57 & 3.65 & 2.51 & 1.41\\
 & 28 & 3.53 & 2.30 & 0.68 & 3.79 & 2.78 & 1.68\\\midrule
\rowcolor{oursrow}One-step training & 1 & \textbf{1.62} & \textbf{0.35} & \textbf{0.20} & \textbf{1.88} & \textbf{0.89} & \textbf{0.73}\\\bottomrule
\end{tabular}
\end{table}

%% file: sections/A-model.tex
\section{Rendering and Model Analysis}
\label{app:model}
We first show what rendering adds to the warped layers and whether generic operations can replace it, and then isolate the effects of conditioning, sampling, training and editable-region constraints. Unless stated otherwise, we select settings on the validation set described in Appendix~\ref{app:impl}, which Table~\ref{tab:s2abl} also reports. Condition and editable-region ablations start from the four-step setting of Table~\ref{tab:s2abl}, and the other analyses use the reported one-step setting; Appendix~\ref{app:val_renderer} lists the validation measurements.

\begin{figure*}[t]
\centering
\includegraphics[width=0.98\textwidth]{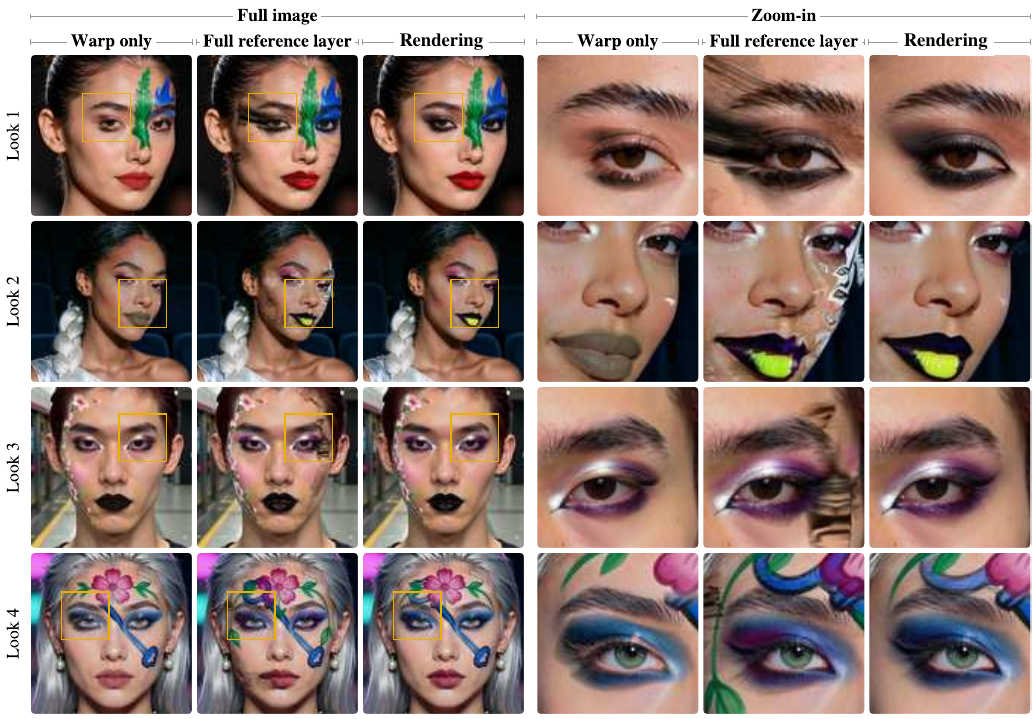}
\caption{\textbf{Full reference layer versus rendering.} Hybrid looks: warp only, the full reference layer pasted inside ${\boldsymbol{\Omega}}$ and rendering (reported configuration), with zooms where the latter two differ most. Pasting the full reference layer leaves removal residue, fragments of neighboring designs and broken strokes flat on the skin; rendering keeps the design more coherent.}
\label{fig:renderer_realism}
\end{figure*}

\noindent\textbf{Diagnostic quantities.} The \emph{full reference layer} is the warped layer before transfer-rule clipping; the renderer receives it as a condition (Table~\ref{tab:s2abl}), and as a diagnostic we also paste it directly inside ${\boldsymbol{\Omega}}$ (Figure~\ref{fig:renderer_realism}). Where the rules removed coverage, \emph{Gap} is the fraction of the RGB difference between the warp and this pasted layer that rendering closes. \emph{Band $\Delta E$} measures the mean CIE76 change in the band of skin inside ${\boldsymbol{\Omega}}$ but outside the design.

\input{sections/A-renderer_cmp}

\subsection{Conditioning and Sampling}

\noindent\textbf{Condition tokens.} The renderer relies mainly on the full reference layer. Removing this layer lowers DINO$_{\text{part}}$ and raises the color error while raising Face-ID (Table~\ref{tab:s2abl}, Table~\ref{tab:val_cond}), which indicates weaker transfer rather than better adaptation, and removing every condition lowers DINO$_{\text{part}}$ further. With one step, graying or removing the part crops, caption and maps lowers Face-ID and raises the color error (Table~\ref{tab:val_cond}), so they complement the full reference layer.

\begin{figure}[t]
\centering
\includegraphics[width=\linewidth]{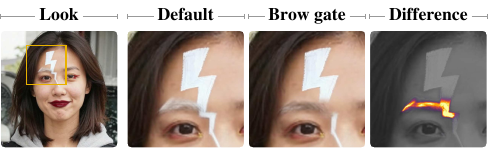}
\caption{\textbf{Optional brow gate.} A validation look and one brow (about $3\times$ zoom, yellow box), rendered by default and with the brow tokens blocked from the reference; the difference ($\times4$) is overlaid as heat on the dimmed gated render. The gate keeps the source's brow but interrupts the design where it crosses the brow.}
\label{fig:renderer_browgate}
\end{figure}

\noindent\textbf{Inference configuration.} The reported configuration, selected on the validation set, uses one Euler step from a moderate noise level and a small dilation of ${\boldsymbol{\Omega}}$ (Table~\ref{tab:hparams}). It keeps the brows editable, so that designs crossing the brows stay continuous (Figure~\ref{fig:renderer_browgate}). A single transformer evaluation costs a quarter of the four-step setting (Table~\ref{tab:s2abl}) and far less than the multi-step baselines (Table~\ref{tab:efficiency}).

\noindent\textbf{Sampler.} The start noise $\sigma$ sets the trade-off between identity and fidelity, and a single step from the reported noise level improves identity and color over four steps, also when both use the same small dilation (Table~\ref{tab:s2abl}, Table~\ref{tab:val_cond}).

\begin{figure*}[t]
\centering
\includegraphics[width=0.90\textwidth]{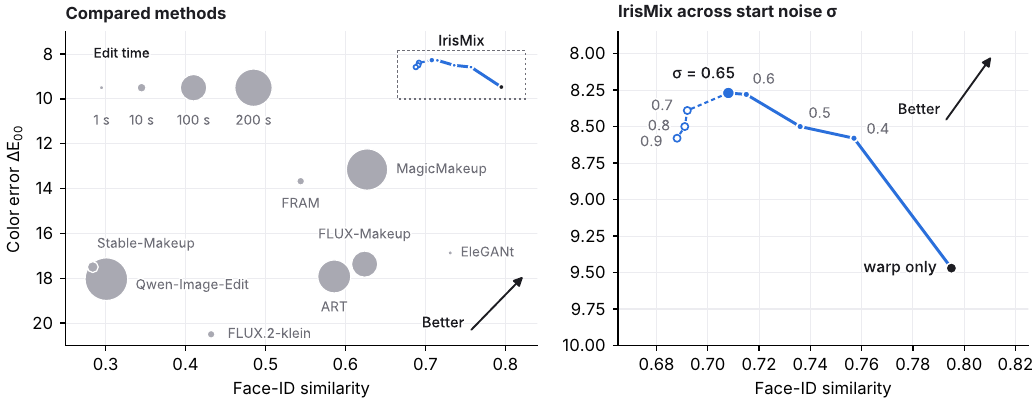}
\caption{\textbf{Comparison of identity and color fidelity.} Our curve has an advantage over the baselines on both axes. Rendering trades some identity for color accuracy up to $\sigma=0.65$; more noise loses both, and marker area shows that fidelity need not cost edit time.}
\label{fig:pareto}
\end{figure*}

\noindent\textbf{Identity and color fidelity.} Figure~\ref{fig:pareto} sweeps one-step rendering on the multi-reference hybrid looks. Up to the reported noise level, rendering trades some identity for a lower Color $\Delta E_{\text{part}}$ (CIEDE2000, $\Delta E_{00}$ in the plot), and more noise lowers both identity and color accuracy. Every comparison method is worse on both axes than at least one point of this curve.

\subsection{Editable Region and Output Layer}
\noindent\textbf{Editable region.} Without the masked residual, the decoded crop replaces the source: L2-M rises, skin outside ${\boldsymbol{\Omega}}$ changes, the skin band around the design is repainted (higher Band $\Delta E$) and Face-ID drops (Table~\ref{tab:s2abl}, Table~\ref{tab:val_cond}). We use a small dilation because it lowers Band $\Delta E$ while maintaining DINO$_{\text{part}}$ (Table~\ref{tab:val_cond}).

\noindent\textbf{Output layer.} Table~\ref{tab:val_layer} removes the minimum coverage from $\tilde{\boldsymbol\alpha}=\max({\boldsymbol\alpha}_{\text{d}},{\boldsymbol\alpha},{\boldsymbol\alpha}_{\hat{\mathbf I}})$ on validation renders. Without minimum coverage ${\boldsymbol\alpha}_{\hat{\mathbf I}}$, ${\mathbf{F}}'$ falls outside $[0,1]$ on a large share of covered pixels, and clipping it causes a visible re-compositing error inside ${\boldsymbol\Omega}$. Pasted coverage ${\boldsymbol\alpha}$ guarantees that every pasted pixel stays covered. Design coverage ${\boldsymbol\alpha}_{\text{d}}$ adds coverage on faint or skin-colored design regions, which keeps them accessible to recoloring even when rendering leaves their color unchanged.

\subsection{Training}
\noindent\textbf{Training ablations.} Each variant changes one component of the training recipe, with the reported one-step sampling throughout; the first and last rows of Table~\ref{tab:val_train} are the training rows of Table~\ref{tab:s2abl}.

Initialization from the decomposer benefits identity preservation: training LoRA$_{\text{ren}}$ from scratch lowers Face-ID at similar part similarity (Table~\ref{tab:val_train}). Replacing self-transfer with perturb-and-restore pairs of each item's own layer keeps Face-ID higher as in Table~\ref{tab:s2abl}, but lowers DINO$_{\text{part}}$ and raises the color error (Table~\ref{tab:val_train}). Figure~\ref{fig:selftransfer} shows why: perturb-and-restore pairs teach the renderer to stay close to the warp and simplify unusual designs, so it softens the feather strokes and replaces the two-tone lip with a uniform red, whereas self-transfer pairs, whose inputs carry real alignment and resampling errors, teach it to restore the design from the full reference layer.

\subsection{Regional Effects and Challenging Cases}
\input{sections/A-spill}

\noindent\textbf{Attention paths.} On FLUX.2, the color and style of an in-context reference reach the output through the text padding tokens, while identity passes from image to image~\cite{ge2026binding}. To trace how the reference reaches the output, we mask attention at inference on the validation looks with the reported sampler. Blocking text-to-reference attention in every block barely changes the result, so the reference does not act through the text. Blocking target-to-reference attention from an early block onward returns the output to warp only, and blocking it only in the late blocks still lowers Gap substantially. Like identity~\cite{ge2026binding}, makeup therefore follows image-to-image attention spread over many blocks.

\begin{figure*}[t]
\centering
\IfFileExists{figures/analysis/fig_failures.pdf}{\includegraphics[width=0.98\textwidth]{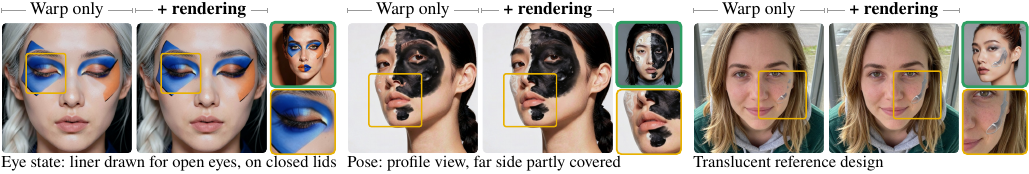}}{\fbox{\parbox{.9\linewidth}{\centering pending: make\_fig\_failures.py}}}
\caption{\textbf{Challenging cases.} Each case shows a LayerBeauty test item with warp only, + rendering, the reference framed in the color of its part and a zoom of the rendered result (yellow box); the label names the difficulty, and the translucent-design source is a LayerBeauty-DailyScene test face. We handle occluders with the occlusion-aware transfer illustrated in Figure~\ref{fig:occluder}.}
\label{fig:failures}
\end{figure*}

\begin{figure*}[t]
\centering
\includegraphics{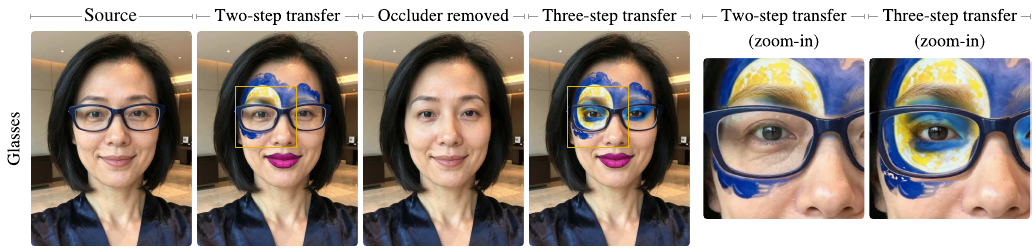}
\caption{\textbf{Challenging case for occlusion-aware transfer.} When face parsing misses parts of a frame, those parts are not restored on top: small gaps appear in the frame and specks of the made-up face show beside it (right zoom).}
\label{fig:occluder_fail}
\end{figure*}

\noindent\textbf{Glasses and bangs on the source.} The decomposer is fine-tuned to keep occluders out of the layer. Without occluder removal (\emph{two-step}: decomposition, then transfer), transfer paints under parsed hair and glasses, so it protects clear lenses together with the frames and loses the eye design behind them (Figure~\ref{fig:occluder}). Occlusion-aware transfer (\emph{three-step}) removes the occluder with the editor used for LayerBeauty, transfers to the occluder-free source $\mathbf T_{\text{free}}$, and puts the occluder back from the source photo $\mathbf T$ over our result $\hat{\mathbf I}$ on $\mathbf T_{\text{free}}$, $\hat{\mathbf I}_{\text{occ}}=\mathbf T+\boldsymbol\rho\odot(\hat{\mathbf I}-\mathbf T)$. The occluder layer $\boldsymbol\alpha_o$ is recovered from the two images with the minimum-coverage rule, and the transmittance $\boldsymbol\rho$ raises $1-\boldsymbol\alpha_o$ to a power that compensates for this lower bound. Face parsing of the photo refines $\boldsymbol\rho$: parsed hair stays opaque even where $\boldsymbol\alpha_o$ is low, and for glasses only the frames are put back, because the lenses refract the eyes, which then differ between the two images. In Figure~\ref{fig:occluder}, this transfer retains eye makeup behind the lenses and places hair over the design. Figures~\ref{fig:failures} and~\ref{fig:occluder_fail} show challenging cases.

\subsection{Lighting, Expression and Head Angle}
\label{app:lighting}
\noindent Figure~\ref{fig:lighting} applies one three-reference look, with the eyes, lips and face paint taken from three CreativeScene references, to one synthetic portrait whose lighting, expression and head angle were changed by FLUX.2-klein image editing. The three references are decomposed once, and only the transfer rules and the one-step render run again for each photo. The comparison methods use the multi-reference protocols of Appendix~\ref{app:hybrid}. In our results the face paint takes on the color and brightness of each light, stays on the closed eyelids and follows the turned and raised head, while the rest of each photo is largely kept. MagicMakeup transfers the design but keeps it at about the same brightness in every light, so it stands out in the dark and colored scenes, and it widens the design when the head turns. ART also carries the flowers of the eye reference into the face, because each of its steps transfers a whole reference, and its repeated steps add grain to the background in the dark scene. FRAM and MAD keep the photo but transfer little of the design.

%% file: sections/A-renderer_cmp.tex
\subsection{Qualitative Effect of Rendering}
\label{app:renderer_cmp}
\begin{figure}[t]
\centering
\includegraphics[width=\linewidth]{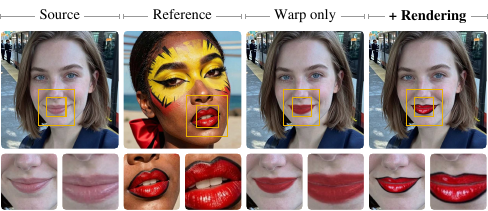}
\caption{\textbf{Effect of in-context rendering.} Rendering reduces compositing artifacts and restores details: the flat pasted color regains the reference's gloss and lip liner under the source's lighting.}
\label{fig:renderer_cmp}
\end{figure}
\begin{figure*}[t]
\centering
\includegraphics{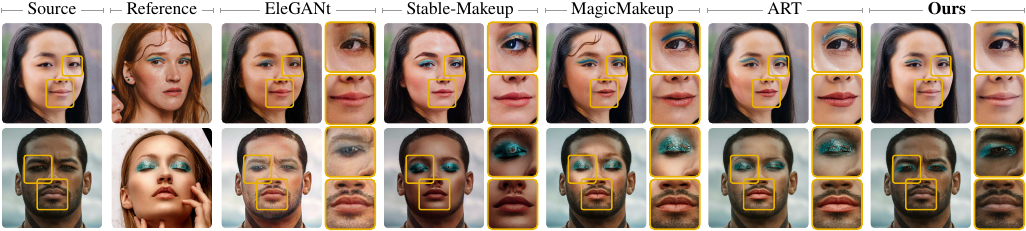}
\caption{\textbf{Effect of compositing on real photographs.} We transfer the reference layer onto MF2K photographs without rendering, preserving uncovered source pixels even under heavy artistic makeup. We quantify preservation in Table~\ref{tab:compare}: warping has an advantage in Face-ID and leaves outside-face pixels unchanged. The evaluated baselines alter identity, skin or background. We mark the zoomed regions with boxes.}
\label{fig:mf2k}
\end{figure*}

Figure~\ref{fig:renderer_cmp} compares the same decomposed layers before and after rendering, with the transfer rules fixed and with test references and sources unseen in training. The lip case below isolates a limitation of warping, which by itself already preserves uncovered pixels of real photographs (Figure~\ref{fig:mf2k}).

\noindent\textbf{Appearance mismatch.} Warped lips carry flat color without highlights or liner; rendering restores the reference's gloss and liner under the source's lighting.

\subsection{Alternatives to Rendering}
\label{app:renderctrl}
\begin{table}[!ht]
\centering
\caption{\textbf{Alternatives to the renderer} on the 100 hybrid looks of Table~\ref{tab:compare}, all applied to the same warp output and scored in one evaluation.}
\label{tab:renderctrl}
\scriptsize\setlength{\tabcolsep}{2.5pt}
\resizebox{0.8\linewidth}{!}{%
\begin{tabular}{@{}lrrr@{}}
\toprule
Method & DINO$_{\text{part}}$\up & Color $\Delta E_{\text{part}}$\dn & Assign\up\\\midrule
Warp only & 0.76 & 9.5 & 0.68\\
+ Poisson blending & 0.76 & 9.5 & 0.69\\
+ off-the-shelf editor & 0.75 & 16.8 & 0.45\\
+ untrained renderer & 0.76 & 9.8 & 0.72\\
+ rendering (ours) & 0.80 & 8.3 & 0.78\\\bottomrule
\end{tabular}}
\end{table}
We test whether a generic operation can replace the renderer by post-processing the warp-only outputs of Table~\ref{tab:compare}. \emph{Poisson blending} clones the warped makeup into the source in the gradient domain, near the region that the warp changed. The \emph{off-the-shelf editor} asks FLUX.2-klein to make the pasted makeup look naturally worn while keeping its design, and we keep its edit inside the same region. The \emph{untrained renderer} runs our conditions and one-step sampling on base FLUX.2-klein without the renderer LoRA.

Table~\ref{tab:renderctrl} shows that Poisson blending nearly matches the warp on every metric and that the untrained renderer only smooths it slightly, so neither refits liner nor restores gloss. The editor recolors many designs, for example blue eye makeup to green or gold, which lowers correct assignment and raises the color error markedly; its hard region boundary also leaves visible seams at the lips. Only the trained renderer raises part similarity, correct assignment and color accuracy together.

%% file: sections/A-spill.tex
\noindent\textbf{Spill.} \emph{Spill} restricts Leak $\Delta E$ (Table~\ref{tab:compare}) to unrequested pixels outside the warped coverage of the requested designs. This region depends only on the source, the references and the fixed warp, so it is the same for every method, and Spill enters the locality group of Table~\ref{tab:rank}.

%% file: sections/B-impl.tex
\section{Implementation Details}
\label{app:impl}

\begin{table*}[t]
\centering
\caption{\textbf{Training and inference settings.} The decomposer and the renderer adapt the same frozen backbone with one LoRA each; loss terms as in Sec.~\ref{sec:method}.}
\label{tab:hparams}
\scriptsize\setlength{\tabcolsep}{4pt}\renewcommand{\arraystretch}{1.12}
\begin{tabular}{@{}p{0.13\textwidth}p{0.405\textwidth}p{0.405\textwidth}@{}}
\toprule
 & Decomposer (decompose and harmonize) & Renderer\\\midrule
Backbone & \multicolumn{2}{l}{FLUX.2-klein-4B; VAE and text encoder frozen}\\
Trainable parts & LoRA of rank 32 on the attention and MLP projections; skips from the VAE encoder's first convolution and first two down blocks to decoder up blocks 0--2 through zero-initialized $1{\times}1$ convolutions & second LoRA of rank 32 on the same projections, initialized from the decomposer checkpoint (decomposition and harmonization) before occluder fine-tuning; skips as in the decomposer with new zero-initialized convolutions\\
Text input & two fixed instructions, encoded once and cached (Figure~\ref{fig:prompts}) & the per-part caption alone, cached per reference\\
Resolution & $512^2$ per panel of the canvas $[{\mathbf{B}}\,|\,{\mathbf{W}}\,|\,{\mathbf{K}}]$ & $512^2$ face crop, pasted back at full resolution through the masked residual; maps and crops at $256^2$ (Figure~\ref{fig:renderer_cond})\\
Training path & rectified flow on the canvas, then fine-tuning in the one-step regime; $H$ starts at $\sigma=0.6$ & input-anchored path; 65\% of the samples use $\sigma=0.7$ and one of the timesteps of a four-step grid, the rest $\sigma\sim\mathcal U[0.5,0.9]$ and $t=\sigma u$, $u\sim\mathcal U(0.02,1]$\\
Sampling & one step & one Euler step from $\sigma=t=0.65$; ${\boldsymbol{\Omega}}$ dilated by 2\,px\\
Loss weights & $\mathcal L_{\text{canvas}}$: L1 + 0.5\,LPIPS on ${\mathbf B}$ + 0.25 Laplacian pyramid on ${\boldsymbol{\alpha}}$; $\mathcal L_{\text{harmonize}}$: L1 in ${\boldsymbol{\Omega}}_{\text{H}}$ (normalized by its area) + 0.5\,LPIPS over the image; $\lambda_{\text{cycle}}=0.5$, $\lambda_{\text{identity}}=0.25$ & $\lambda_{\text{L1}}=1$, $\lambda_{\text{LPIPS}}=0.5$, faded in below $t=0.6$ (full at $t\le0.2$); $\mathcal L_{\text{aux}}$ in the text\\
Optimizer & AdamW, learning rate $5\times10^{-5}$ & AdamW, learning rate $2\times10^{-5}$, no weight decay, 100 warm-up steps, cosine decay to 10\%, gradient clipping at 1, EMA 0.999 with bias correction\\
Batch and steps & one sample per task; 2{,}000 steps, then 600 fine-tuning steps (select step 300) & batch 8 (4 with 2-step accumulation); 3{,}000 steps on the training pairs and auxiliary items of Table~\ref{tab:decomposer_data}\\
Training memory and time & 18.5\,GiB peak with the VAE decoder gradient-checkpointed & 40.5\,GiB peak, 14.1\,s per step\\
Parameters & \multicolumn{2}{l}{3.88\,B frozen transformer shared by the decomposer and the renderer; each LoRA 33.4\,M (rank 32) plus 0.23\,M skips; VAE 84\,M; text encoder 4.02\,B}\\
Hardware & \multicolumn{2}{l}{one NVIDIA RTX PRO 6000 Blackwell Max-Q (96\,GB) for training and all our timings}\\
Reported model & one-step model after 2{,}000 steps of joint decomposition and harmonization (L1 in ${\boldsymbol{\Omega}}_{\text{H}}$, stronger perturbations), fine-tuned with occluder pairs (highlighted in Table~\ref{tab:cycle}) & bias-corrected EMA weights after 3{,}000 steps, selected on the renderer validation set\\\bottomrule
\end{tabular}
\end{table*}
Table~\ref{tab:hparams} summarizes the settings, including the weights of Eq.~\eqref{eq:decompose}; Tables~\ref{tab:efficiency} and~\ref{tab:edit_cost} report inference memory, FLOPs and runtime.

\subsection{Decomposition and Harmonization}

\begin{figure*}[t]
\centering
\includegraphics[width=0.98\textwidth]{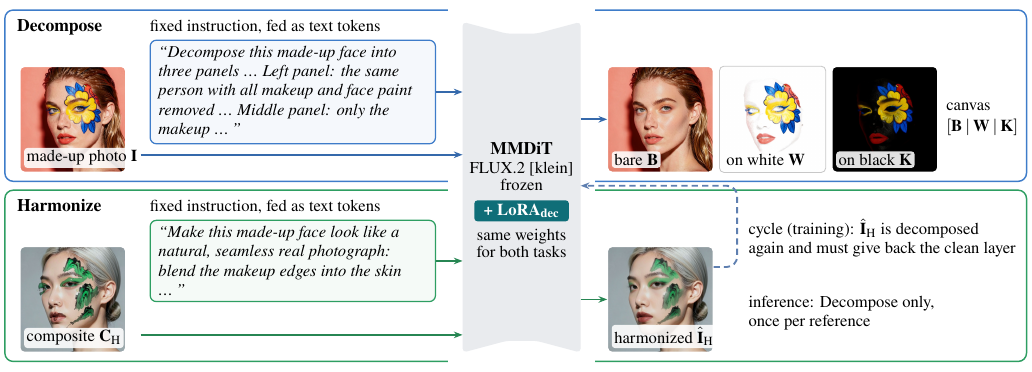}
\caption{\textbf{Decomposition and harmonization with one LoRA.} Depending only on the fixed instruction (excerpts; full text in Figure~\ref{fig:prompts}), the frozen MMDiT with the single LoRA$_{\text{dec}}$ decomposes a made-up photograph into the canvas or harmonizes a perturbed paste ${\mathbf{C}}_{\text{H}}$ into a photograph (one-step outputs on validation items). Training mixes both tasks and chains them in the cycle; inference runs only Decompose, once per reference.}
\label{fig:twotasks}
\end{figure*}

\begin{figure}[t]
\centering
\setlength{\fboxsep}{4pt}
\fcolorbox{black!25}{black!3}{\parbox{\dimexpr\columnwidth-2\fboxsep-2\fboxrule\relax}{\scriptsize
\textbf{Decompose.} Decompose this made-up face into three panels with exactly the same framing as the photo. Left panel: the same person with all makeup and face paint removed, natural bare skin, natural lips and brows, identical identity, expression, hair, lighting and background. Middle panel: only the makeup, over a pure white background. Right panel: the same makeup, over a pure black background. Skin, eyes, teeth, hair, clothes and background are not part of the makeup.\\[3pt]
\textbf{Harmonize.} Make this made-up face look like a natural, seamless real photograph: blend the makeup edges into the skin and match the makeup to the face's lighting. Keep the makeup design, colours and position exactly, and keep the person's identity, expression, hair and background exactly the same.}}
\caption{\textbf{Decomposer instructions.} One LoRA serves both tasks: the fixed instruction alone tells it whether to split a made-up face into bare face and layer or to blend a pasted layer into the photo.}
\label{fig:prompts}
\end{figure}

\noindent\textbf{Decomposer canvas.} The output latent is three reference widths wide; its tokens and the reference tokens share the joint attention, and the decoded canvas is split into its three panels. Zero-initialized skips from the encoder of the reference feed the bare panel~\cite{pix2pixturbo}.

\noindent\textbf{Decomposer losses.} Rectified-flow pre-training~\cite{liu2023flow} on the canvas upweights the squared error on the dilated makeup region in the layer panels and on the lips and eyes in the bare panel. In the one-step regime, $\mathcal L_{\text{canvas}}$ adds LPIPS~\cite{lpips} and a Laplacian-pyramid loss on the closed-form ${\boldsymbol{\alpha}}$~\cite{fbamatting}, with per-panel weights that switch off panels a dataset cannot supervise (Table~\ref{tab:decomposer_data}). Both terms of $\mathcal L_{\text{harmonize}}$ compare with ${\mathbf{I}}$, the made-up image that the unperturbed layer reproduces.

\noindent\textbf{Occluder fine-tuning.} Glasses and bangs must stay above the makeup rather than enter the layer. We briefly fine-tune the decomposer on the training data of Table~\ref{tab:decomposer_data} with added occluder pairs (Table~\ref{tab:hparams}). Each pair starts from a real photograph with glasses or bangs and an edited copy without them. A CreativeScene layer is warped onto the occluder-free face, and the occluder is composited back on top, so the bare panel keeps the occluder. For glasses, the layer target is the makeup visible around the frames; for bangs, the photograph is the base, makeup appears only on visible facial skin and the target is the full design. Faces with the occluder alone and an empty layer serve as negatives. We visually review all pairs and split identities using the protocol in Appendix~\ref{app:protocol}, reserving part of them for model selection.

\noindent\textbf{Harmonize perturbations.} Independent random draws perturb the pasted layer in ${\mathbf{C}}_{\text{H}}$ (Figure~\ref{fig:composite}, bottom) with a smooth warp, wrong occlusion by the amodal layer over the hair, hard or blurred edges, a strength error, a shading and color mismatch and JPEG compression. The rows of Table~\ref{tab:cycle} before stronger perturbations use only a milder warp and a milder color mismatch.

\begin{figure*}[t]
\centering
\includegraphics[width=0.98\textwidth]{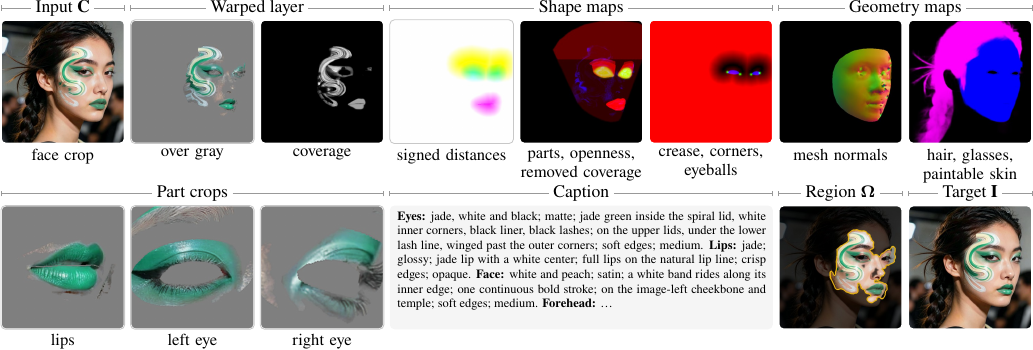}
\caption{\textbf{Renderer inputs and supervision.} Conditions: composite, full warped layer and coverage, three shape and two geometry maps (three channels each), part crops and caption. The editable region $\boldsymbol\Omega$ is not tokenized, and the target $\mathbf I$ is used only as supervision.}
\label{fig:renderer_cond}
\end{figure*}

\subsection{Rendering: Conditions and Training}

\begin{figure*}[t]
\centering
\includegraphics[width=\textwidth]{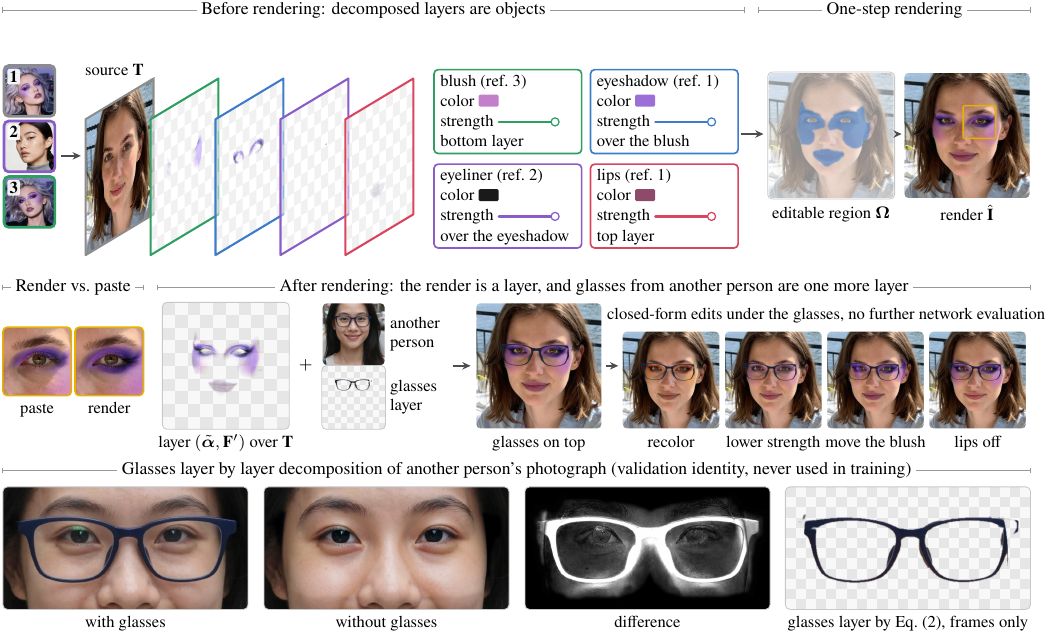}
\caption{\textbf{Editable layers before and after rendering.} Top: decomposed parts become layer objects on the source $\mathbf T$ with adjustable color, strength, position and order (here one reference's eyeliner over another's eyeshadow); morphology on the layers gives $\boldsymbol\Omega$, and the one-step render refines the paste. Middle: the render is a layer $(\tilde{\boldsymbol\alpha},\mathbf F')$ over $\mathbf T$; glasses from another person are one more layer on top, and edits beneath them stay closed-form. Bottom: the glasses layer, decomposed by Eq.~\eqref{eq:minalpha} and scaled by face width.}
\label{fig:layerobj}
\end{figure*}

Figure~\ref{fig:layerobj} follows one look through rendering: the makeup is a set of editable layers both before and after the render.

\noindent\textbf{Map contents.} The shape maps encode distances to the eyelids, lip outline and brow line, part labels, eye openness, removed coverage, and the eyelid crease, eye corners and eyeballs. The geometry maps encode mesh normals, hair, glasses and paintable skin. Captions describe color, finish, motif, placement, edges and opacity from LayerBeauty prompts, and decomposed reference layers supply measured colors.

\noindent\textbf{Condition encoding.} The composite ${\mathbf C}$, full warped layer over gray and coverage are VAE-encoded at $512^2$ on the target's $32\times32$ position grid. Five maps use $256^2$ resolution on the same grid at half density; three part crops use $256^2$ with separate position offsets (Figure~\ref{fig:renderer_cond}). All condition tokens share the target timestep and participate in joint attention.

\noindent\textbf{Editable region and layer recovery.} The editable region is computed before rendering, from the source and references:
\begin{equation}
\begin{aligned}
{\boldsymbol{\Omega}}&=g_{2}\ast\mathrm{dil}_{r}\big(\mathbf{M}\cup[{\boldsymbol{\alpha}}>0.02]\big)\odot(1-\mathbf{P}),\\
\mathbf{P}&=\big(\mathbf{P}_{\text{eye}}\cup\mathbf{P}_{\text{mouth}}\cup\mathbf{P}_{\text{hair}}\cup\mathbf{P}_{\text{glasses}}\cup\mathbf{P}_{\text{unreq}}\big)\setminus\mathbf{M}.
\end{aligned}
\label{eq:omega}
\end{equation}
\begin{figure}[!ht]
\centering
\includegraphics[width=\linewidth]{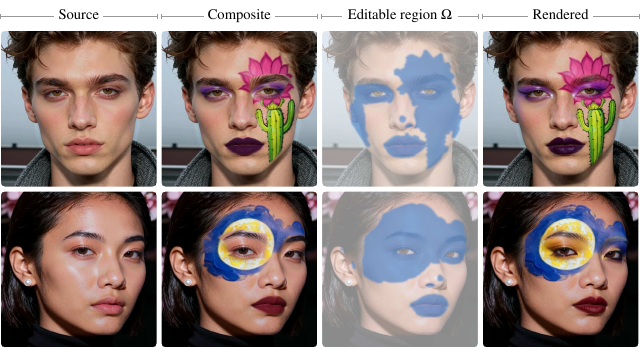}
\caption{\textbf{Editable region ${\boldsymbol\Omega}$ on two hybrid looks.} ${\boldsymbol\Omega}$ (blue) covers the warped designs and pasted makeup and excludes the eyes, the mouth interior, hair and unrequested parts; the renderer changes only these pixels.}
\label{fig:omega}
\end{figure}
\begin{figure}[!ht]
\centering
\includegraphics[width=\linewidth]{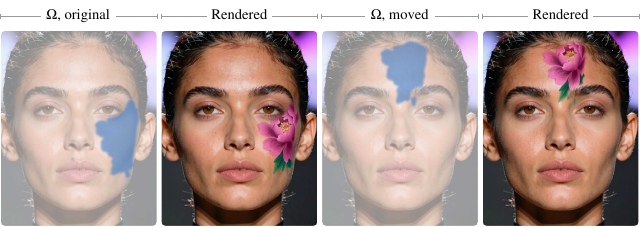}
\caption{\textbf{Editable region ${\boldsymbol\Omega}$ after layer edits.} ${\boldsymbol\Omega}$ is rebuilt from the edited layer before each render: moving the peony to the forehead moves the editable region with it, and the cheek it leaves returns to the source.}
\label{fig:omega_move}
\end{figure}
Figure~\ref{fig:omega} shows ${\boldsymbol\Omega}$ on hybrid looks; since it is rebuilt from the current layers before each render, it follows layers that users move (Figure~\ref{fig:omega_move}). The design mask $\mathbf M$ grows confident full warped coverage into adjacent weaker coverage, discards tiny components and also includes pasted coverage. Here $\mathrm{dil}_r$ denotes dilation by $r$ pixels and $g_{2}$ a small Gaussian blur. Table~\ref{tab:hparams} gives $r$, and Appendix~\ref{app:model} analyzes larger dilations.

Protected regions $\mathbf P$ come from face parsing and landmarks: slightly eroded eyeballs, which retain lash-line liner, the inner-lip polygon, confidently parsed hair and hats, glasses, and unrequested parts. Subtracting $\mathbf M$ from their union keeps design pixels editable even when they overlap these regions, and every pixel outside ${\boldsymbol\Omega}$ keeps its source value.

\noindent\textbf{Recovering the output layer.} Design coverage ${\boldsymbol\alpha}_{\text d}$ is restricted to paintable skin. Minimum coverage ${\boldsymbol\alpha}_{\hat{\mathbf I}}$ follows Eq.~\eqref{eq:minalpha} with $(\hat{\mathbf I},\mathbf T)$, accounting for coverage restored by the renderer. With pasted coverage, they define $\tilde{\boldsymbol\alpha}$ (Sec.~\ref{sec:renderer}). Re-compositing the recovered layer reproduces $\hat{\mathbf I}$ without error where $\tilde{\boldsymbol\alpha}>\tau$ and with a negligible error elsewhere.

\noindent\textbf{Renderer training inputs.} Self-transfer warps the layer to a second training identity with a different eye openness and back with the same mesh; lips and brows follow the rules applied on the item's own face. To match the decomposer's output, most items scale their coverage by the decomposer's measured retention with a random per-item strength. The input equals ${\mathbf{I}}$ outside the dilated design, and items whose eye openness differs noticeably between ${\mathbf{I}}$ and the bare face are dropped. Real photographs of MT~\cite{beautygan} (training split), BeautyFace~\cite{beautyrec} and CelebA-HQ~\cite{celebamaskhq} use their closed-form layer and predicted bare face, with the loss restricted to the face.

\noindent\textbf{Primary losses.} $\mathcal L_{\text{FM}}$ in Eq.~\eqref{eq:render} follows the input-anchored path from ${\mathbf{z}}_{\mathbf{I}}$ to ${\mathbf{x}}_\sigma$, replacing the path of Sec.~\ref{sec:layer}:
\begin{equation*}
\mathcal L_{\text{FM}}=\mathbb E_{\sigma,t,\boldsymbol\epsilon}\Big[{\textstyle\sum_p}\,w_p\big\lVert\big(\mathbf v_\theta(\mathbf x_t,t,\mathbf c)-\mathbf v^\ast\big)_p\big\rVert_2^2\big/{\textstyle\sum_p}\,w_p\Big],
\end{equation*}
with latent tokens $p$, $\mathbf x_t=\mathbf z_{\mathbf I}+t\,\mathbf v^\ast$, $\mathbf v^\ast=(\mathbf x_\sigma-\mathbf z_{\mathbf I})/\sigma$, $t\le\sigma$ and ${\mathbf{x}}_\sigma=(1-\sigma){\mathbf{z}}_{\mathbf{C}}+\sigma{\boldsymbol{\epsilon}}$. The regional weight $w_p$, which is high inside ${\boldsymbol\Omega}$ and low outside, is multiplied by the token-pooled pixel weight
\begin{equation*}
1+3\,\mathrm{clip}\!\left(\frac{\Delta E^\ast_{ab}(\mathbf I,\mathbf B)}{20}
+\frac{0.5\lVert\nabla(\mathbf I-\mathbf B)\rVert}{0.1},0,1\right).
\end{equation*}
LPIPS is the spatial VGG distance map, masked by ${\boldsymbol\Omega}$ before averaging; L1 and LPIPS are normalized by the area of ${\boldsymbol\Omega}$; this pixel weight stresses makeup and edges in L1.

\noindent\textbf{Auxiliary losses.} $\mathcal L_{\text{aux}}$ combines preservation, structure and appearance terms, with weights in parentheses. Preservation uses L1 to keep pixels inside ${\boldsymbol\Omega}$ but outside the design at ${\mathbf C}$ (1). Structure uses masked Laplacian-pyramid L1 (0.3), BlendFace~\cite{blendface} identity (0.5) and 2D-FAN~\cite{fan2017} landmarks (0.05); the latter two start later in training. Appearance uses per-component CIEDE2000~\cite{ciede2000} (0.02), SigLIP~\cite{siglip} distance on lip and eye crops (1), lip-gloss statistics on photograph targets (0.1), and a hinge discriminator on crop SigLIP features (0.05), which also starts later.

\noindent\textbf{Renderer checkpoint selection.} We periodically score the bias-corrected EMA weights on the renderer validation set, which combines hybrid looks of several validation references with single-reference looks on validation bare faces. We rank checkpoints by DINO$_{\text{part}}$ and Gap, defined in Appendix~\ref{app:model}, retaining those whose Face-ID stays close to that of the pasted full reference layer and whose color $\Delta E_{\text{part}}$ is no worse than the warp alone. The selected final checkpoint (Table~\ref{tab:hparams}) ranks first on both appearance criteria.

\noindent\textbf{Warp and zones.} The warp uses the MediaPipe face-mesh landmarks~\cite{grishchenko2020attentionmesh}, extended by rings over the forehead. The seam between the upper and lower zones is searched in a band between the eyes and the lip centroid; it never crosses the lips and minimizes the smoothed ${\boldsymbol{\alpha}}$ plus a weak pull toward the middle of this band.

%% file: sections/C-data.tex
\section{LayerBeauty Construction}
\label{app:data}
LayerBeauty uses Z-Image-Turbo~\cite{zimage} for portrait generation and FLUX.2-klein~\cite{flux2} for editing. DailyScene begins with bare faces generated from story cards. CreativeScene begins with made-up portraits generated from theme cards; the editor then removes makeup from the full face and, in a second pass, the eyes. Every retained pair includes its prompt, part descriptions and identity labels.

\begin{figure*}[p]%
\centering
\includegraphics{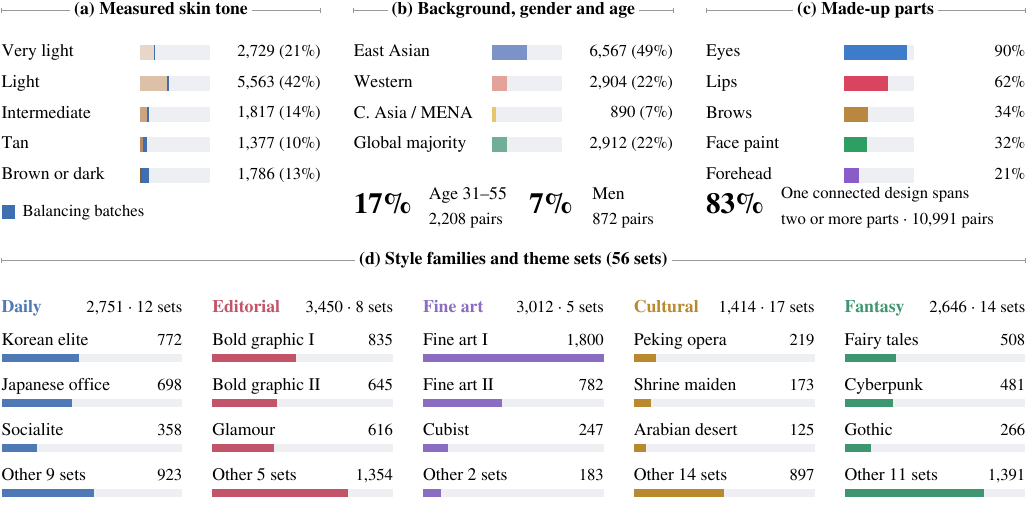}
\caption{\textbf{CreativeScene composition.} Skin tone (a; blue: balancing batches), background, gender and age (b), made-up parts (c; a part counts when opaque design covers 5\% of it) and the three largest theme sets of each of the five style families (d).}
\label{fig:suppstats}
\par\vspace{2pt}
\includegraphics{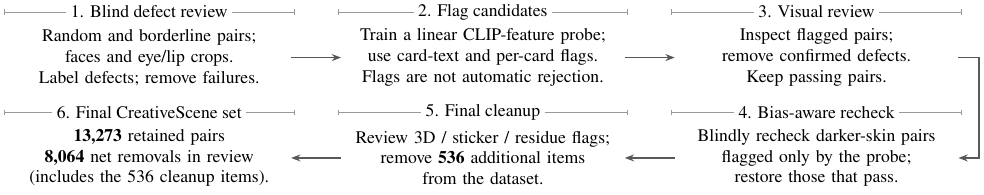}
\caption{\textbf{Visual quality control.} Flagged pairs are reviewed; darker-skin pairs flagged only by the CLIP probe are rechecked (Appendix~\ref{app:data}).}
\label{fig:defect_audit}
\par\vspace{2pt}
\begin{minipage}{\textwidth}
\newcommand{\grouphead}[1]{\noindent\makebox[\linewidth]{{\color{black!40}\rule[.5ex]{.4pt}{0pt}\rule[\dimexpr.55ex-1.25pt]{.4pt}{2.5pt}\leaders\hrule height \dimexpr.55ex+.2pt depth -\dimexpr.55ex-.2pt\hfill}\kern2pt{\footnotesize\bfseries #1}\kern2pt{\color{black!40}\leaders\hrule height \dimexpr.55ex+.2pt depth -\dimexpr.55ex-.2pt\hfill\rule[\dimexpr.55ex-1.25pt]{.4pt}{2.5pt}}}}
\captionsetup[subfigure]{font=footnotesize,skip=1pt}
\grouphead{LayerBeauty-CreativeScene}\par\vspace{1pt}
\begin{subfigure}[t]{115.3pt}
\centering
\includegraphics{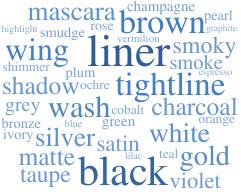}
\caption{Eyes}\label{fig:words-creative-eyes}
\end{subfigure}\hfill
\begin{subfigure}[t]{115.3pt}
\centering
\includegraphics{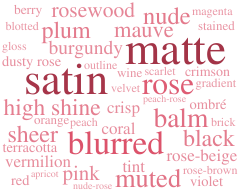}
\caption{Lips}\label{fig:words-creative-lips}
\end{subfigure}\hfill
\begin{subfigure}[t]{115.3pt}
\centering
\includegraphics{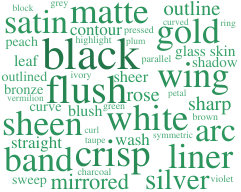}
\caption{Face paint}\label{fig:words-creative-face}
\end{subfigure}\hfill
\begin{subfigure}[t]{115.3pt}
\centering
\includegraphics{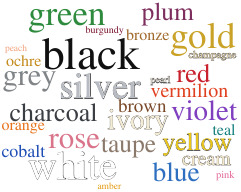}
\caption{Palette}\label{fig:words-creative-palette}
\end{subfigure}
\par\smallskip
\grouphead{LayerBeauty-DailyScene}\par\vspace{1pt}
\begin{subfigure}[t]{115.3pt}
\centering
\includegraphics{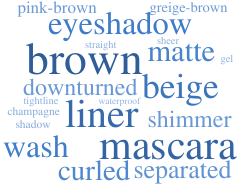}
\caption{Eyes}\label{fig:words-daily-eyes}
\end{subfigure}\hfill
\begin{subfigure}[t]{115.3pt}
\centering
\includegraphics{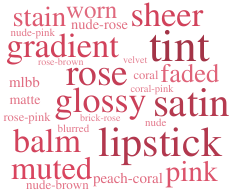}
\caption{Lips}\label{fig:words-daily-lips}
\end{subfigure}\hfill
\begin{subfigure}[t]{115.3pt}
\centering
\includegraphics{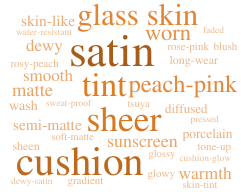}
\caption{Base and cheeks}\label{fig:words-daily-face}
\end{subfigure}\hfill
\begin{subfigure}[t]{115.3pt}
\centering
\includegraphics{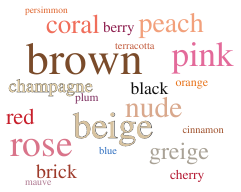}
\caption{Palette}\label{fig:words-daily-palette}
\end{subfigure}
\end{minipage}
\caption{\textbf{Makeup vocabulary.} Part-specific terms of CreativeScene (a--d) and DailyScene (e--h); word size follows the number of pairs.}
\label{fig:suppwords}
\end{figure*}

\noindent\textbf{Dataset overview.} As shown in Figure~\ref{fig:datapipe}, filtering retains a subset of the CreativeScene renders. Figures~\ref{fig:suppstats} and~\ref{fig:suppwords} summarize statistics and vocabulary, and Appendix~\ref{app:gallery} shows examples.

\subsection{Prompt Construction}
\noindent\textbf{Card content.} Story cards (DailyScene) and theme cards (CreativeScene) describe adults only. Cards exclude sexualized clothing or poses, scene text and logos, and facial objects outside the makeup description. CreativeScene themes also exclude real people, brands, artists, film or anime characters, sacred or religious symbols, and realistic wounds or blood.

\noindent\textbf{Card validation.} A theme card must avoid global exclusion patterns, family-specific rules and regional quotas. Structural checks require a detailed focal face-art description, a short palette and unobstructed eyes, lips and design. Each card must differ in wording from every existing card, and combinations of colors and motifs close to evaluation designs are excluded. The rules also reject sticker-like or raised finishes, occluders and props.

\noindent\textbf{Prompt sanitization.} An adult persona sets age and ethnicity according to the card's background. We remove hands or tools at the face, lying poses and application instructions, and we rewrite attached decorations as flat paint. A fixed clause requests pigment texture and empty hands and excludes attached, glossy or sticker-like elements and text. The clause avoids naming tools because negated objects can appear without classifier-free guidance. Lighting, backgrounds, accessories, peripheral occlusion and seeds vary.

\noindent\textbf{Balancing.} For non-regional themes, a fixed share of prompts uses global-majority personas (African, African-diaspora, South and Southeast Asian, Latin American, Pacific and mixed backgrounds with darker Monk Skin Tone levels). Smaller shares use male personas, except for female-coded themes, or older adults with mature skin. Dedicated balancing batches make every prompt global-majority, male or older, and per-pair sampling weights by measured skin tone, gender and age subset are provided.

\subsection{Image Filtering and Review}
\noindent\textbf{Structural consistency.} We reject CreativeScene pairs whose eye area changes too much between the bare and made-up images. In Figure~\ref{fig:datapipe}, \emph{edit drift} denotes DailyScene edits that change mouth shape or smear skin, and \emph{aborted} denotes CreativeScene renders without a removal result.

\noindent\textbf{Skin-tone correction.} The editor can lighten skin during removal, including the neck. We therefore keep the made-up image outside the face of the bare image and re-anchor facial skin to the unpainted skin of the made-up image with a spatially varying lightness gain, which can lighten or darken.

\noindent\textbf{Skin-tone rejection.} After correction, we measure median CIELAB lightness and the individual typology angle, $\mathrm{ITA}=\arctan((L^\ast-50)/b^\ast)$. Darker skin, identified by a low ITA or a darker-skin prompt, faces strict limits on lightening, darkening, ITA shift and change in the face--neck lightness gap, whereas lighter skin is rejected only for large lightness changes. The limits are measured on unpainted regions to avoid face-paint bias. Figure~\ref{fig:datapipe} shows the pairs that the eye and skin gates remove.

\noindent\textbf{Additional DailyScene filters.} DailyScene uses quality, phone-in-frame (OWLv2~\cite{owlv2}), skin, diversity, identity-novelty, and lip and eye artifact gates. A final pass removes hands on faces, dark-brown lips and manually flagged pairs, and Table~\ref{tab:decomposer_data} lists the remaining pairs. Ambiguous cases are checked manually and excluded if in doubt.

\begin{table*}[t]
\centering
\caption{\textbf{Training data and splits.} LayerBeauty subsets, Composite pairs, FFHQ-Makeup and occluder pairs, each split by identity (Appendix~\ref{app:protocol}); bold totals match Figure~\ref{fig:datapipe}. Decomposer training samples the datasets with the listed shares, occluder fine-tuning adds the glasses and bangs pairs of the last row (Appendix~\ref{app:impl}), and the renderer trains on the listed pairs plus 1{,}788 auxiliary items (1{,}417 synthetic, 371 real photographs). Layers come from Eq.~\eqref{eq:minalpha} or are composited by Eq.~\eqref{eq:layer}. ${\mathbf B}$ denotes the bare face and ${\mathbf W}$/${\mathbf K}$ the white/black layer composites.}
\label{tab:decomposer_data}
\providecommand{\dashfill}{\xleaders\hbox to 3pt{\hss\rule{1.5pt}{0.3pt}\hss}\hfill}
\footnotesize\setlength{\tabcolsep}{2.7pt}
\begin{tabular}{@{}ll>{\raggedright\arraybackslash}p{4.2cm}llrrrrrrr@{}}
\toprule
 & & & & & \multicolumn{5}{c}{Pairs} & \multicolumn{2}{c}{Use in training}\\
\cmidrule(lr){6-10}\cmidrule(l){11-12}
Dataset & Subset & Bare face and layer & Split by & Panels & Train & Val & Test & Unseen & Total & Decomp. & Render.\\\midrule
CreativeScene & makeup-first & FLUX.2-klein removal; \mbox{by Eq.~\eqref{eq:minalpha}} & identity & ${\mathbf{B}}$, ${\mathbf{W}}$, ${\mathbf{K}}$ & 11{,}739 & 323 & 493 & 718 & \textbf{13{,}273} & 16.7\% & 2{,}536\\\midrule
DailyScene & bare-first & generated bare face; FLUX.2-klein adds makeup; \mbox{composited by Eq.~\eqref{eq:layer}} & identity & ${\mathbf{B}}$, ${\mathbf{W}}$, ${\mathbf{K}}$ & 4{,}366 & 122 & 306 & -- & \textbf{4{,}794} & 22.2\% & 4{,}290\\\midrule
Composite & warped layers & independent bare face; warped CreativeScene layer; \mbox{composited by Eq.~\eqref{eq:layer}} & layer, face & ${\mathbf{B}}$, ${\mathbf{W}}$, ${\mathbf{K}}$ & 3{,}745 & 41 & 162 & -- & 3{,}948 & 33.3\% & --\\\noalign{\vskip1.5pt\hbox to\textwidth{\dashfill}\vskip1pt}
FFHQ-Makeup & filtered & real photograph & FFHQ id & ${\mathbf{B}}$ & 9{,}738 & 191 & 363 & -- & 10{,}292 & 27.8\% & --\\\midrule
Occluder & glasses, bangs & real photograph with the occluder and an edited copy without it; warped CreativeScene layer & identity & ${\mathbf{B}}$, ${\mathbf{W}}$, ${\mathbf{K}}$ & 820 & 135 & -- & -- & 955 & fine-tuning & --\\\bottomrule
\end{tabular}
\end{table*}

\noindent\textbf{Defect review.} Figure~\ref{fig:defect_audit} separates manual labeling, CLIP-probe screening, visual review and restoration. The coding agent screens all CreativeScene pairs using faces and eye and lip crops, and human reviewers assisted by MCP agent tools then decide disputed pairs blindly. The review categories cover raised or sticker-like finishes, hands or tools, removal residue, skin-tone shift, lip artifacts and other artifacts. Confirmed failures are removed; pairs flagged only by the probe are restored if a blind human recheck passes.

%% file: sections/G-gallery.tex
\subsection{Gallery}\label{app:gallery}
\begin{figure*}[p]
\centering
\includegraphics[width=\textwidth,height=0.86\textheight,keepaspectratio]{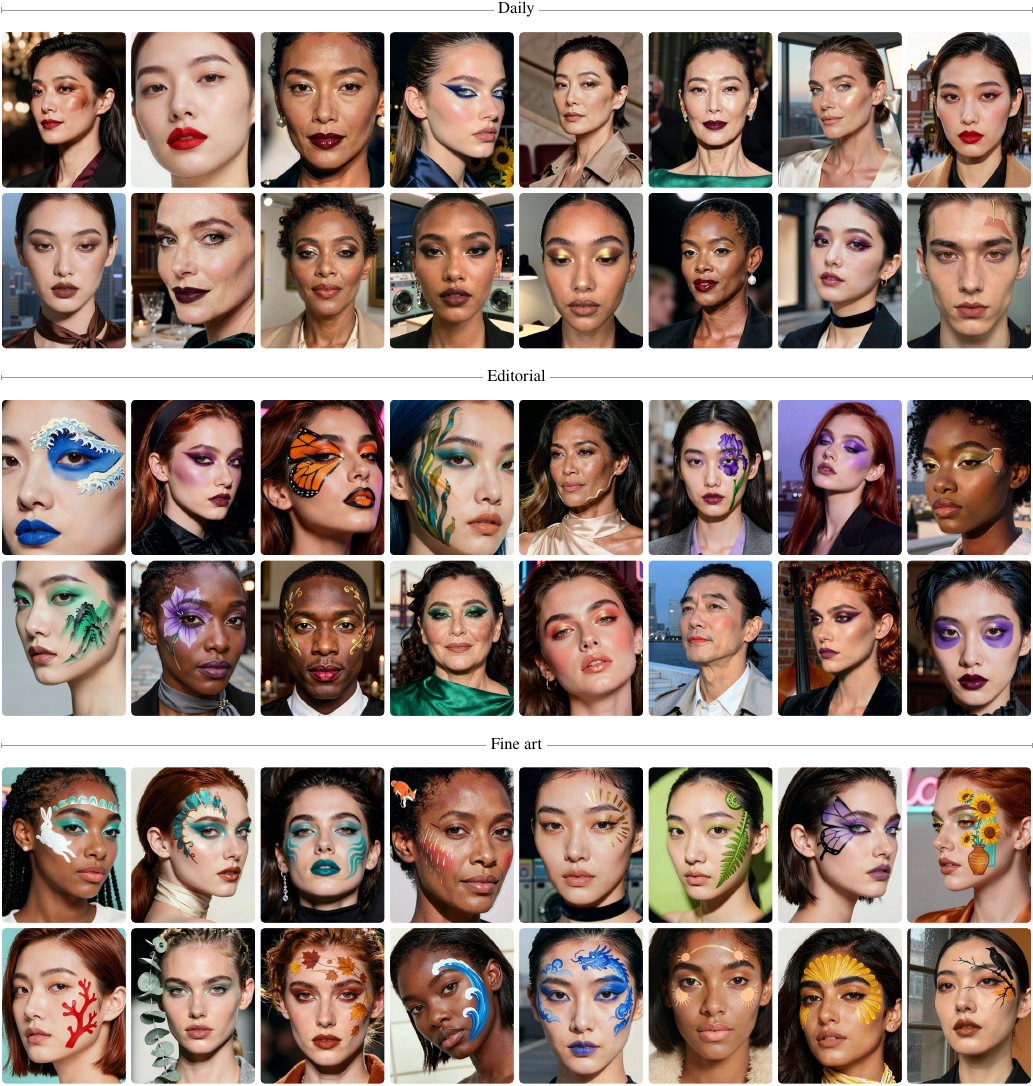}
\caption{\textbf{LayerBeauty gallery: daily, editorial and fine-art designs.} We show 16 CreativeScene portraits in each block, grouped by the style family of their theme cards (Figure~\ref{fig:suppstats}): daily (top), editorial (middle) and fine-art (bottom). We illustrate variations in liner geometry, lip color and finish in the daily group; saturated eye makeup, metallic accents and graphic contours in the editorial group; and floral, wave and abstract motifs extending across facial regions in the fine-art group.}\label{fig:gallery1}
\end{figure*}

\begin{figure*}[p]
\centering
\includegraphics[width=\textwidth,height=0.86\textheight,keepaspectratio]{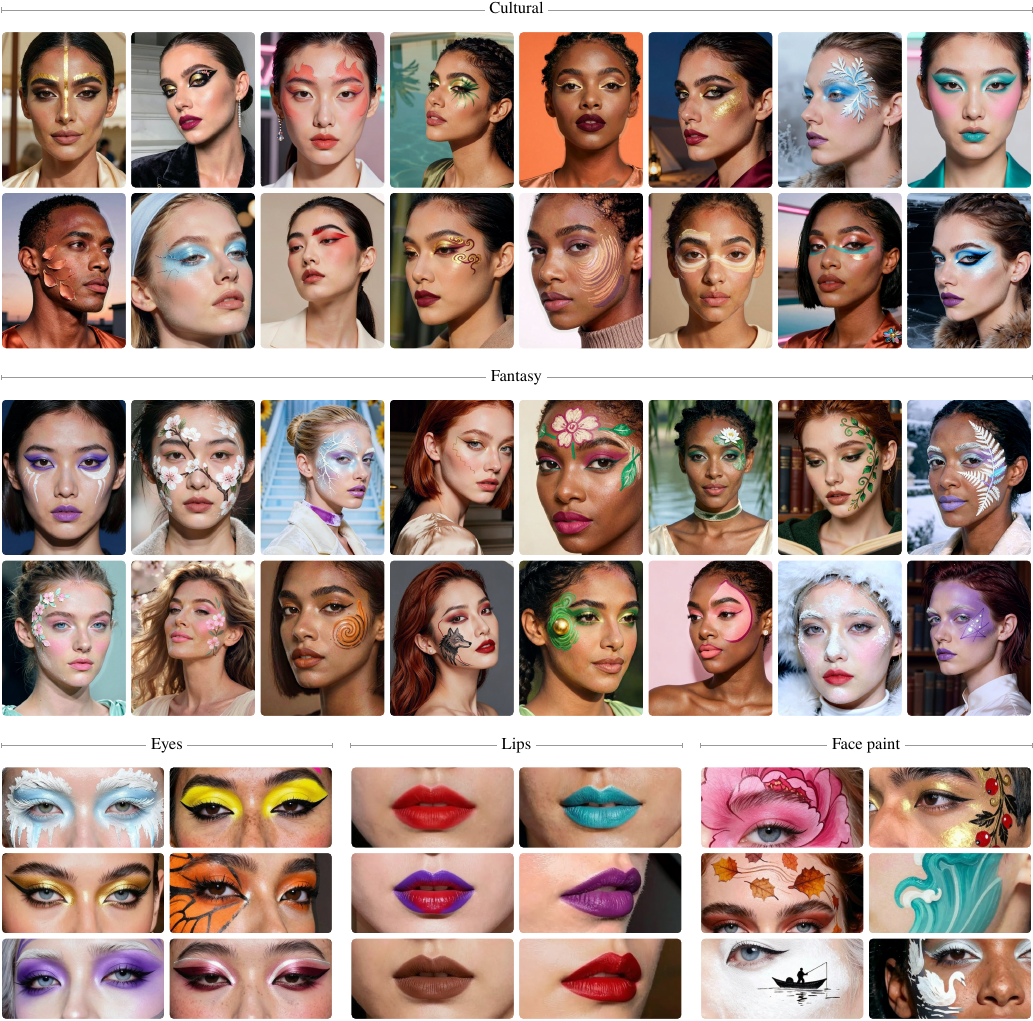}
\caption{\textbf{LayerBeauty gallery: cultural and fantasy themes with local details.} We show 16 culturally themed CreativeScene portraits in the upper block and 16 fantasy-themed portraits in the middle block. We display six additional close-ups per part below: eyes (left), lips (center) and face paint (right). These crops expose liner boundaries, metallic and gradient lip finishes, and painted motifs such as flowers, leaves and waves at a scale where their local structure is visible.}
\label{fig:gallery2}
\end{figure*}

\begin{figure*}[p]
\centering
\includegraphics[width=\textwidth,height=0.86\textheight,keepaspectratio]{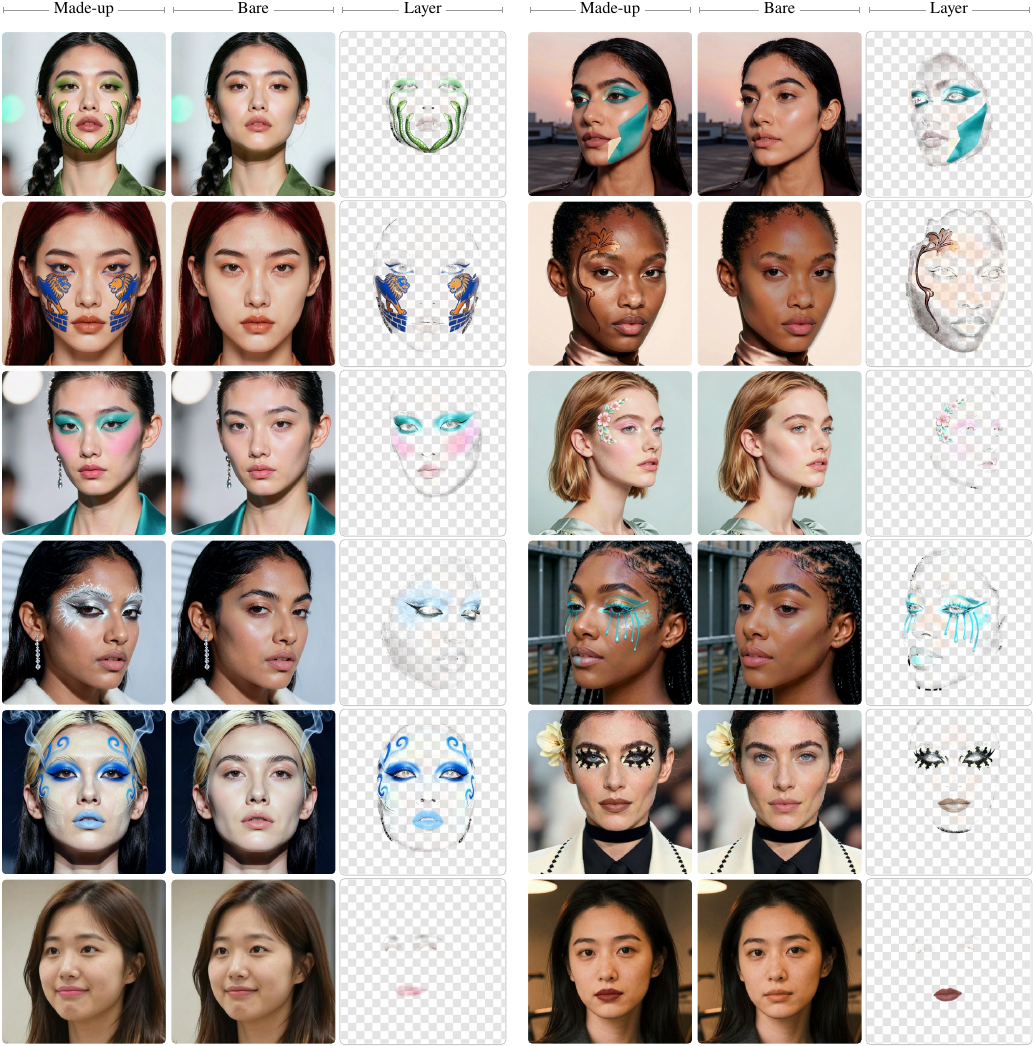}
\caption{\textbf{LayerBeauty gallery: paired portraits and explicit layers.} We arrange 12 dataset examples in six rows, with two triplets per row. Each triplet shows the made-up portrait, paired bare face and makeup layer over a checkerboard. In rows~1--5, we show CreativeScene examples with editor-estimated bare faces and minimum-coverage layers computed from the pairs. In the last row, we show DailyScene examples with the original bare faces and the layers used to construct their made-up counterparts. The checkerboard reveals the coverage of both extended face paint and localized lip or eye makeup.}
\label{fig:gallery3}
\end{figure*}

\begin{figure*}[p]
\centering
\includegraphics[width=\textwidth,height=0.86\textheight,keepaspectratio]{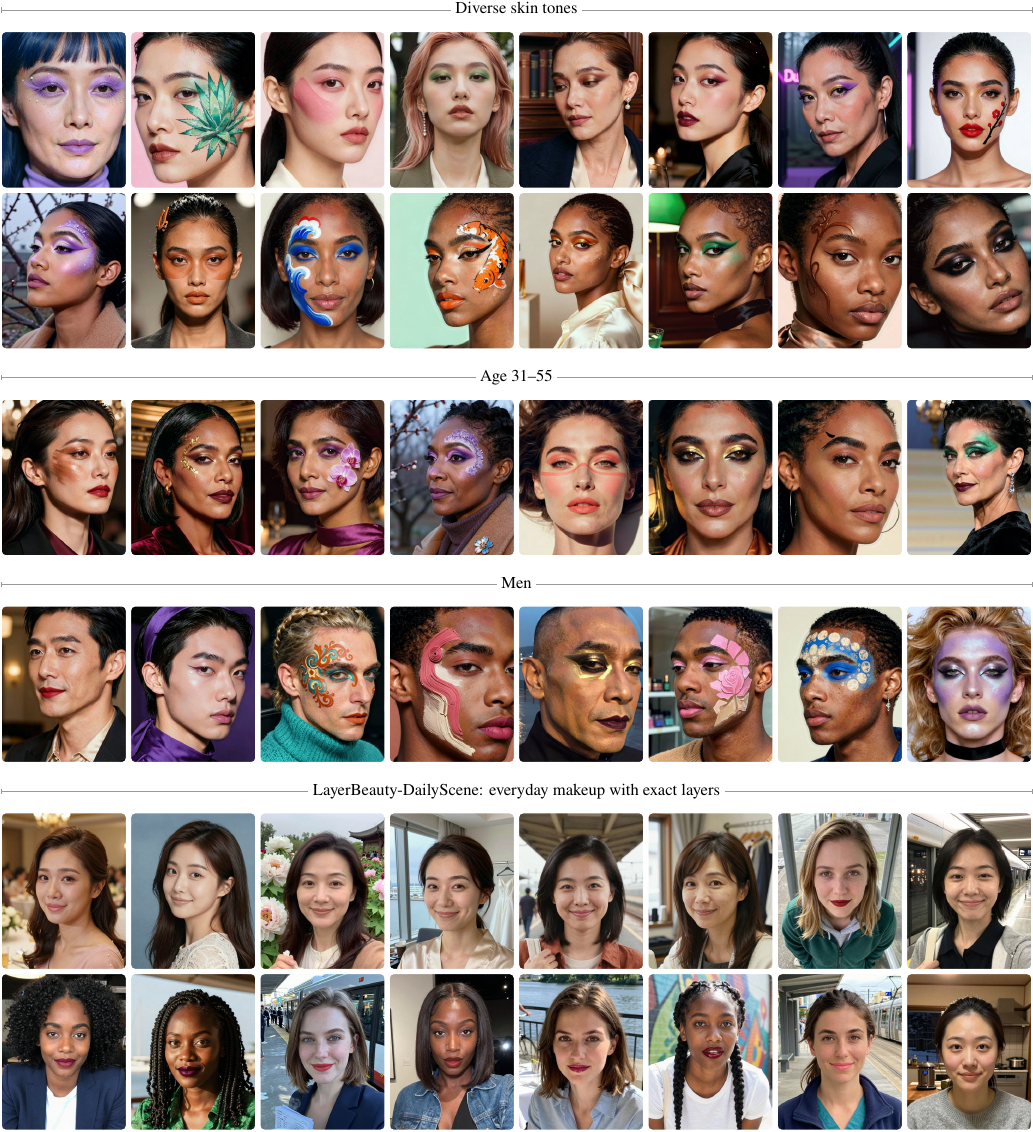}
\caption{\textbf{LayerBeauty gallery: diverse people.} The first 16 CreativeScene portraits span the range of measured bare-face skin tones. We then show eight examples from the 31--55 age subset and eight men from the balancing batches. In the bottom block, we show 16 DailyScene portraits with everyday makeup across indoor and outdoor settings. These examples complement the style-based organization in Figures~\ref{fig:gallery1} and~\ref{fig:gallery2} by showing the people and settings on which the designs appear.}
\label{fig:gallery4}
\end{figure*}

Figures~\ref{fig:gallery1}--\ref{fig:gallery4} show LayerBeauty pairs that passed the filters above.

%% file: sections/D-protocol.tex
\begin{figure*}[p]%
\centering
\includegraphics{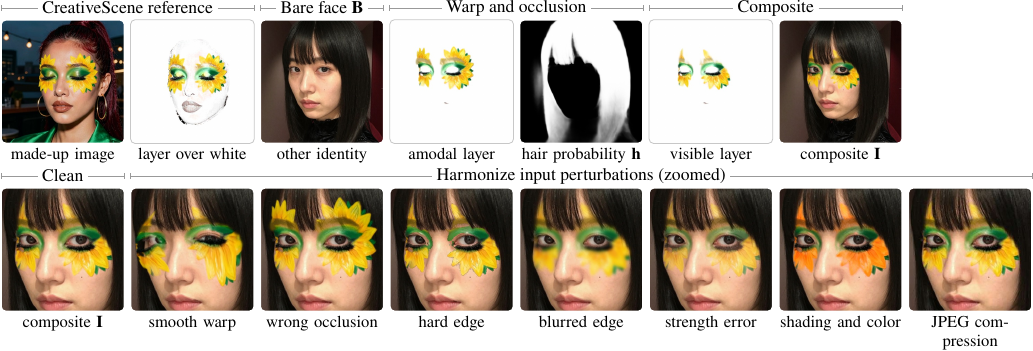}
\caption{\textbf{Layer-based supervision and harmonization training.} A CreativeScene layer warped onto an independent bare face, with hair occlusion, gives a pair composited by construction; the lower row shows the compositing errors that harmonization learns to correct.}
\label{fig:composite}
\par\vspace{2pt}
\includegraphics{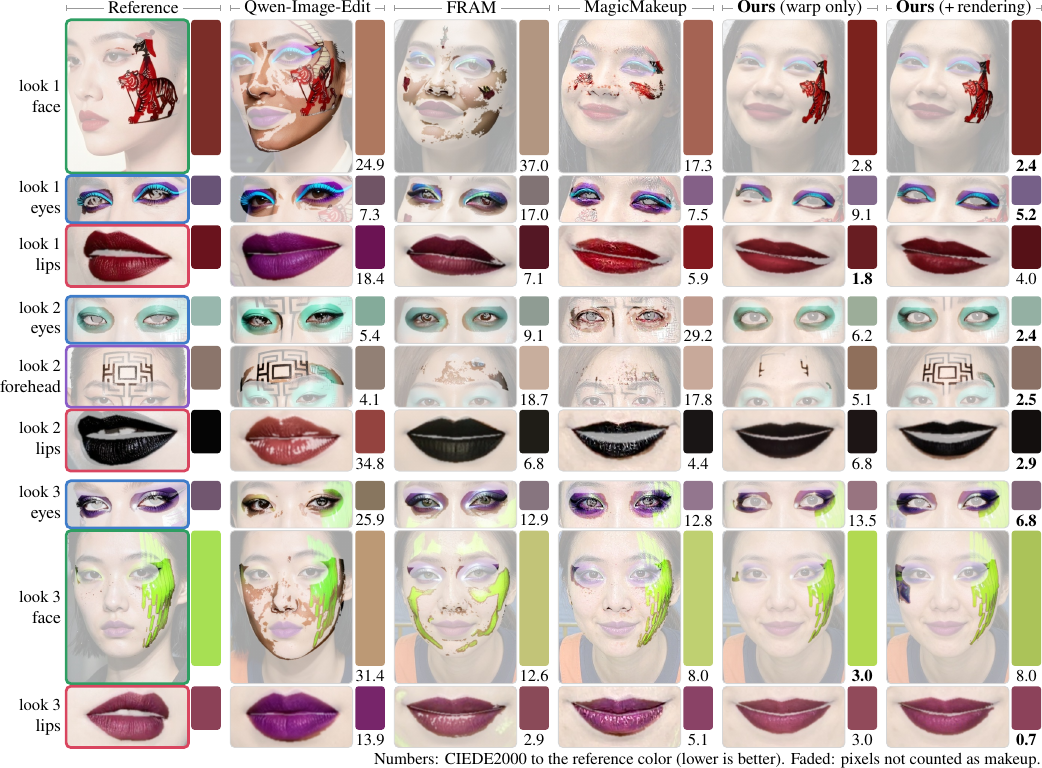}
\caption{\textbf{Comparison of per-part color fidelity and source preservation.} Explicit layers keep each part close to its reference color with fewer changes to the source's features, whereas baselines alter facial features and lip contours, consistent with their lower Face-ID in Table~\ref{tab:compare}.}
\label{fig:colour}
\end{figure*}

\section{Experimental Protocol}
\label{app:protocol}
\noindent\textbf{Model selection.} Hyperparameters, checkpoints and inference settings are chosen on validation splits. Test tables report the selected configurations and the stated ablations. We run all baselines with their official code, weights and settings (Appendix~\ref{app:hybrid}).

\subsection{Splits and Composite Construction}

\noindent\textbf{Splits.} Table~\ref{tab:decomposer_data} lists the splits of each dataset. CreativeScene is split by person group (one shared person description). DailyScene keeps the training and test identities of its generation and sets aside a small share of its training identity clusters for validation. FFHQ-Makeup is split by FFHQ~\cite{ffhq} id and supervises only the bare panel, since its layer does not reproduce the image by construction. Three CreativeScene theme sets (European fairy tales, cyberpunk and cyber-runner face art) form an \emph{unseen-style} split that is never trained on. No identity is shared between the training, validation and test lists, and no test layer or face enters a training Composite pair.

\nocite{ffhqaging}

\noindent\textbf{Composite construction.} As illustrated in Figure~\ref{fig:composite}, we pair a CreativeScene layer with an independent bare face from the same split. A piecewise-affine warp over a triangulation of MediaPipe landmarks~\cite{grishchenko2020attentionmesh} resamples the color, coverage and part weights of the layer. Pairing requires agreement in pose and expression.

\noindent\textbf{Occlusion by hair.} Makeup lies \emph{under} the hair: the bare face's soft hair/hat probability ${\mathbf{h}}$ from face parsing (SegFormer~\cite{segformer} fine-tuned on CelebAMask-HQ~\cite{celebamaskhq,faceparsing_hf}) occludes it, ${\mathbf{I}}={\mathbf{B}}+(1-{\mathbf{h}})\odot{\boldsymbol{\alpha}}\odot({\mathbf{F}}-{\mathbf{B}})$, \ie Eq.~\eqref{eq:layer} with coverage $(1-{\mathbf{h}})\odot{\boldsymbol{\alpha}}$, while the stored layer is \emph{amodal}, containing the full design under the hair. The decomposer predicts the amodal layer, so makeup hidden by the reference's hair is completed rather than lost, and the renderer occludes it with the source's hair.

\noindent\textbf{Transfer rules.} Composite pairs and the rendering warp transfer the design of a CreativeScene layer and drop only faint washes that cannot be separated from removal residue (skin texture, specular sheen, freckles); daily washes come from the composited LayerBeauty-DailyScene layers instead. Lips follow the source's lip shape and are always kept, brows keep the source's shape and receive only their tint, and the eye rim keeps liner and lash lines. Nostrils, teeth, eyeballs and the mouth interior never carry makeup, and hair occludes it as above; the remaining face inside its contour is the paintable skin to which transferred coverage is restricted.

\noindent\textbf{Exclusions.} We drop every reference with a hand or tool near the face, detected by OWLv2~\cite{owlv2} and MediaPipe hand landmarks~\cite{mediapipehands}. Among the planned Composite pairs, we skip light looks with little visible design, which DailyScene covers, and pairs whose bare face wears glasses. The validation and test Composite pairs are frozen evaluation sets, and training Composite pairs train only the decomposer.

\noindent\textbf{FFHQ-Makeup.} Its authors regenerate the made-up photographs, which changes background, skin texture and occasionally identity, so we keep only pairs with high structural agreement between the bare and made-up photographs, and the bare target takes the low-frequency color of the real bare face inside the face and the high-frequency texture of the made-up photograph.

\subsection{Evaluation Sets}
\noindent\textbf{Transfer benchmarks.} We evaluate on MF2K test pairs, CPM-Real pattern pairs and multi-reference hybrid looks whose references come from the held-out CreativeScene test and unseen-style splits; Appendix~\ref{app:hybrid} gives the settings.

\noindent\textbf{Decomposition and diagnostics.} Validation Composite pairs give Table~\ref{tab:cycle} and select checkpoints, the validation occluder pairs of Table~\ref{tab:decomposer_data} select the occluder fine-tuning step, and validation images serve the joint-canvas analysis (Appendix~\ref{app:onecanvas}).

%% file: sections/E-eval.tex
\section{Evaluation Details}
\label{app:hybrid}
\input{sections/tab_efficiency}\subsection{Baseline Inputs and Inference}

\noindent\textbf{MagicMakeup.} We use the official MagicMakeup checkpoint on FLUX.1-Kontext-dev~\cite{kontext} and unchanged inference code with its default settings. Each call takes one source, one reference and a mask that selects the eyes, the lips or the full face. Three-reference looks chain calls with official masks: full-face transfer for face paint and forehead comes first, then eyes, then lips, and each output becomes the next source. This order protects localized edits, and cheek- and forehead-only modes are unavailable.

\noindent\textbf{Multi-image editors.} Qwen-Image-Edit-2511~\cite{qwenimageedit2511} and FLUX.2-klein use their official sampling settings with a fixed seed and receive one prompt with the source and the three references. On MF2K and CPM-Real, the input and prompt are adapted to one source and one reference.

\noindent\textbf{Single-reference methods.} Stable-Makeup and ART (official LoRA) have no region control and are applied once per reference in the same order as MagicMakeup. We show intermediate steps of the single-reference methods in Figure~\ref{fig:gpt}. FLUX-Makeup uses the sampling settings and prompt published in its paper. MAD~\cite{mad} uses its official low-resolution translation model with inpainting and attention-based refinement, one call per reference, and pastes the translated face back into the source, which keeps its L2-M low.

\noindent\textbf{Native regional composition.} FRAM takes one reference per region (skin and nose, eyes and brows, mouth); forehead designs use the skin region, and unrequested regions use the source. FRAM therefore cannot represent looks with separate forehead and face-paint references. Its face crop uses official soft-border pasting. EleGANt combines lip, skin and eye references in one run with the same assignments and crop pasting.

\subsection{Metrics and Interpretation}
\label{app:metrics}
\noindent\textbf{Preservation.} We score all methods on identical pairs using MagicMakeup's official \texttt{evaluate/} code at the resolution of each benchmark. Face-ID is the source--output cosine similarity of AdaFace ViT-B/KP-RPE~\cite{adaface,kprpe,adaface_cvlface}. L2-M is the mean squared error outside the source face contour. Neither metric alone measures successful makeup transfer, because an unchanged source preserves identity perfectly.

\noindent\textbf{Reference similarity.} Part-wise DINO and CLIP similarities (DINO$_{\text{part}}$ and CLIP$_{\text{part}}$, from the DINO-I and CLIP-I of the official protocol) compare output and reference crops located by MediaPipe, graying pixels outside the part and averaging over the three parts. Whole-image variants follow the official protocol. \emph{Assign} is the fraction of requested parts whose CIELAB change is closest to the change of their assigned reference. We average metrics over all hybrid looks.

\noindent\textbf{Unrequested changes.} \emph{Leak} is mean CIE76 $\Delta E^\ast_{ab}$ from the source within unrequested face parts. We restrict this measure to pixels outside the requested designs to compute \emph{Spill}, as defined in Appendix~\ref{app:model}.

\noindent\textbf{Color fidelity.} Color $\Delta E_{\text{part}}$ is the CIEDE2000 difference between the makeup colors of output and reference, where makeup pixels are selected by a CIE76 color-difference threshold between the reference and its bare face, and between the output and its source. As shown in Figure~\ref{fig:colour}, baselines can alter both makeup hues and facial features.

\begin{table}[t]
\centering\setlength{\aboverulesep}{0.2ex}\setlength{\belowrulesep}{0.3ex}
\caption{\textbf{Recovered and ground-truth reference layers.} Hybrid looks of Table~\ref{tab:compare} with reference layers recovered by the decomposer from the reference portraits, as in Table~\ref{tab:compare}, or taken from the LayerBeauty ground truth.}
\label{tab:e2e}
\scriptsize\setlength{\tabcolsep}{0pt}
\begin{tabular*}{\linewidth}{@{\extracolsep{\fill}}llrrrrr@{}}
\toprule
Reference layers & Output & ID\up & L2-M\dn & DINO$_{\text{part}}$\up & Color $\Delta E_{\text{part}}$\dn & Assign\up\\\midrule
\rowcolor{oursrow}Decomposer (Table~\ref{tab:compare}) & warp only & 0.79 & 11.8 & 0.76 & 9.5 & 0.68\\
\rowcolor{oursrow} & + rendering & 0.71 & 15.4 & 0.80 & 8.3 & 0.78\\
Ground truth & warp only & 0.78 & 13.3 & 0.77 & 8.9 & 0.69\\
 & + rendering & 0.67 & 17.8 & 0.81 & 7.7 & 0.81\\\bottomrule
\end{tabular*}
\end{table}

\noindent\textbf{Recovered and ground-truth layers.} For our hybrid rows in Table~\ref{tab:compare}, the decomposer recovers layers from the same reference portraits the baselines receive. Table~\ref{tab:e2e} compares them with the ground-truth layers of these references. The recovered layers reach nearly the part similarity and color error of the ground-truth layers and preserve identity better.

\begin{table}[t]
\centering
\caption{\textbf{Grouped average rank on the hybrid looks.} Lower is better. Fidelity: DINO$_{\text{part}}$, CLIP$_{\text{part}}$, Color $\Delta E_{\text{part}}$, Assign; identity: ID, L2-M; locality: Leak $\Delta E$, Spill. Unedited Source performs no transfer.}
\label{tab:rank}
\scriptsize\setlength{\tabcolsep}{3pt}
\begin{tabular}{@{}lrrr@{}}
\toprule
Method & Fidelity & Identity & Locality \\\midrule
\rowcolor{oursrow}\method{} warp only & 3.5 & 1.0 & 1.0 \\
\rowcolor{oursrow}\method{} + rendering & 1.5 & 2.5 & 2.5 \\
EleGANt & 7.0 & 3.5 & 4.0 \\
FRAM & 3.5 & 5.0 & 2.5 \\
MagicMakeup & 4.25 & 4.0 & 7.0 \\
FLUX-Makeup & 7.0 & 6.0 & 6.0 \\
FLUX.2-klein & 7.5 & 8.5 & 7.5 \\
ART & 9.0 & 6.0 & 5.5 \\
Stable-Makeup & 6.75 & 9.0 & 9.0 \\
Qwen-Image-Edit & 5.0 & 9.5 & 10.0 \\
\bottomrule
\end{tabular}
\end{table}

\noindent\textbf{Grouped rank.} Table~\ref{tab:rank} averages per-metric ranks within fidelity, identity and locality over the editing methods: warp only leads on identity and locality, and rendering leads on fidelity.

\noindent\textbf{Pattern makeup on CPM-Real.} References are CPM-Real~\cite{cpm} photographs that a fixed CLIP zero-shot rule marks as face paint or pattern makeup and that pass the same content screens as MF2K; sources are MF2K bare faces. CPM-Real's terms forbid redistributing its images, so we report quantitative results only.

\begin{figure*}[t]
\centering
\includegraphics[width=0.74\textwidth]{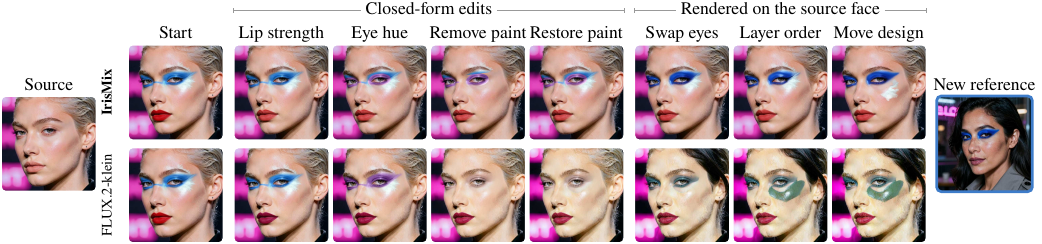}
\caption{\textbf{Comparison of multi-round editing with FLUX.2-klein.} From the same start, regenerating the previous output accumulates changes in skin, hair and background, whereas restarting from the layers and the source face keeps person and scene stable.}
\label{fig:multiround_cmp}
\end{figure*}

\subsection{Runtime and Repeated Editing}
\input{sections/tab_edit_cost}
\noindent\textbf{Time per image.} Every time in Table~\ref{tab:compare} and Table~\ref{tab:efficiency} comes from one benchmark with batch one on one RTX PRO 6000 Blackwell Max-Q, warmed-up runs with per-image caches cleared before each run, and baselines with official settings and precision, including all chained calls. Our runtime is similar across the evaluated looks. Reference decomposition is reusable, and Table~\ref{tab:compare} adds it for the three references of each look.

\noindent\textbf{Amortized editing cost.} Table~\ref{tab:edit_cost} lists per-operation costs. Recovery is paid once per reference, layer edits need no network, and one render adapts the appearance of a stacked or moved design.

\subsection{Multi-Round Editing}
\label{app:multiround}
\noindent\textbf{Setup.} We apply seven edits to hybrid looks: lip strength, eye hue, removing and restoring the face paint, swapping the eyes, reordering layers and moving the design. FLUX.2-klein starts from our initial render, and each round takes its previous output and a text instruction, plus the new reference for the swap. Our closed-form rounds edit the layer in closed form, and the other rounds recompose the layers and render once on the source face. Drift is the CIEDE2000 difference between consecutive rounds on the face outside the part edited in that round (dilated); cumulative drift compares each round with the start on the face outside every part edited so far, \ie in never-edited regions. The metrics were fixed before the runs.

\begin{table}[!ht]
\centering
\caption{\textbf{Multi-round editing over 30 looks.}}
\label{tab:multiround}
\scriptsize\setlength{\tabcolsep}{3pt}
\begin{tabular}{@{}lrrlrr@{}}
\toprule
 & \multicolumn{2}{c}{Drift $\Delta E$} & & \multicolumn{2}{c}{Success}\\
\cmidrule(lr){2-3}\cmidrule(lr){5-6}
Round & Ours & FLUX.2-klein & Success measure & Ours & FLUX.2-klein\\\midrule
Lip strength & 0.03 & 3.29 & lip $\Delta E$ ratio (ideal 0.5) & 0.55 & 0.98\\
Eye hue & 0 & 2.59 & hue error ($^\circ$) & 16.7 & 17.9\\
Remove face paint & 0 & 9.06 & $\Delta E$ to source & 0 & 7.05\\
Restore face paint & 0 & 1.95 & $\Delta E$ to before & 0 & 5.78\\
Swap eyes & 1.11 & 2.38 & $\Delta E$ to reference & 8.9 & 14.1\\
Layer order & 0.38 & 5.05 & correct order (5 looks) & 5/5 & 1/5\\
Move design & 1.66 & 1.55 & design moved & 97\% & 27\%\\\bottomrule
\end{tabular}
\end{table}

\begin{figure}[!ht]
\centering
\includegraphics[width=0.68\linewidth]{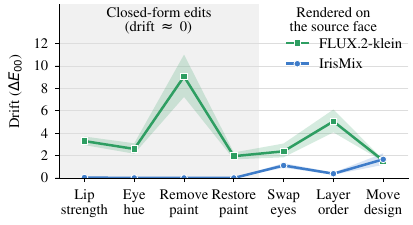}
\caption{\textbf{Comparison of drift per round.} Closed-form rounds leave other regions nearly unchanged, and rendered rounds restart from the source face.}
\label{fig:multiround_drift}
\end{figure}

\noindent\textbf{Results.} Table~\ref{tab:multiround} and Figure~\ref{fig:multiround_drift} report each round, and Figure~\ref{fig:multiround_cmp} shows an example. Closed-form rounds leave other regions nearly unchanged, and rendered rounds change them slightly but restart from the source face. In the move round, our layers move the design in most looks, whereas FLUX.2-klein rarely does (Table~\ref{tab:multiround}). A longer prompt that asks FLUX.2-klein to preserve identity and scene gives similar drift.

\begin{table}[!ht]
\centering
\caption{\textbf{Multi-round editing with full history.} FLUX.2-klein regenerates every round from the source face and all references, prompted with the edit list or the target state; 30 looks. Sequential is the FLUX.2-klein column of Table~\ref{tab:multiround}; per-round drift is measured outside the part edited in that round, and cumulative drift in never-edited regions.}
\label{tab:multiround_hist}
\scriptsize\setlength{\tabcolsep}{3pt}
\begin{tabular}{@{}lrrrr@{}}
\toprule
 & & & \multicolumn{2}{c}{Full history}\\\cmidrule(l){4-5}
 & Ours & Sequential & Edit list & Target state\\\midrule
Mean drift per round\dn & 0.46 & 3.70 & 5.09 & 8.95\\
Cumulative drift, round 7\dn & 3.53 & 11.62 & 14.91 & 12.40\\
Lip strength lip $\Delta E$ ratio (ideal 0.5) & 0.55 & 0.98 & 1.36 & 0.95\\
Eye hue error ($^\circ$)\dn & 16.7 & 17.9 & 37.1 & 36.8\\
Swap: $\Delta E$ to reference\dn & 8.9 & 14.1 & 25.6 & 28.2\\
Correct layer order (5 looks)\up & 5/5 & 1/5 & 1/5 & 1/5\\
Design moved\up & 97\% & 27\% & 33\% & 27\%\\\bottomrule
\end{tabular}
\end{table}

\noindent\textbf{Full history.} To test whether the gap stems only from FLUX.2-klein seeing its previous output, we also give it the full history: every round regenerates from the source face with the three references of the look, including the new eye reference after the swap. The prompt is either the transfer prompt followed by all edits so far, or a description of the target state only, written before its runs. Neither variant helps (Table~\ref{tab:multiround_hist}). With the edit list, both the mean drift per round and the cumulative drift rise significantly over sequential editing under a paired bootstrap. The target-state prompt drifts more per round, although its cumulative drift does not differ significantly from that of sequential editing. Both variants clearly raise the hue error and move the swapped eyes further from the new reference. Regeneration thus replaces accumulated error with fresh changes in every round, whereas our layers keep unedited parts fixed.

\begin{table}[!ht]
\centering
\caption{\textbf{Edit coverage by method family.}}
\label{tab:coverage}
\scriptsize\setlength{\tabcolsep}{3pt}
\begin{tabular}{@{}lccc@{}}
\toprule
Edit & Ours & Image editors & Reference transfer\\\midrule
Strength, hue, remove, restore & closed form & regenerate & ---\\
Swap reference & one render & regenerate & re-run\\
Layer order, move design & closed form + one render & regenerate & ---\\
\bottomrule
\end{tabular}
\end{table}

\noindent\textbf{Predictability.} Our edits act directly on explicit layers, so they are predictable and reversible, whereas image editors regenerate the image in every round (Table~\ref{tab:coverage}).

\begin{figure*}[t]
\centering
\includegraphics[width=\textwidth]{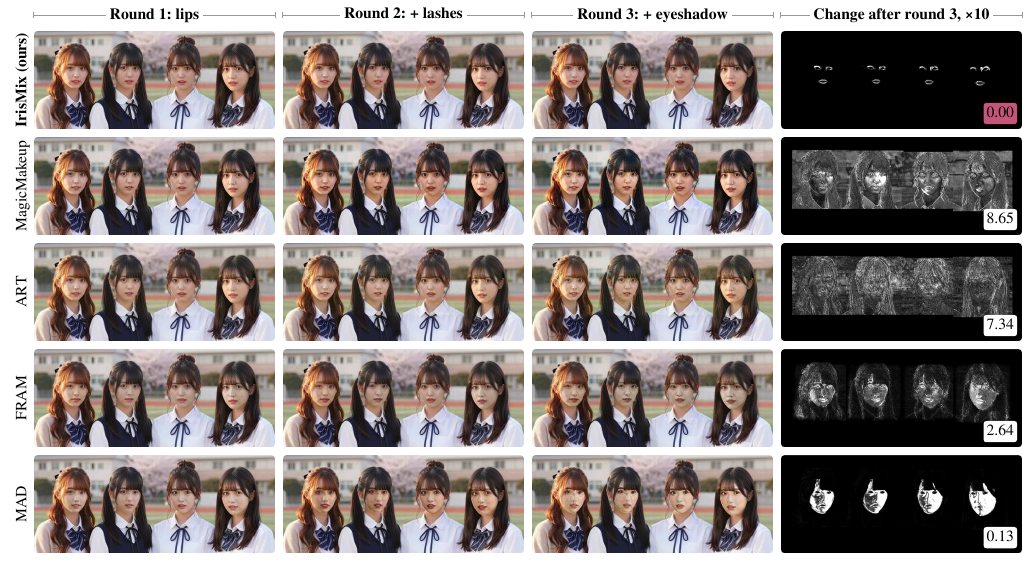}
\caption{\textbf{Multi-face multi-round makeup transfer on an everyday group photo.} The rounds add lips, lashes and eyeshadow to all four faces. The last column shows the absolute change from the makeup-removed photo after three rounds ($\times$10) with its mean outside the faces (0--255). The baselines edit their own previous output in every round, so each round regenerates the whole face crop and the changes accumulate: MagicMakeup and ART alter the hair, clothing and background inside every crop, FRAM alters the hair and skin, and MAD repaints the skin of every face. IrisMix adds each round as a layer to the cached stack of each face and renders the stack once on the makeup-removed face, so no round edits an earlier output and the photo outside the made-up regions stays unchanged.}
\label{fig:group_rounds_daily}
\end{figure*}

\begin{figure*}[t]
\centering
\includegraphics[width=\textwidth]{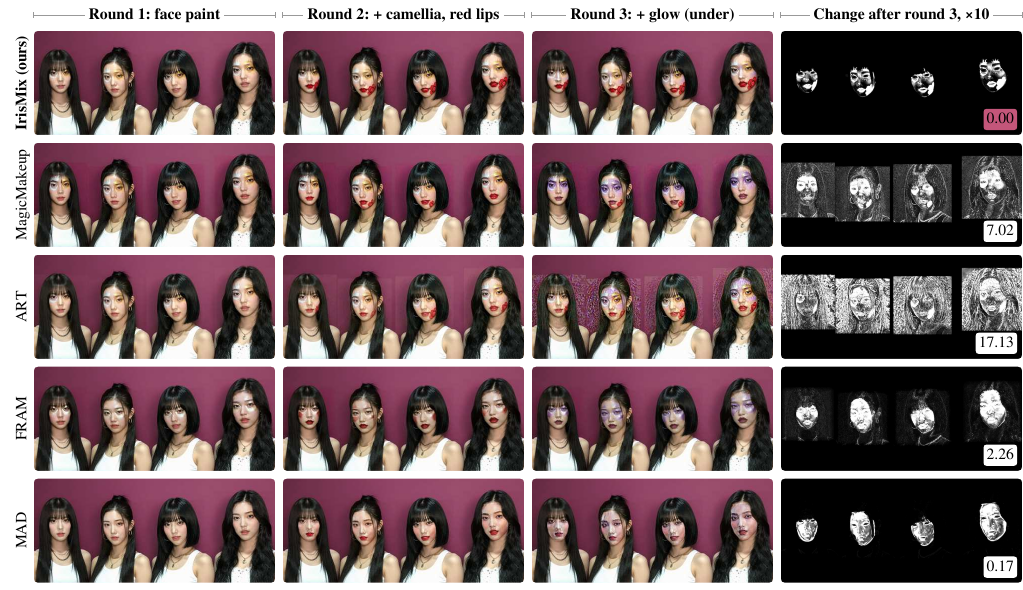}
\caption{\textbf{Multi-face multi-round makeup transfer on a group photo with face paint.} The rounds add face paint, then a camellia with red lips, then a glow placed under the earlier layers. The last column is as in Figure~\ref{fig:group_rounds_daily}. The baselines bake every round into the pixels, so a later design can only be painted over the earlier ones and every round changes the crop again: MagicMakeup and FRAM lay the glow on top of the camellia and the face paint, ART accumulates grain over the hair and background, and MAD repaints the skin of every face. IrisMix keeps each round as a separate layer, so the glow is inserted under the earlier layers, the stack is rendered once on the makeup-removed face, and the photo outside the made-up regions stays unchanged.}
\label{fig:group_rounds_creative}
\end{figure*}

\noindent\textbf{Multi-face multi-round makeup transfer.} Figures~\ref{fig:group_rounds_daily} and~\ref{fig:group_rounds_creative} apply three rounds of makeup to every face of two generated group photos, one with everyday makeup and one with face paint. All methods start from the same makeup-removed photo, use the same square crop and the same reference for each face, and continue from their own previous output, and each crop is pasted back into the photo. Our rounds add one layer to the stack of each face and render the whole stack once on the makeup-removed face, so no round edits a previous output; in the face-paint photo the third layer is placed under the earlier ones. In the everyday photo, the lash and eyeshadow rounds use the exact LayerBeauty-DailyScene layers of those references, as daily washes do in the transfer rules (Appendix~\ref{app:protocol}). Outside the faces our photos stay nearly unchanged after three rounds, whereas MagicMakeup, ART and FRAM change hair and background more with every round. MAD stays local but changes the skin inside the faces.

\begin{figure*}[t]
\centering
\includegraphics[width=\textwidth]{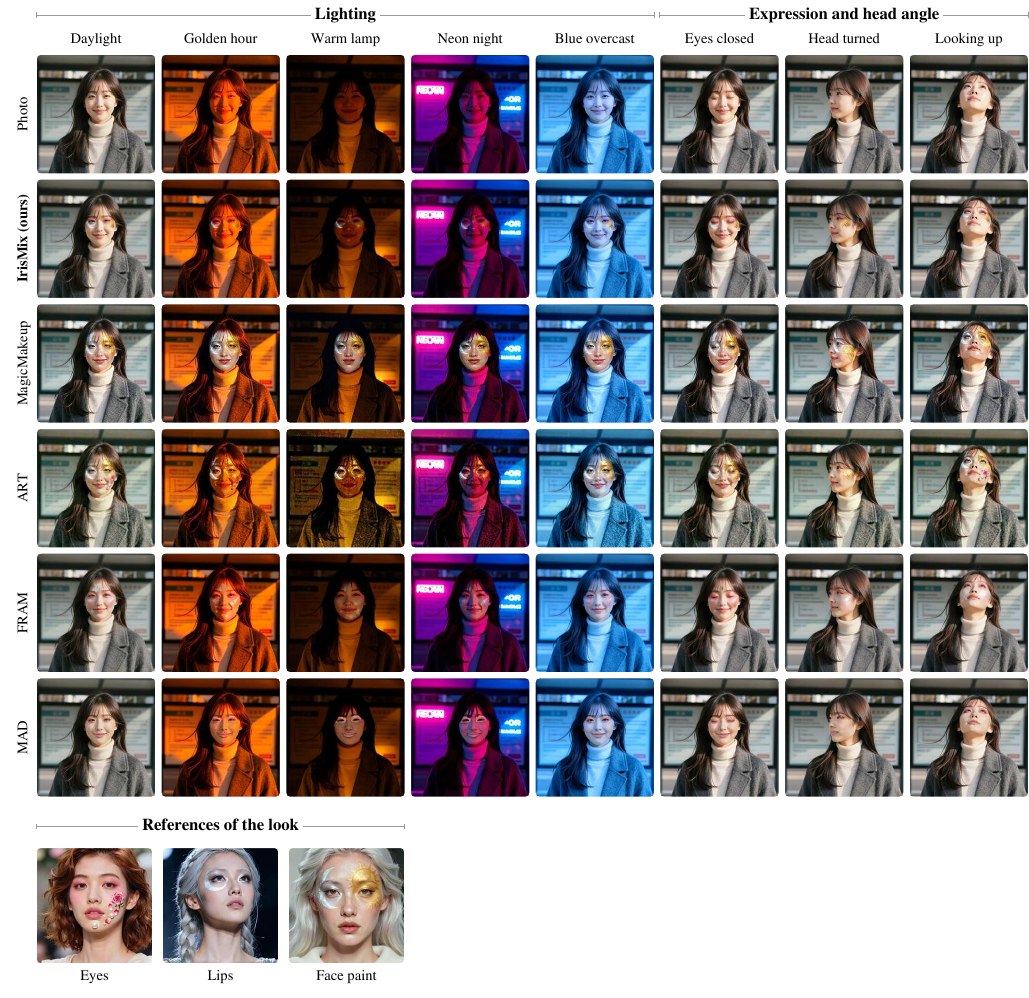}
\caption{\textbf{Comparison of one look under different lighting, expressions and head angles.} Each column shows a photo of the same synthetic person and the results for the look combining the three references at the bottom. IrisMix decomposes the references once and renders their layers on each photo, so the face paint takes on the color and brightness of each light, stays on the closed eyelids and follows the turned and raised head. MagicMakeup keeps the design at about the same brightness in every light, so it stands out in the dark and colored scenes, and it widens the design when the head turns. ART transfers each reference as a whole, so the flowers of the eye reference spread over the face, and its repeated steps add grain to the dark scene. FRAM and MAD keep the photo but transfer little of the design.}
\label{fig:lighting}
\end{figure*}

\subsection{Renderer Analysis on the Validation Set}
\label{app:val_renderer}
Tables~\ref{tab:val_cond}--\ref{tab:val_layer} report the validation measurements behind the renderer analysis of Appendix~\ref{app:model} and extend Table~\ref{tab:s2abl}, whose rows they reproduce. \emph{Band $\Delta E$} is defined at the start of Appendix~\ref{app:model}.

\begin{table*}[!t]
\centering
\caption{\textbf{Renderer conditions, sampling and editable region on the validation looks.} The first row shows the warp alone, without rendering. The lower blocks start from the four-step setting of Table~\ref{tab:s2abl}.}
\label{tab:val_cond}
\scriptsize\setlength{\tabcolsep}{2.5pt}
\resizebox{0.78\textwidth}{!}{\input{figures/tables/supp_model_cond_val}}
\end{table*}

\begin{table*}[!t]
\centering
\caption{\textbf{Renderer training ablations on the validation looks.} Each variant changes one component of the training recipe; the first and last rows are the training rows of Table~\ref{tab:s2abl}.}
\label{tab:val_train}
\scriptsize\setlength{\tabcolsep}{2.5pt}
\resizebox{0.78\textwidth}{!}{\input{figures/tables/supp_model_train_val}}
\end{table*}

\begin{table*}[!t]
\centering
\caption{\textbf{Terms of the output coverage $\tilde{\boldsymbol\alpha}$ on validation renders.} Re-compositing error (mean and 99th percentile, $\times100$), share of covered pixels whose color $\mathbf F'$ must be clipped, and error of halving the strength.}
\label{tab:val_layer}
\scriptsize\setlength{\tabcolsep}{2.5pt}
\resizebox{0.72\textwidth}{!}{\input{figures/tables/supp_model_layer_val}}
\end{table*}

%% file: sections/tab_efficiency.tex
\begin{table}[!htb]
\centering\setlength{\aboverulesep}{0.2ex}\setlength{\belowrulesep}{0.3ex}
\caption{\textbf{Effect of one-step layers on inference cost.} A reference is decomposed once in one step and reused, so each rendered image needs a single transformer evaluation, and edits of the returned layer need none. We run each method at batch 1 on one RTX PRO 6000 Blackwell Max-Q with its official settings and precision. We report sampling steps including chained calls, generator and text-encoder parameters (B), peak GPU memory (GiB), and median runtime per image in seconds. We time decomposition per image and our decomposer per reference; Table~\ref{tab:compare} adds it for the three references of a look. MAD appears in Table~\ref{tab:compare} as a transfer method; this table lists its removal path.}
\label{tab:efficiency}
\scriptsize\setlength{\tabcolsep}{2.6pt}\renewcommand{\arraystretch}{0.9}
\resizebox{\linewidth}{!}{%
\begin{tabular}{@{}lrrrrrr@{}}
\toprule
Method & Steps & Params & Mem & TFLOPs & Time (s) & FPS\\\midrule
\multicolumn{7}{@{}l}{\textit{Makeup transfer, per image}}\\
Qwen-Image-Edit & 40 & 20.4\,+\,8.3 & 58.0 & 50{,}382 & 261.3 & 0.004\\
FLUX.2-klein & 4 & 3.9\,+\,4.0 & 17.5 & 1{,}091 & 7.0 & 0.14\\
MagicMakeup & 3$\times$28 & 12.9\,+\,5.3 & 39.7 & 24{,}948 & 245.1 & 0.004\\
FLUX-Makeup & 3$\times$28 & 12.6\,+\,4.9 & 35.3 & 16{,}565 & 97.4 & 0.010\\
FRAM & 50 & 1.6\,+\,0.3 & 10.5 & 122 & 7.2 & 0.14\\
ART & 3$\times$28 & 12.0\,+\,4.9 & 35.0 & 23{,}719 & 155.3 & 0.006\\
Stable-Makeup & 3$\times$30 & 1.9\,+\,0.1 & 9.3 & 310 & 15.4 & 0.065\\
EleGANt & 1 & 0.01 & 0.5 & 0.73 & 1.0 & 1.00\\
\midrule\multicolumn{7}{@{}l}{\textit{Decomposition, per image}}\\
FLUX.2-klein removal & 2$\times$4 & 3.9\,+\,4.0 & 9.8 & 644 & 4.4 & 0.23\\
MAD & 2$\times$700 & 0.3 & 2.3 & 8{,}281 & 116.7 & 0.009\\
Qwen-Layered & 50 & 20.4\,+\,8.3 & 60.3 & 20{,}018 & 111.9 & 0.009\\
Stable-Layers & 50 & 20.5\,+\,8.3 & 45.1 & 19{,}763 & 164.4 & 0.006\\
\midrule
\rowcolor{oursrow}\method{} decomposer (per reference) & 1 & 3.9\,+\,4.0 & 10.5 & 44.7 & 0.45 & 2.20\\
\rowcolor{oursrow}\method{} warp only & 0 & 0.08 & -- & 0.88 & 1.3 & 0.78\\
\rowcolor{oursrow}\method{} + rendering & 1 & 3.9\,+\,4.0 & 18.0 & 66.5 & 2.4 & 0.41\\
\rowcolor{oursrow}\method{} edit (strength, color, part) & 0 & 0 & -- & 0 & 0.0013 & 751\\
\bottomrule
\end{tabular}}\vspace{-2mm}
\end{table}

%% file: sections/tab_edit_cost.tex
\begin{table}[!ht]
\centering
\caption{\textbf{Comparison of operation costs.} Recovery is paid once per reference; layer edits need no network. Render image includes warping and composing, so a look of three references costs three decompositions and one render (Table~\ref{tab:compare}). Batch-one runtime on one RTX PRO 6000 Blackwell Max-Q.}
\label{tab:edit_cost}
\footnotesize\setlength{\tabcolsep}{2pt}
\begin{tabular*}{\linewidth}{@{}l@{\extracolsep{\fill}}lrrrr@{}}
\toprule
Operation & Unit & Steps & TFLOPs & Mem. (GiB) & Time (s)\\\midrule
Decompose & reference & 1 & 44.7 & 10.5 & 0.45\\
Warp + compose & image & 0 & 0.88 & --- & 1.3\\
Render image & image & 1 & 66.5 & 18.0 & 2.4\\
Layer edit & edit & 0 & 0 & --- & \textbf{0.0013}\\
\bottomrule
\end{tabular*}
\end{table}

%% file: figures/tables/supp_model_cond_val.tex
\begin{tabular}{@{}lrrrr@{}}
\toprule
Setting & ID\up & DINO$_{\text{part}}$\up & Color $\Delta E_{\text{part}}$\dn & Band $\Delta E$\dn \\\midrule
Warp only & 0.817 & 0.703 & 9.36 & 0.01 \\
\midrule
\rowcolor{oursrow}Reported: $\sigma=0.65$, 1 step, $\boldsymbol\Omega$ 2\,px & 0.687 & 0.756 & 8.47 & 0.47 \\
\quad $\sigma=0.7$ & 0.675 & 0.756 & 8.59 & 0.54 \\
\quad $\sigma=0.7$, brows outside $\boldsymbol\Omega$ & 0.674 & 0.756 & 8.65 & 0.52 \\
\quad $\sigma=0.7$, 4 steps, $\boldsymbol\Omega$ 6\,px & 0.666 & 0.758 & 8.72 & 2.03 \\
\midrule
\multicolumn{5}{@{}l}{\emph{Changes to $\sigma=0.7$, 4 steps, $\boldsymbol\Omega$ 6\,px}}\\
w/o full reference layer & 0.778 & 0.731 & 9.32 & 1.82 \\
w/o every condition & 0.804 & 0.709 & 9.88 & 1.23 \\
w/o crops, caption and maps (gray), 1 step & 0.660 & 0.754 & 8.86 & 2.60 \\
w/o their tokens, 1 step & 0.673 & 0.754 & 8.76 & 2.39 \\
\midrule
w/o masked residual & 0.640 & 0.756 & 9.09 & 2.52 \\
$\boldsymbol\Omega$ dilated 2\,px & 0.671 & 0.757 & 8.69 & 0.49 \\
$\boldsymbol\Omega$ dilated 12\,px & 0.664 & 0.757 & 8.73 & 2.40 \\
$\boldsymbol\Omega$ dilated 24\,px & 0.664 & 0.756 & 8.74 & 2.40 \\
\bottomrule
\end{tabular}

%% file: figures/tables/supp_model_train_val.tex
\begin{tabular}{@{}lrrr@{}}
\toprule
Training & ID\up & DINO$_{\text{part}}$\up & Color $\Delta E_{\text{part}}$\dn \\\midrule
\rowcolor{oursrow}w/ self-transfer & 0.687 & 0.750 & 8.40 \\
w/ self-transfer, LoRA$_{\text{ren}}$ from scratch & 0.666 & 0.749 & 8.45 \\
w/o self-transfer & 0.743 & 0.743 & 9.07 \\
\bottomrule
\end{tabular}

%% file: figures/tables/supp_model_layer_val.tex
\begin{tabular}{@{}lrrrr@{}}
\toprule
$\tilde{\boldsymbol\alpha}$ & Recomp.\dn & Recomp.\ p99\dn & Clipped $\mathbf F'$ (\%)\dn & $\times0.5$ error\dn \\\midrule
\rowcolor{oursrow}$\max(\boldsymbol\alpha_{\text{d}},\boldsymbol\alpha,\boldsymbol\alpha_{\hat{\mathbf I}})$ (ours) & 0.13 & 2.6 & 0.0 & 0.07 \\
$\max(\boldsymbol\alpha_{\text{d}},\boldsymbol\alpha)$ & 5.24 & 41.3 & 47.5 & 2.62 \\
\bottomrule
\end{tabular}

%% file: sections/G-realworld.tex
\section{Real-World Layer Decomposition and Makeup Transfer}
\label{app:realworld}
This section applies the reported decomposer and renderer unchanged to publicly released photographs and, for decomposition, also to three digital illustrations.

\subsection{Makeup Transfer to Real Photographs}
\begin{figure*}[t]
\centering
\includegraphics[width=\textwidth]{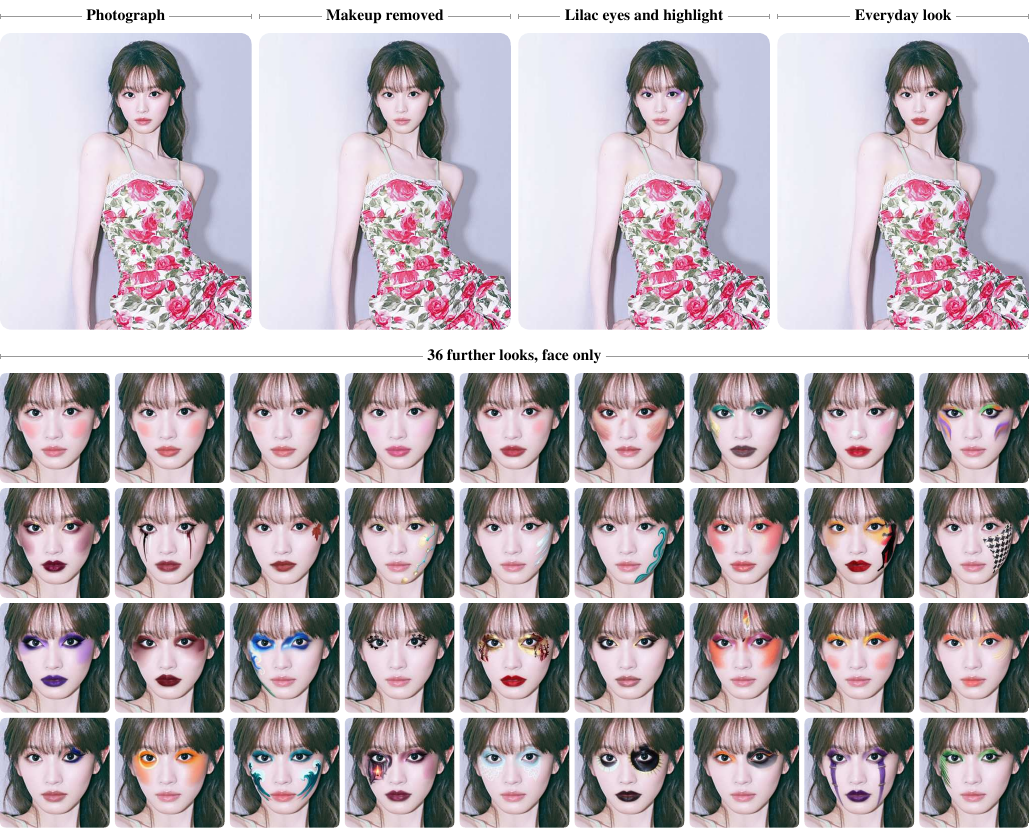}
\caption{\textbf{Re-make-up of a real portrait.} From left: the photograph, the bare face recovered by the decomposer, and two looks transferred from CreativeScene references onto that bare face; below, thirty-six further looks on the same bare face, shown as face crops. The photograph is a publicly released portrait used for research purposes only.}
\label{fig:realphoto}
\end{figure*}

\begin{figure*}[t]
\centering
\includegraphics[width=\textwidth]{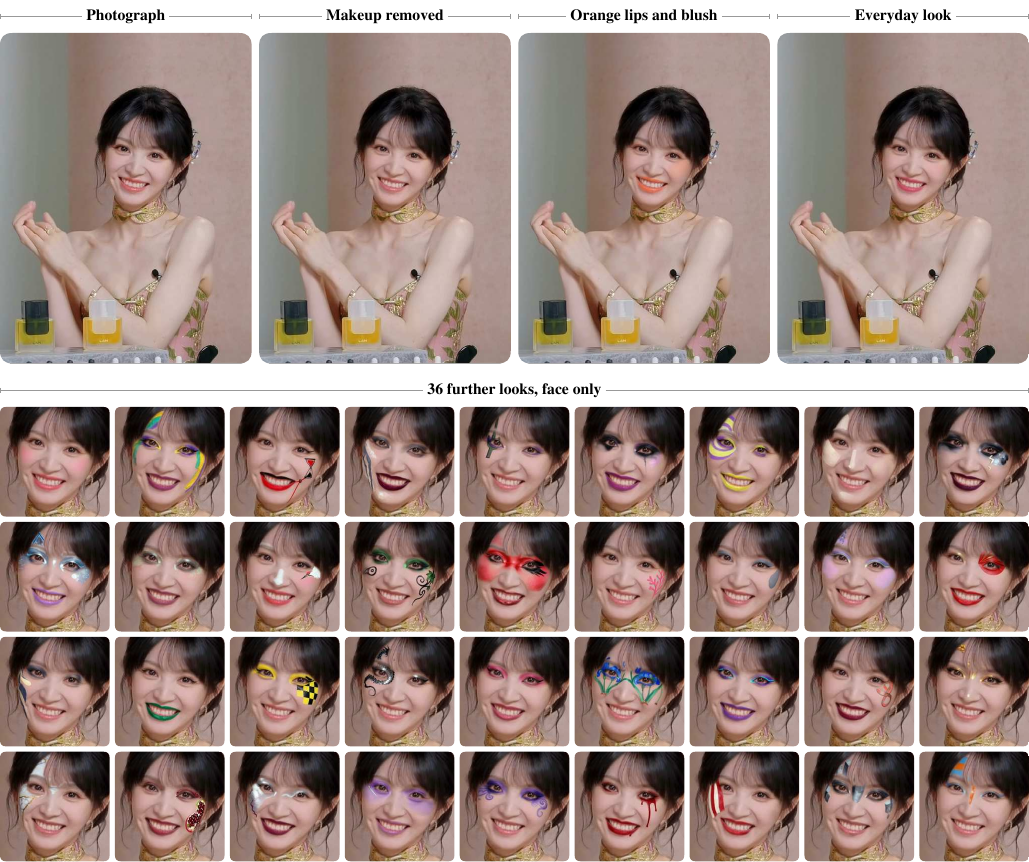}
\caption{\textbf{Re-make-up of a second real portrait.} Layout as in Figure~\ref{fig:realphoto}. The photograph is a publicly released portrait used for research purposes only.}
\label{fig:realphoto2}
\end{figure*}

\begin{figure*}[t]
\centering
\includegraphics[width=\textwidth]{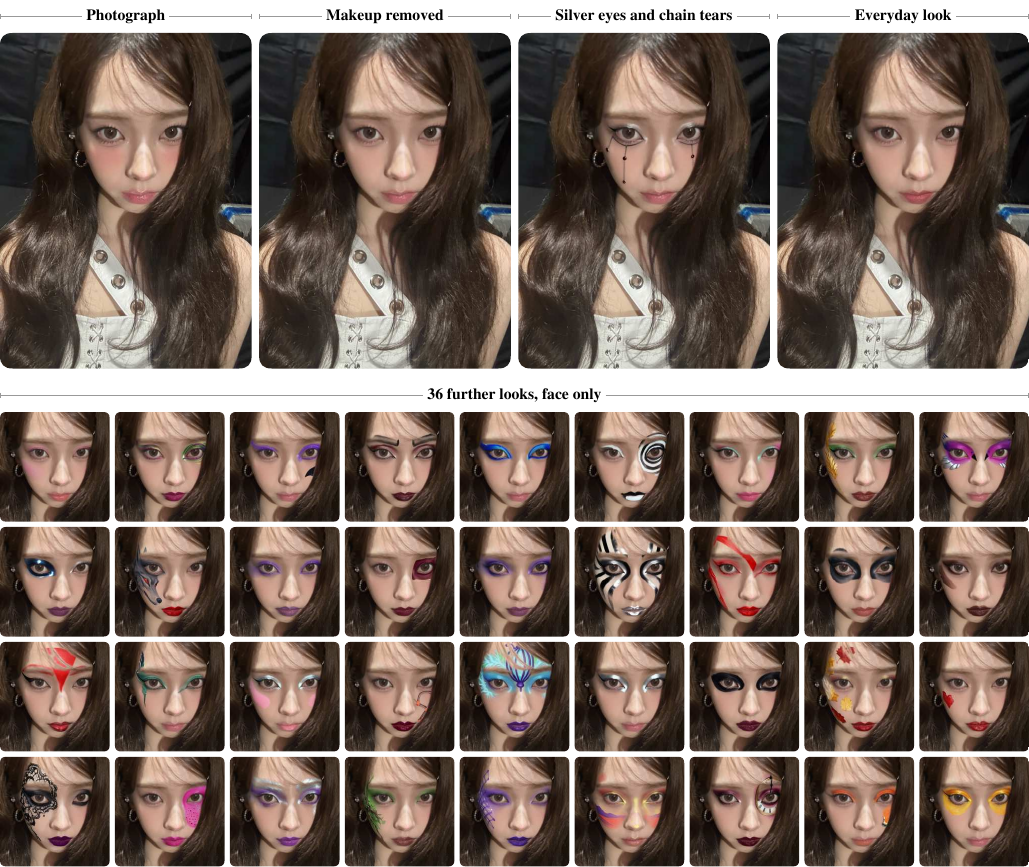}
\caption{\textbf{Re-make-up of a close-up selfie.} Layout as in Figure~\ref{fig:realphoto}. The photograph is a publicly released portrait used for research purposes only.}
\label{fig:realphoto3}
\end{figure*}

\begin{figure*}[p]
\centering
\includegraphics[width=\textwidth]{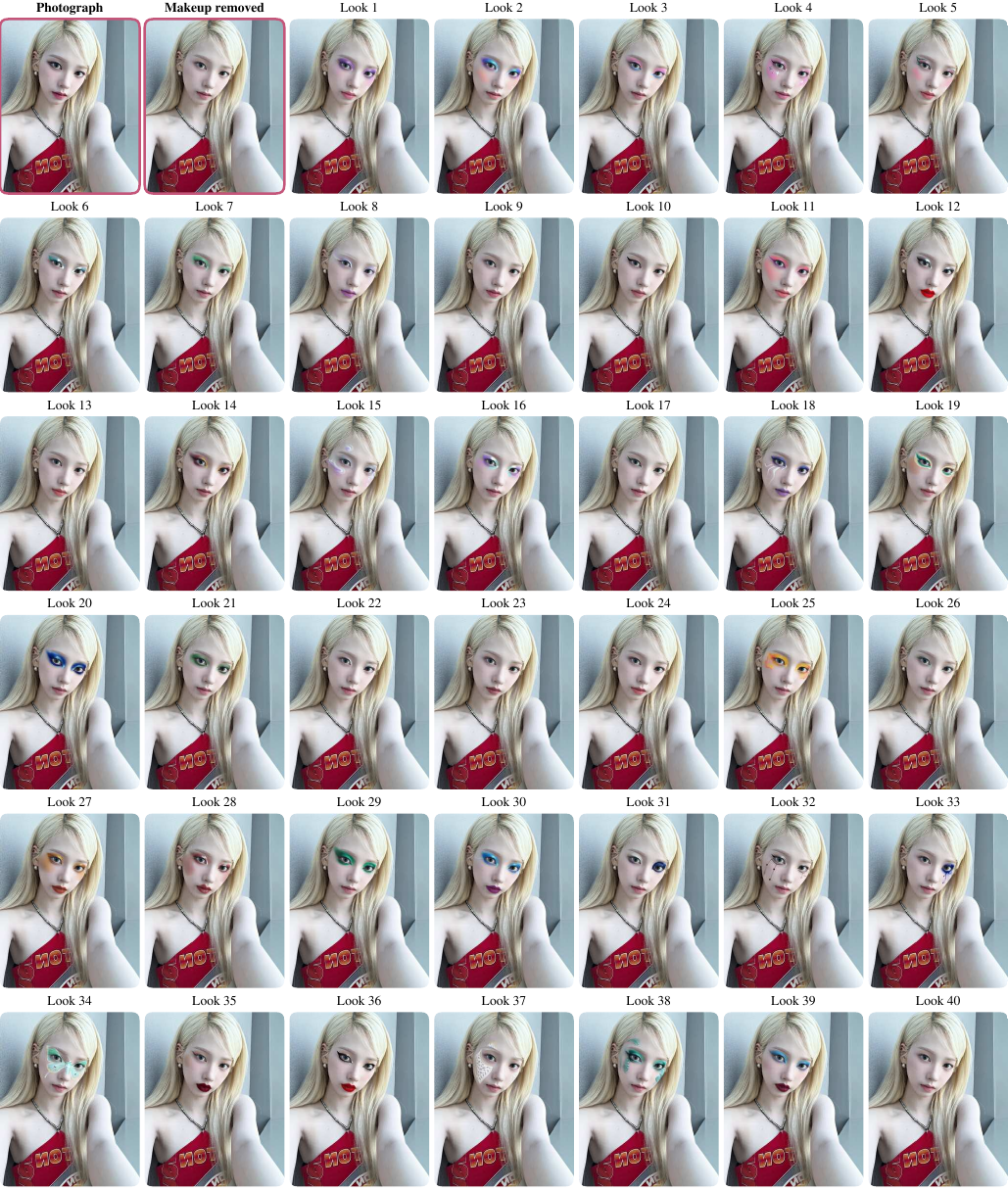}
\caption{\textbf{Forty looks on one real portrait.} The photograph, the bare face recovered by the decomposer (framed), and forty looks from forty CreativeScene references rendered on that bare face. The photograph is a publicly released portrait used for research purposes only.}
\label{fig:realphoto_grid}
\end{figure*}

\begin{figure*}[p]
\centering
\includegraphics[width=\textwidth]{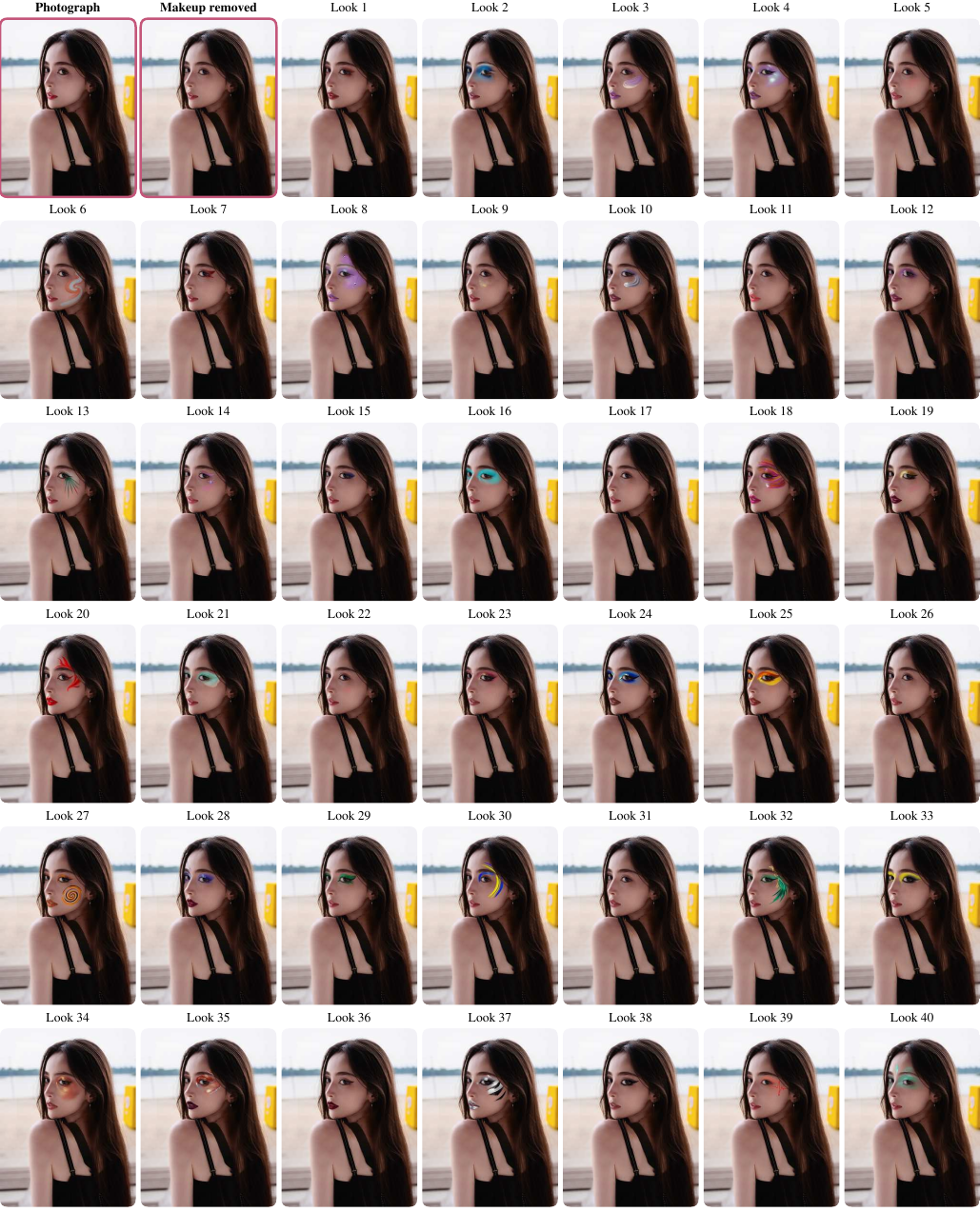}
\caption{\textbf{Forty further looks on a real portrait seen from the side.} Layout as in Figure~\ref{fig:realphoto_grid}, with forty other CreativeScene references. The photograph is a publicly released portrait used for research purposes only.}
\label{fig:realphoto_grid2}
\end{figure*}

\begin{figure*}[p]
\centering
\includegraphics[width=\textwidth]{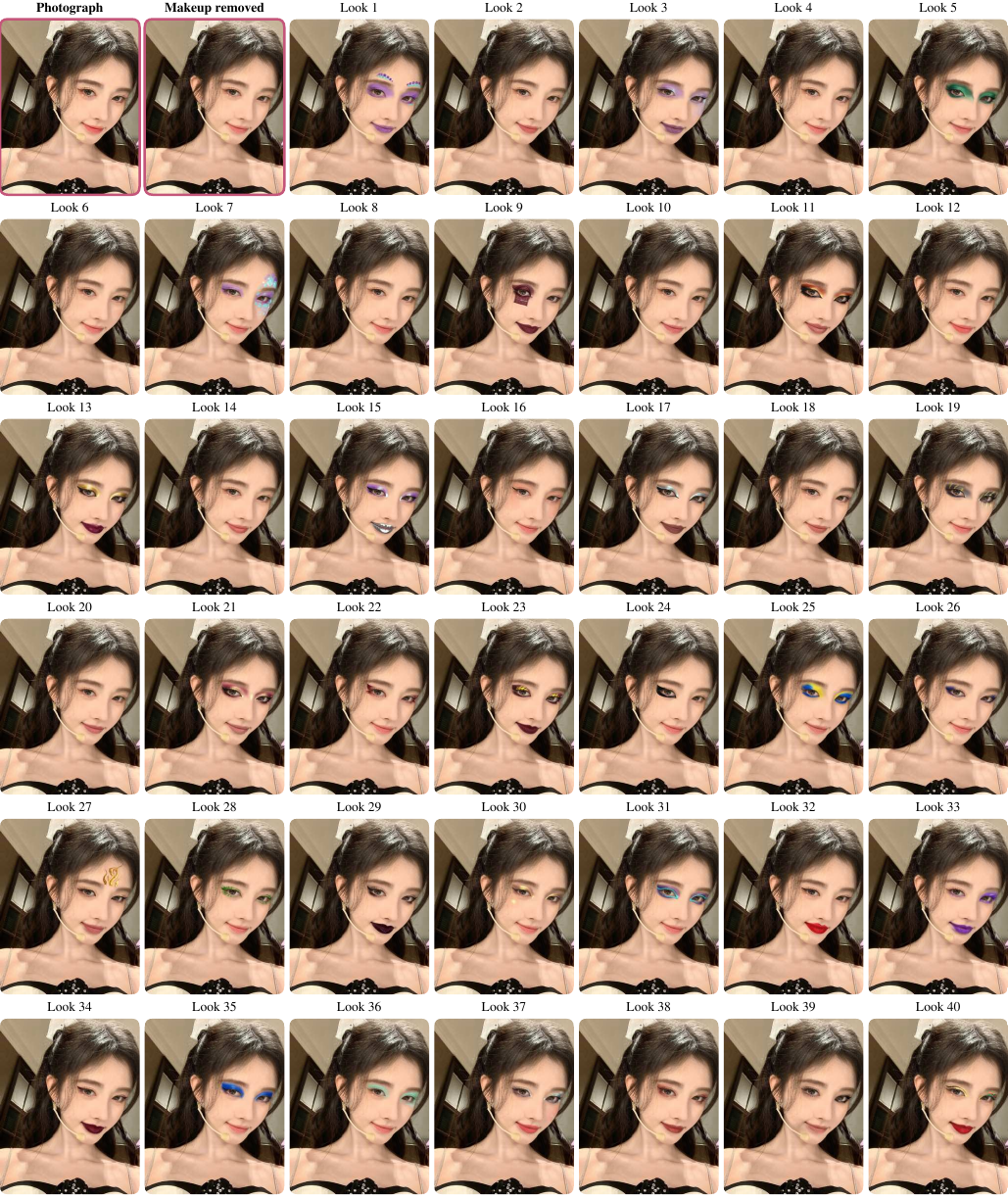}
\caption{\textbf{Forty looks on a selfie with a headset.} Layout as in Figure~\ref{fig:realphoto_grid}. The headset is kept, and the lip color stays off the microphone. The photograph is a publicly released portrait used for research purposes only.}
\label{fig:realphoto_grid3}
\end{figure*}

\begin{figure*}[p]
\centering
\includegraphics[width=\textwidth]{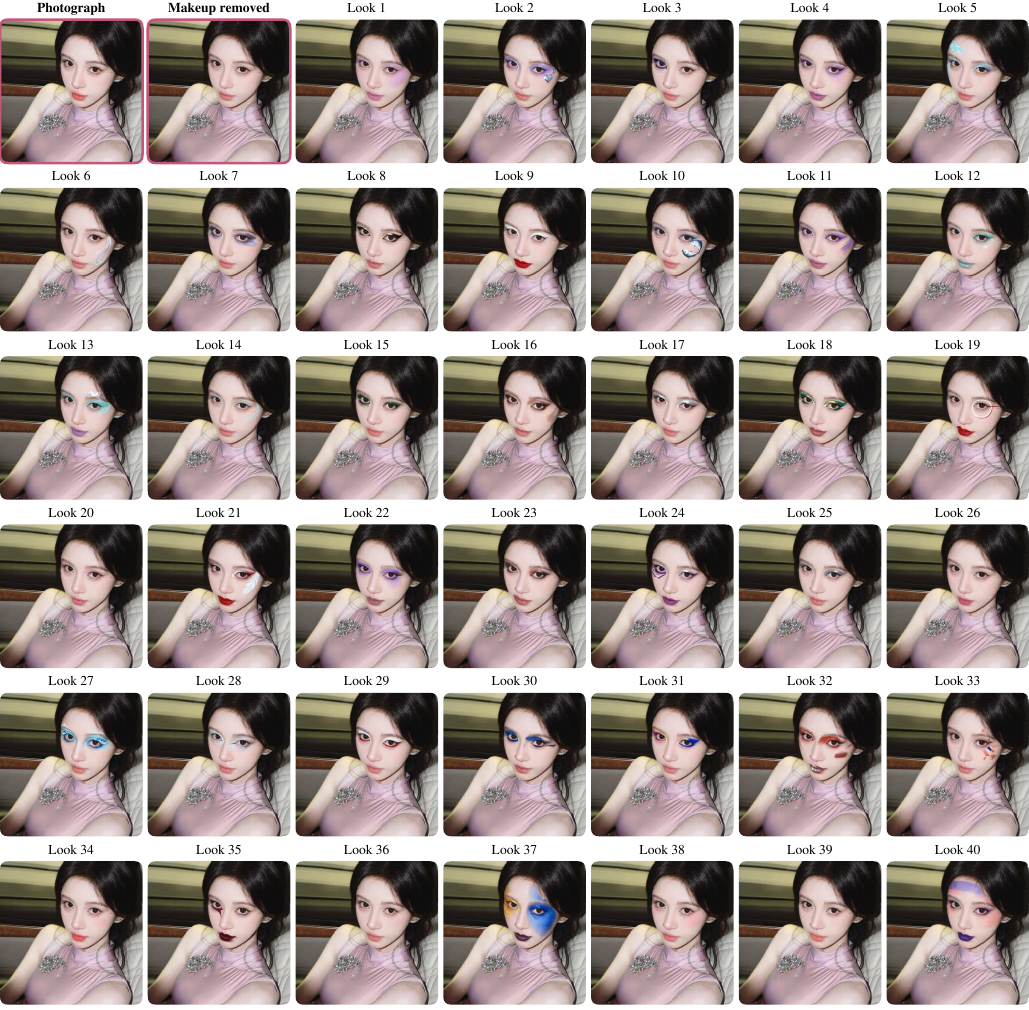}
\caption{\textbf{Forty looks on a portrait taken in a car.} Layout as in Figure~\ref{fig:realphoto_grid}, with forty further CreativeScene references. The photograph is a publicly released portrait used for research purposes only.}
\label{fig:realphoto_grid4}
\end{figure*}

Figures~\ref{fig:realphoto}, \ref{fig:realphoto2} and~\ref{fig:realphoto3} show three real portraits, including one with an open smile and a close-up selfie. The decomposer first removes the makeup the person already wears, and looks from CreativeScene references are then rendered on the recovered bare face: two in full frame and thirty-six more as face crops. Only the face changes, the hair, clothing and background of each photograph are kept, and in the smiling portrait the lip color stays on the lips and leaves the teeth unchanged. Figures~\ref{fig:realphoto_grid}--\ref{fig:realphoto_grid4} each render forty looks from CreativeScene references, chosen across many theme sets, on a real portrait after its makeup is removed once; the portraits include a view from the side, a selfie with a headset and a photograph taken in a car.

\subsection{Decomposition of Real Photographs}
\begin{figure*}[p]
\centering
\includegraphics[width=\textwidth,height=0.9\textheight,keepaspectratio]{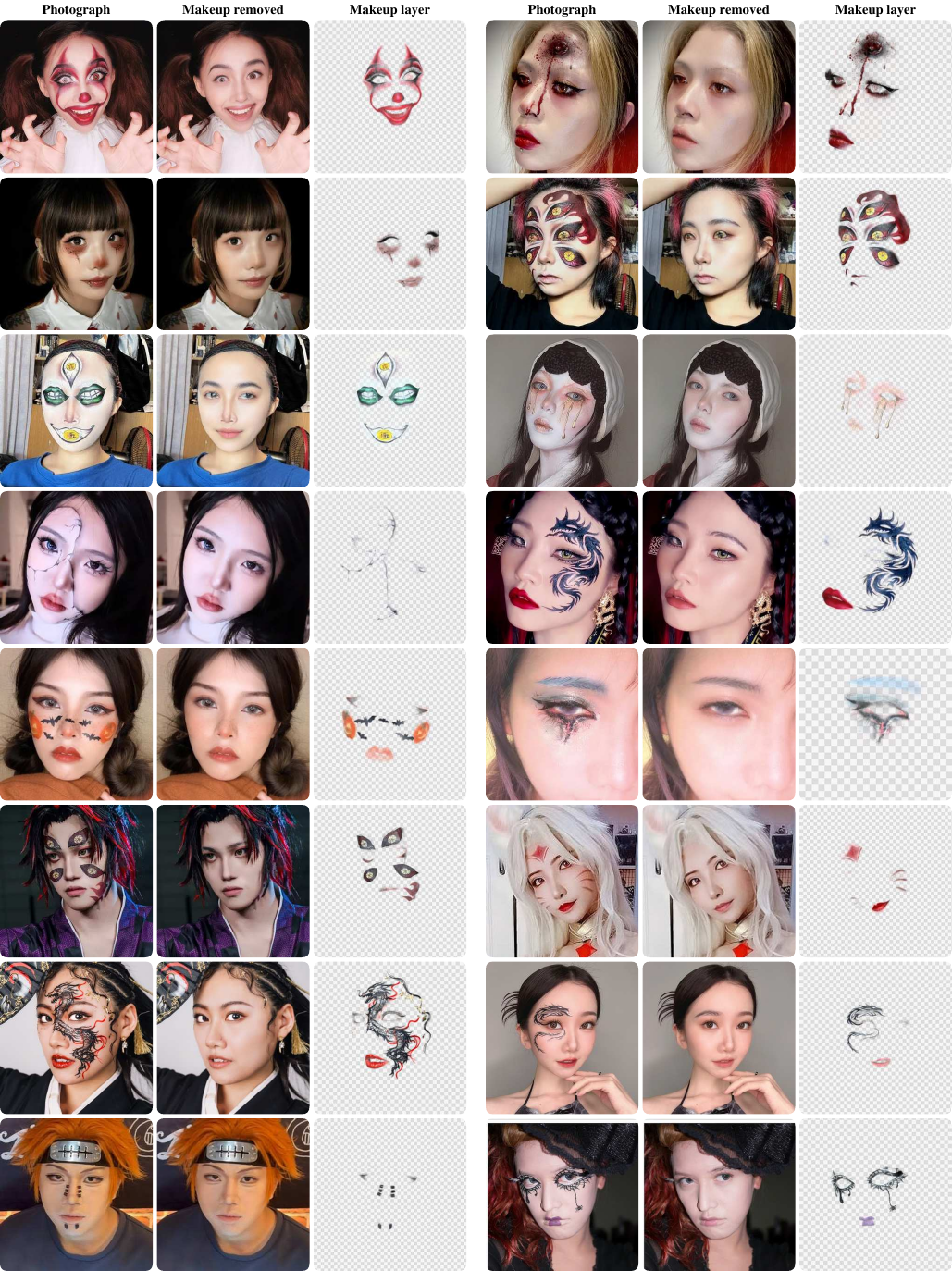}
\caption{\textbf{Decomposition of real portraits with Halloween, special-effects and cosplay makeup.} Each example shows the photograph, the bare face recovered by the decomposer and the extracted makeup layer, drawn over a checkerboard where the layer is transparent. The photographs are publicly released portraits used for research purposes only.}
\label{fig:realphoto_decomp_halloween}
\end{figure*}

\begin{figure*}[p]
\centering
\includegraphics[width=\textwidth,height=0.88\textheight,keepaspectratio]{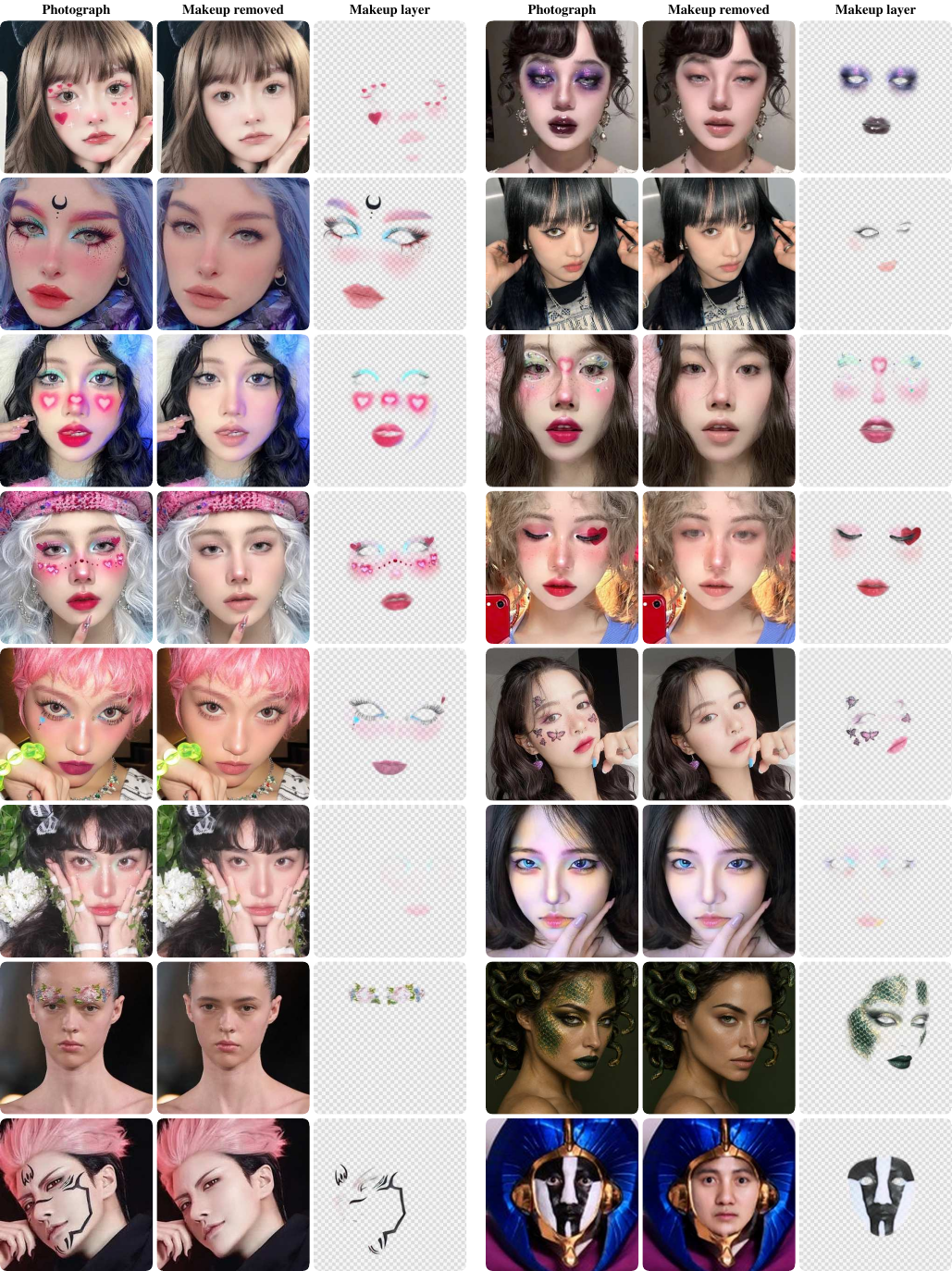}
\caption{\textbf{Decomposition of real portraits with decorative makeup.} Layout as in Figure~\ref{fig:realphoto_decomp_halloween}. Hearts, rhinestones, stickers, glitter and painted scales are extracted together with the eye and lip makeup, and so are the face markings of two cosplays in the last two examples. The photographs are publicly released portraits used for research purposes only.}
\label{fig:realphoto_decomp_decorative}
\end{figure*}

\begin{figure*}[p]
\centering
\includegraphics[width=\textwidth,height=0.9\textheight,keepaspectratio]{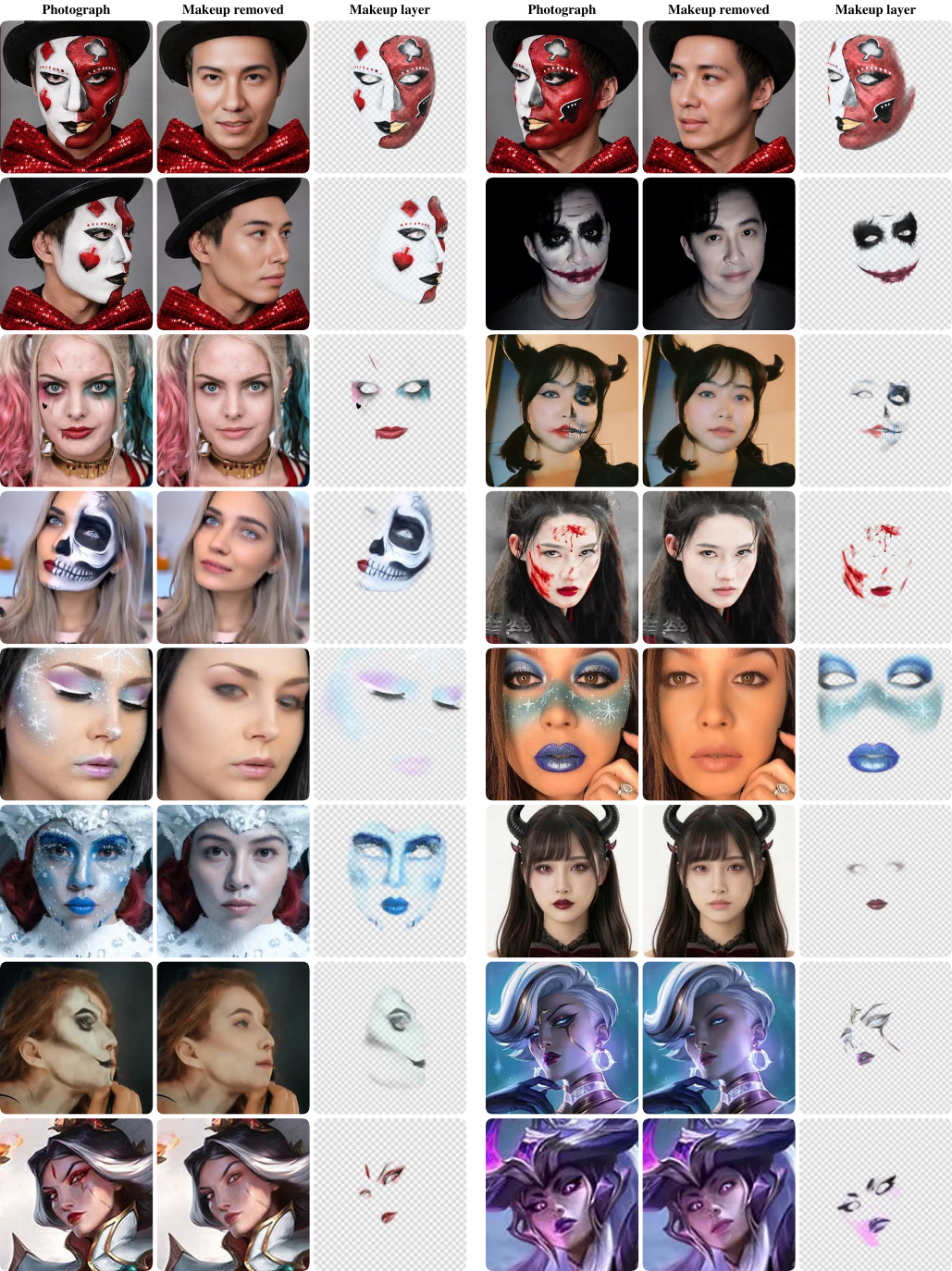}
\caption{\textbf{Decomposition of portraits with face paint and special-effects makeup.} Layout as in Figure~\ref{fig:realphoto_decomp_halloween}. The first three examples show one painted face from three viewpoints. The last three examples are digital illustrations rather than photographs. All images are publicly released and used for research purposes only.}
\label{fig:realphoto_decomp_paint}
\end{figure*}

Figures~\ref{fig:realphoto_decomp_halloween}--\ref{fig:realphoto_decomp_paint} run only the decomposer on real portraits with Halloween, special-effects, cosplay, decorative and full-face paint makeup, and on three digital illustrations, and show the input image, the recovered bare face and the extracted makeup layer. Painted designs, stickers, rhinestones and lip color move to the layer, while hair, accessories, colored contact lenses and the background stay in the bare face.

\subsection{Transfer Between Real Photographs}
\begin{figure*}[p]
\centering
\includegraphics[width=\textwidth]{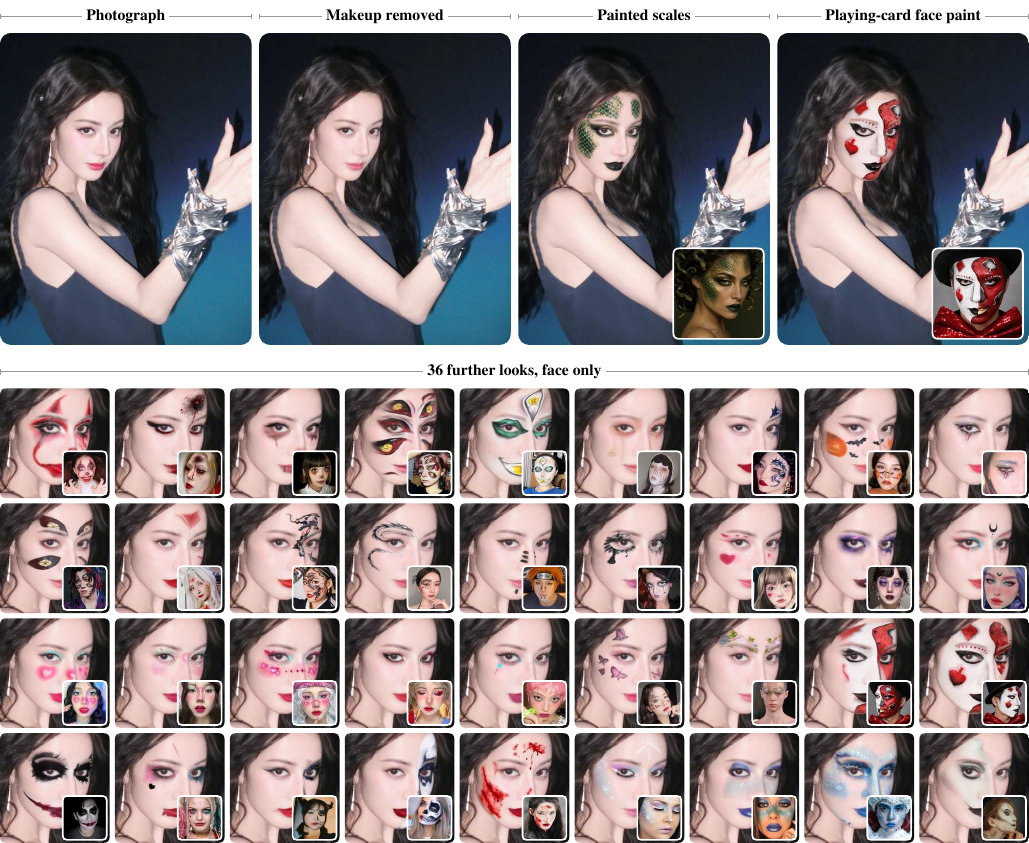}
\caption{\textbf{Real-to-real transfer.} Layout as in Figure~\ref{fig:realphoto}, but every look comes from a real portrait instead of a CreativeScene reference: the makeup layers recovered in Figures~\ref{fig:realphoto_decomp_halloween}--\ref{fig:realphoto_decomp_paint} are rendered on the bare face of another real portrait. The inset in each panel shows the reference photograph. All photographs are publicly released portraits used for research purposes only.}
\label{fig:realphoto_r2r}
\end{figure*}

Figure~\ref{fig:realphoto_r2r} transfers the makeup layers recovered from those real portraits onto another real portrait, so that both the reference and the source are photographs.

